\documentclass[acmlarge]{acmart}
\usepackage{algorithm}
\usepackage{algpseudocode}
\usepackage{diagbox}
\usepackage{booktabs}
\usepackage{graphicx}
\usepackage{wrapfig}
\usepackage{subcaption}
\usepackage{enumitem}
\usepackage{bbm}
\usepackage{adjustbox}
\usepackage{multirow}
\newcommand{\N}{\text{SynHAT}}
\AtBeginDocument{%
  \providecommand\BibTeX{{%
    \normalfont B\kern-0.5em{\scshape i\kern-0.25em b}\kern-0.8em\TeX}}}

\definecolor{rccolor}{RGB}{0, 120, 255}

\begin{document}
% \pagestyle{plain}
%%
%% The "title" command has an optional parameter,
%% allowing the author to define a "short title" to be used in page headers.
% VLDB
% \title{\N: A Generalizable Coarse-to-Fine Framework for Large-scale Asynchronous Trajectory Data Generation}
\setcopyright{cc}
\setcctype{by}
\acmJournal{IMWUT}
\acmYear{2026} \acmVolume{10} \acmNumber{2} \acmArticle{69}
\acmMonth{6} \acmDOI{10.1145/3810213}

\title{SynHAT: A Two-stage Coarse-to-Fine Diffusion Framework for Synthesizing Human Activity Traces}
% \title{\N: Coarse-then-Fine Granular Asynchronous Trajectory Data Generation with Spatio-Temporal Diffusion Models}

\author{Rongchao Xu}
\affiliation{%
  \institution{Florida State University}
  \city{Tallahassee}
  \state{Florida}
  \country{USA}
}
\email{rx21a@fsu.edu}

\author{Lin Jiang}
\affiliation{%
  \institution{Florida State University}
  \city{Tallahassee}
  \state{Florida}
  \country{USA}
}
\email{lin.jiang@fsu.edu}

\author{Dahai Yu}
\affiliation{%
  \institution{Florida State University}
  \city{Tallahassee}
  \state{Florida}
  \country{USA}
}
\email{dahai.yu@fsu.edu}

\author{Ximiao Li}
\affiliation{%
  \institution{Florida State University}
  \city{Tallahassee}
  \state{Florida}
  \country{USA}
}
\email{xl24g@fsu.edu}

\author{Guang Wang}
\authornote{Dr. Guang Wang is the corresponding author.}
\affiliation{%
  \institution{Florida State University}
  \city{Tallahassee}
  \state{Florida}
  \country{USA}
}
\email{guang@cs.fsu.edu}

% \author{Zhiqing Hong}
% \affiliation{%
%   \institution{Rutgers University}
%   \city{Piscataway}
%   \state{New Jersey}
%   \country{USA}
% }
% \email{zhiqing.hong@rutgers.edu}

% \author{Kunlin Cai}
% \affiliation{%
%   \institution{University of California, Los Angeles}
%   \city{Los Angeles}
%   \country{USA}
% }
% \email{kunlin96@g.ucla.edu}

% \author{Yuan Tian}
% \affiliation{%
%   \institution{University of California, Los Angeles}
%   \city{Los Angeles}
%   \country{USA}
% }
% \email{yuant@ucla.edu}

% \author{Guang Wang}
% \affiliation{%
%   \institution{Florida State University}
%   \city{Tallahassee}
%   \state{Florida}
%   \country{USA}
% }
% \email{guang@cs.fsu.edu}
% \author{Anonymous Author(s)}

\begin{abstract}
Human activity traces (HATs) play a crucial role in numerous real-world applications such as human mobility modeling, trace prediction, Point-of-Interest (POI) recommendation, and urban planning. However, increasing concerns over data privacy have significantly restricted access to authentic large-scale HATs. Fortunately, recent advances in generative AI open new opportunities to synthesize realistic yet privacy-preserving HATs that can support diverse applications. Despite this promise, two key challenges remain. (i) HATs (e.g., user-level POI check-in traces) are highly irregular and dynamic with long and variable time intervals, which makes it difficult to effectively capture their complex spatio-temporal patterns and intrinsic distributions. (ii) Generative models are typically computationally intensive and resource-demanding, such that generating long-term, fine-grained HATs incurs substantial computational overhead.
To address these challenges, we propose \N, a computationally-efficient coarse-to-fine HAT synthesis framework based on a novel spatio-temporal denoising diffusion model. In stage 1, we design a Coarse-grained Human Activity Diffusion model (Coarse-HADiff) to capture the overall spatio-temporal (ST) dependencies of the constructed coarse-grained latent ST traces, which includes a novel Latent Spatio-Temporal UNet (LST-UNet) for denoising through dual Drift-Jitter branches for jointly modeling smooth spatial transitions and temporal variations. In stage 2, we design a three-step pipeline consisting of Behavior Pattern Extraction, Fine-HADiff that shares the same architecture as Coarse-HADiff, and Semantic Alignment to further synthesize fine-grained
latent ST traces based on the output from stage 1.
We extensively evaluate the proposed \N\ framework from diverse perspectives, including 
\textbf{effectiveness} for data \textit{fidelity}, \textit{utility}, and \textit{privacy}, 
\textbf{robustness}, and \textbf{scalability}. Experimental results on real-world HATs from four cities (Tokyo, New York, Stockholm, and Austin) in three countries show that SynHAT significantly outperforms state-of-the-art baselines by 52\% and 33\% on spatial and temporal metrics, respectively.

\end{abstract}



%%
%% Keywords. The author(s) should pick words that accurately describe
%% the work being presented. Separate the keywords with commas.

\maketitle

\section{Introduction}

% \rc{(ready)} Check-in traces collected by location-based social networks are essential for many real-world applications, including urban mobility understanding \cite{wang2019urban}, business location selection \cite{xie2016learning}, advertising \cite{jeon2021lightmove}, and pandemic control \cite{hao2020understanding}. 
% \begin{figure}[t]
%   \centering
%   \includegraphics[width=3.75in]{Figures/trajvisual.png}
%   % \vspace{-10pt}
%   % \caption{Example of an asynchronous POI check-in trace (red) and a synchronous GPS trace (green). The fixed interval of the GPS trace is 30 seconds.}
%   \caption{An illustration of the difference between a discrete asynchronous human activity trace with varying intervals (red) and a continuous synchronous GPS trace with a fixed interval of 30 seconds (green).}
%   \Description{Dataset visualization.}
%   \label{asynchronous}
%   \vspace{-15pt}
% \end{figure}

%in Location-based social networks (LBSNs) 
%Human Activity Trace Synthesis

Fine-grained human activity traces (HATs) are crucial for a wide range of real-world applications, including human mobility modeling~\cite{li2020systematic,hong2024crosshar, wang2019urban,xu2026geogen, yu2025uqgnn, cheng2025bts, hong2025experience}, next Point-of-Interest (POI) recommendation~\cite{next_poi_rec_3, hong2026urbanpoi}, business location optimization~\cite{xie2016learning}, trip purpose inference~\cite{liao2022wheels, jiang2025hcride, 10446624, hong2025llm4har}, and pandemic intervention planning~\cite{jiang2025uncertainty, yu2026healthmamba, 10.1145/3748636.3763223, shen2025learning}. However, obtaining large-scale HATs has become increasingly difficult for both academia and industry due to the substantial privacy concerns arising from the sensitive personal information contained in such data. Using user-level POI check-in traces as a representative example, existing studies have shown that as few as four check-in points can uniquely re-identify 90\% of individuals, since their behavioral patterns and frequently visited locations are often highly distinctive~\cite{de2015unique, shen2026cited, shen2026credit}. Therefore, to enable privacy-preserving data sharing and facilitate research on human activity analysis, it is essential to develop effective methods for generating synthetic yet realistic HATs that maintain high utility without compromising individual privacy.
%However, obtaining large-scale HAT datasets (e.g., user-level POI check-in traces) has become increasingly difficult for both academia and industry due to the high acquisition costs and escalating privacy concerns.

% Hence, to facilitate LBSN trace data sharing for the benefit of the research community, it is imperative to design effective models to help generate synthetic while realistic trace data that preserve the data utilities without compromising privacy.

% Due to the importance of trace data generation, many existing works have focused on this topic by designing different types of models \cite{isaacman2012human, jiang2016timegeo, yin2017generative, ouyang2018non, zhu2023difftraj}. 
In recent years, a wide range of approaches have been proposed~\cite{isaacman2012human, jiang2016timegeo, yin2017generative, ouyang2018non, zhu2023difftraj, xu2025autostdiff} for synthesizing HATs. Early studies primarily relied on rule- or physics-based generative frameworks. For instance, Yin et al.~\cite{yin2017generative} leveraged hidden Markov models to synthesize human activity patterns from cellular data, while Jiang et al.~\cite{jiang2016timegeo} developed TimeGeo, a physics-inspired framework that reproduces individual mobility traces without travel surveys.
More recently, generative adversarial networks (GANs)~\cite{ouyang2018non, feng2020learning, yu2017seqgan, yuan2022activity} have demonstrated remarkable success in producing diverse forms of HATs, ranging from individual mobility traces and POI visit sequences to large-scale location-based social networks. These models highlight the power of adversarial learning to capture both fine-grained behavioral regularities and global spatio-temporal dynamics.
However, GAN-based approaches often suffer from unstable training dynamics and mode collapse, which limit their reliability and consistency in synthesizing complex, long-term and fine-grained human activity behaviors.

% \vspace{-10pt}
\begin{figure*}[t]
  \centering
  \begin{subfigure}[t]{0.48\linewidth}
    \centering
    \includegraphics[height=1.38in]{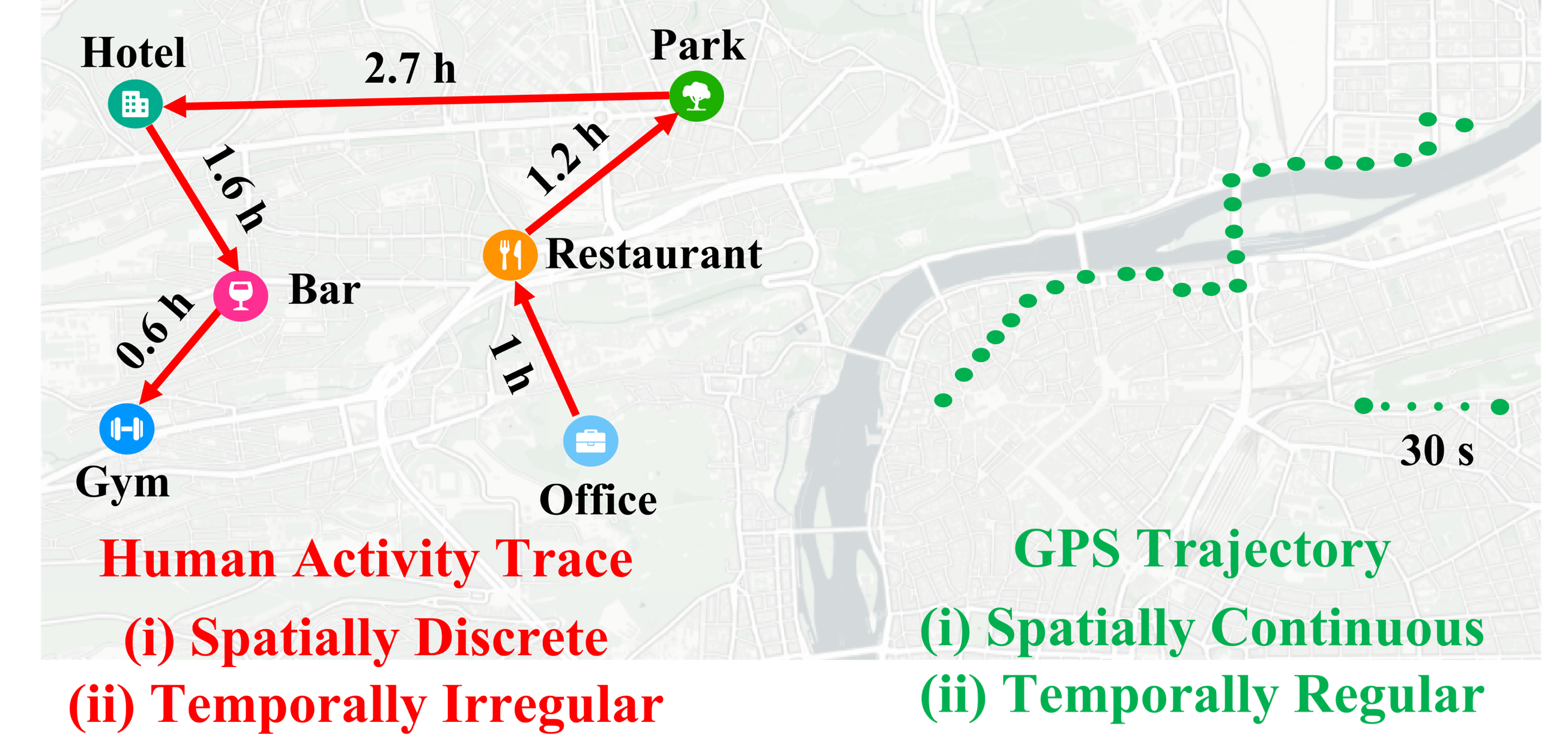}
    \caption{Difference between a HAT (red) with an irregular interval and a GPS trajectory (green) with a 30s-fixed interval.}
    \label{fig:GPSVSHAT}
  \end{subfigure}
  \hfill
  \begin{subfigure}[t]{0.48\linewidth}
    \centering
    \includegraphics[height=1.38in]{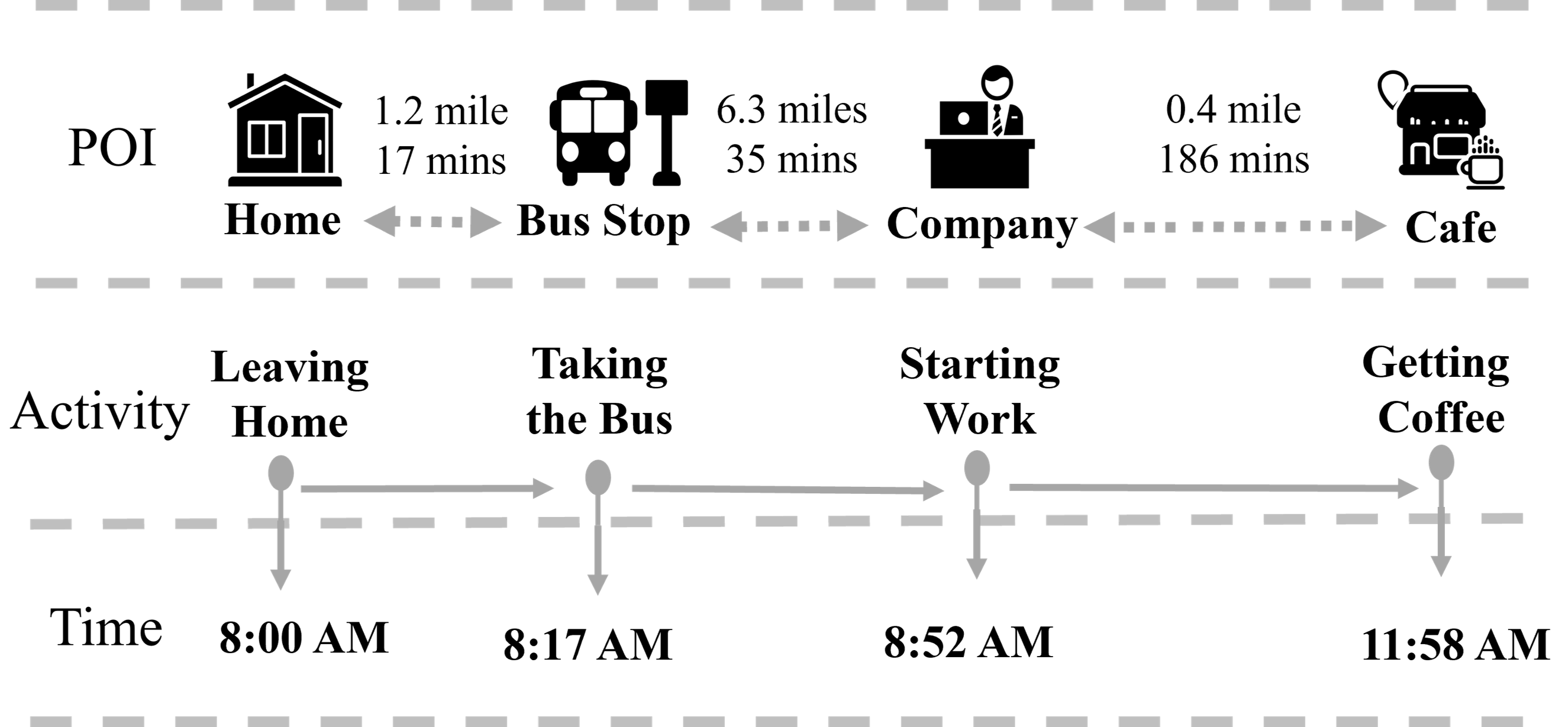}
    \caption{A concrete example of a HAT collected using a smartphone from a user.}
    \label{fig:HATexample}
  \end{subfigure}\vspace{-10pt}
  \caption{Illustrations of the difference between spatially discrete, temporally irregular HATs and spatially continuous, temporally regular GPS trajectories.}
  \label{fig:figure1}
  % \vspace{-15pt}
\end{figure*}
\begin{figure}[h]
\centering
\begin{subfigure}[b]{0.31\linewidth}
  \includegraphics[width=\linewidth]{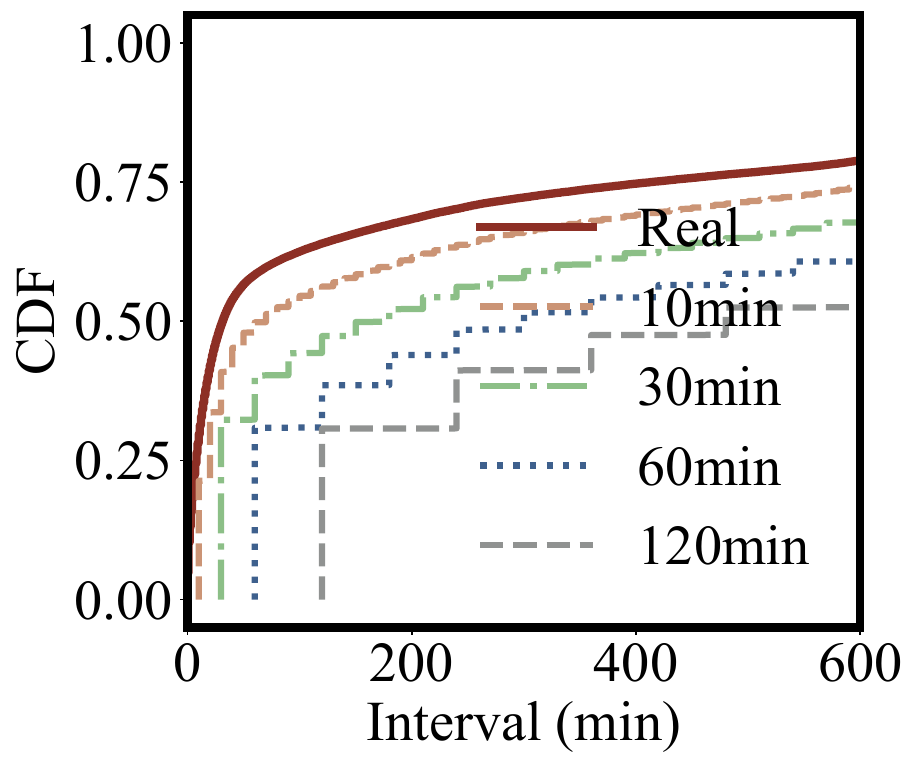}
  \caption{Interval Distribution}
  \label{fig:motivation:b}
\end{subfigure}\hfill
\begin{subfigure}[b]{0.31\linewidth}
  \includegraphics[width=\linewidth]{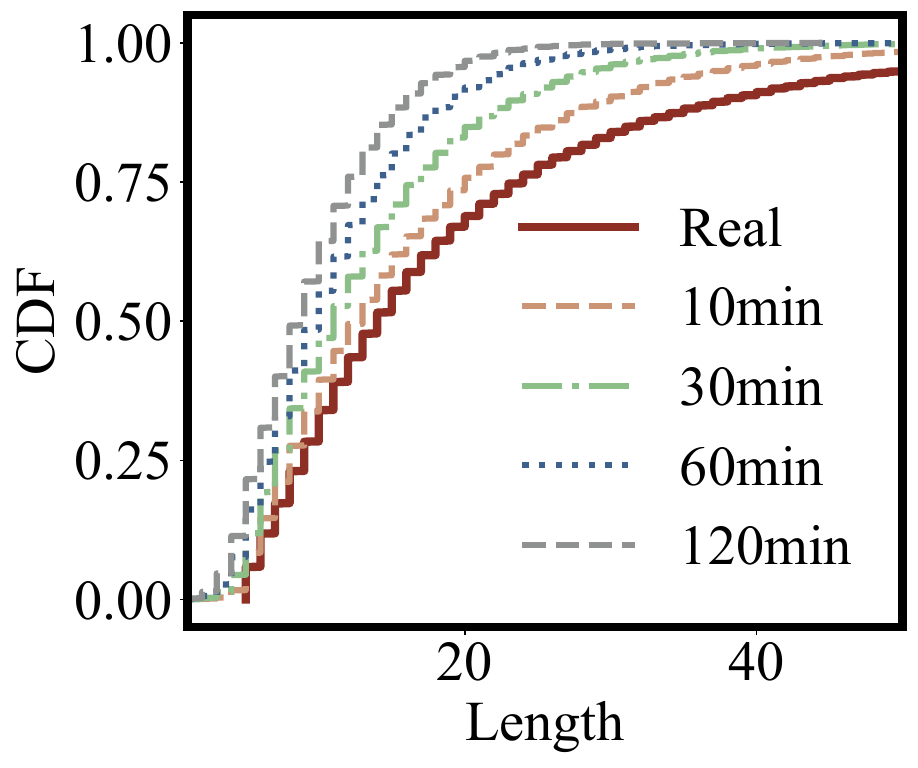}
  \caption{Length Distribution}
  \label{fig:motivation:c}
\end{subfigure}\hfill
\begin{subfigure}[b]{0.35\linewidth}
  \captionsetup{skip=9.3pt}
  \includegraphics[width=\textwidth]{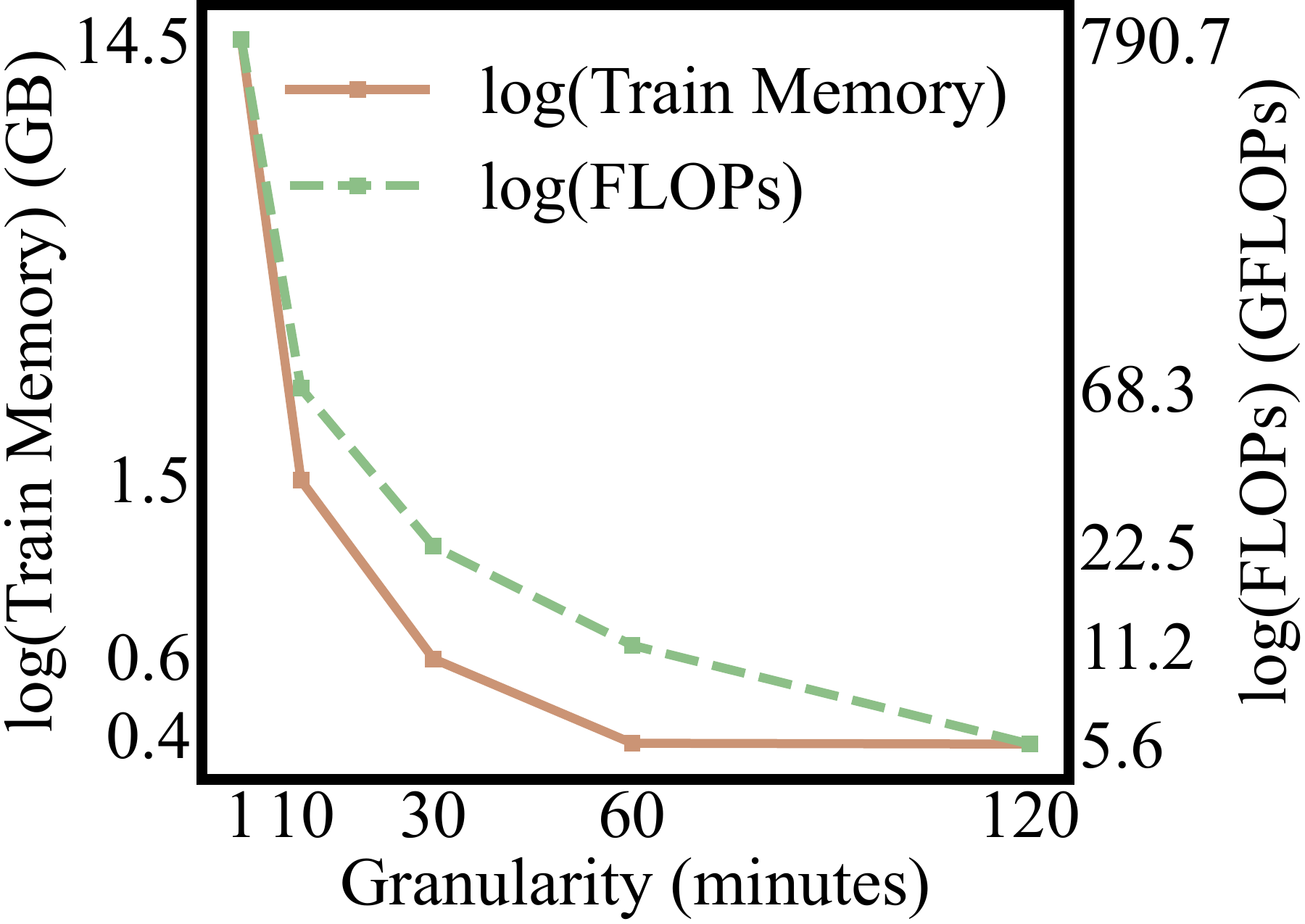}
  \caption{Computation Efficiency}
  \label{fig:motivation:d}
\end{subfigure}

\caption{(a, b) Visualizations of interval and length distributions of real HAT data and interpolated ones at different granularities. (c) Computational costs at different granularities.}
\label{fig:motivation}
% \vspace{-10pt}
\end{figure}

Motivated by the recent success of diffusion models in image~\cite{dhariwal2021diffusion, kulikov2023sinddm} and audio~\cite{huang2022prodiff} generation, researchers have begun exploring their potential for modeling human activities. However, most prior studies have focused on trajectory generation rather than Human Activity Trace (HAT) synthesis. For example, recent works~\cite{zhu2023difftraj, zhu2024controltraj,guo2025leveraging} aim to generate GPS trajectories of a fixed time interval and uniform sequence length (e.g., 100 points per record). In other words, GPS trajectories can be regarded as spatially continuous and temporally regular, where locations are continuously recorded along the path at a fixed time interval. Although these diffusion models have achieved strong performance in trajectory synthesis, extending them to HAT synthesis is non-trivial due to the following \textbf{two challenges}. (i) From the data perspective, HATs are temporally irregular and spatially discrete caused by the inherent uncertainty of human activities, which distinguishes them from GPS trajectories. 
% These irregularities manifest as variable time intervals between consecutive activities and heterogeneous trace lengths across users, often referred to as the spatially discrete and temporally irregular nature of HATs~\cite{bilovs2019uncertainty, jin2023large}. 
An illustrative comparison between HATs and GPS trajectories is shown in Figure~\ref{fig:figure1}. (ii) From the model perspective, adapting diffusion models for HAT synthesis is nontrivial, as these models were originally designed for generating spatially continuous and temporally regular data. As shown in Figures~\ref{fig:motivation:b} and~\ref{fig:motivation:c}, using a fine-grained time interval (e.g., 10 minute) produces HATs that more closely resemble real data (i.e., 1-minute time interval), whereas a coarse-grained interval (e.g., 60 minutes) leads to substantial gaps with authentic data. This means larger time intervals tend to reduce the fidelity of fine-grained HAT generation. However, while fine-grained intervals enhance realism, they also significantly increase computational complexity, training time, and resource consumption, as illustrated in Figure~\ref{fig:motivation:d}. Therefore, a key challenge lies in achieving high-fidelity synthetic HAT generation while maintaining computational efficiency.

 % and a smaller interval (e.g., 1 minute) may reduce the generation efficiency and complexity for long traces. For example, the irregular and dynamic temporal intervals in asynchronous HATs make it difficult to select an appropriate sampling granularity: a larger interval (e.g., 30 minutes) sacrifices temporal precision, whereas a smaller one (e.g., 1 minute) drastically increases generation cost.

To address the above challenges, we propose \N, a computationally efficient two-stage coarse-to-fine framework for realistic HAT synthesis.
% , which balances the synthesis quality and computational efficiency by (i) exploiting the diffusion model’s strong ability to generate regular and continuous data, (ii) skipping inactive time slots to reduce redundant computation.
In the first stage, we first convert the original spatially-discrete, temporally-irregular HATs into coarse-grained spatially-continuous, temporally-regular latent ST traces, which makes it possible to adapt the diffusion model for fine-grained HAT synthesis with a low computational complexity.
We then design a Coarse-grained Human Activity Diffusion model (Coarse-HADiff) to capture the overall spatio-temporal (ST) dependencies of the constructed coarse-grained latent ST traces, which includes a novel Latent Spatio-Temporal UNet (LST-UNet) for denoising through dual Drift–Jitter branches for jointly modeling smooth spatial transitions and temporal variations. 
% This design enables the model to learn stable and global representations of activity patterns, providing robust priors for the subsequent fine-grained data synthesis.
We then transform it to latent ST states by filtering active slots with the temporal mask, which skips the time slots when no activity happens, largely eliminating the redundant computations.
In the second stage, we design a three-step pipeline consisting of Behavior Pattern Extraction (BPEM), Fine-HADiff, and Semantic Alignment to further synthesize fine-grained latent ST states based on the coarse-grained latent ST states from stage one. BPEM first extracts global context from the latent ST states to guide fine-grained generation; Fine-HADiff then performs conditional generation to learn local ST dynamics; and finally, Semantic Alignment efficiently maps each generated spatially-continuous coordinate back to discrete POI-anchored activities, ultimately producing temporally accurate and semantically consistent fine-grained HATs.

In summary, this paper has the following key contributions:
\begin{itemize}
  \item 
  % We extend the diffusion model to accommodate LBSN trace data generation, which is characterized by unfixed time intervals between successive trace points, variable trace lengths, and typically operates in discrete space. 
Conceptually, this is the first computational efficiency-aware HAT synthesis work that aims to generate large-scale long-term HATs to benefit data owners and research communities for different real-world use cases, such as privacy-preserving data publishing and data augmentation.

    \item Technically, we propose \N, an efficient two-stage coarse-to-fine diffusion framework for HAT synthesis. A shared Human Activity Diffusion (HADiff) architecture is proposed and applied in both stages with different inputs. It includes a novel Latent Spatio-Temporal UNet (LST-UNet) for denoising through dual Drift–Jitter branches for jointly modeling smooth spatial transitions and temporal variations. In stage 1, we convert the real HATs into coarse-grained latent ST traces for HADiff training to output the synthetic latent traces as the input of the second stage. In stage 2, we design a three-step pipeline composed of BPEM, Fine-HADiff, and Semantic Alignment to capture more precise spatio-temporal dynamics, enhance temporal fidelity, and produce the final fine-grained synthetic HATs.

  \item Experimentally, we extensively evaluate our \N\ using real-world HAT datasets from four cities in three countries. Experimental results demonstrate that our proposed \N\ significantly outperforms state-of-the-art models on data fidelity, data utility, and computation efficiency with satisfactory privacy-preserving performance.
  For example, our \N\ achieves 52\% and 33\% improvements in fidelity in the spatial and temporal metrics over the best baseline on the TKY dataset, 
highlighting its strong capability to generate realistic, fine-grained HATs. 
The code of this project is available at 
\href{https://github.com/Rongchao98/SynHAT}{\textcolor{blue}{https://github.com/Rongchao98/SynHAT}} and \href{https://huggingface.co/spaces/Rongchao0605/SynHAT}{\textcolor{blue}{https://huggingface.co/spaces/Rongchao0605/SynHAT}}.
\end{itemize}

% \newpage

\section{PRELIMINARIES}

\subsection{Problem Statement} \label{sec:problem statement}
% \rc{(ready)}
%In this paper, we regard POI check-in data as a representative form of HATs. POI (Point of Interest) check-ins are spatiotemporal records capturing when and where individuals visit locations such as restaurants, offices, parks, or transit stations. Each check-in event represents a concrete instance of human behavior at a particular time and place. Consequently, a sequence of such check-ins constitutes a human activity trace, reflecting an individual’s temporal dynamics and mobility patterns across different locations.

% This section is organized into three parts. We first introduce the HAT collection process and conduct a real-world data analysis to illustrate the limitations of existing work and motivate our research. Next, we formally define the HAT synthesis task, which aims to generate fine-grained synthetic traces that preserve the spatio-temporal characteristics and behavioral patterns of real-world data. Finally, we introduce the fundamentals of the diffusion probabilistic model, which serve as the basis of our framework.

%, which are more important for real-world applications such as POI recommendation

\textbf{Definition 1. (Human Activity Trace).}
Human Activity Traces (HATs) are temporally ordered records of human behaviors collected from sensing devices such as smartphones, wearables, GPS trackers, and Wi-Fi systems. Each HAT consists of a sequence of timestamped, POI-anchored activities that capture an individual’s actions in daily life. Since our study focuses on the spatiotemporal characteristics of HAT data, we define each activity as a POI-anchored activity to emphasize its spatial semantics. Specifically, each activity corresponds to a distinct POI, representing a meaningful location visit such as staying at home, arriving at work, or visiting a café. A toy example of a HAT is illustrated in Figure~\ref{fig:HATexample}.

Formally, a HAT for a user is defined as a time-ordered sequence of events $s = [x_1, x_2, \ldots, x_n]$. Each event $x_i=(a_i,t_i)$ records an activity $a_i$ at time $t_i$, where $a_i \in A$ is a POI-anchored activity (e.g., stay at home, a workplace, or a restaurant) drawn from a finite, city-specific set $A$. Each activity $a_i$ is associated with the spatial coordinates of the visited POI, $\mathrm{co}(a_i)=(\mathrm{latitude},\mathrm{longitude})$.
% ; for brevity, we write $\mathrm{co}_i=\mathrm{co}(a_i)$.

% The time intervals between consecutive activities, $t_{i+1} - t_i$, are generally irregular.

\noindent \textbf{Definition 2. (Spatiotemporal Taxonomies).}
In this study, a spatially continuous trajectory means the coordinates of each trajectory point can be at any spatial location, whereas a spatially discrete trace indicates each record can be at only finite locations. Temporal regularity means the time interval between two consecutive records is fixed, while temporal irregular data indicates non-uniform or varying time intervals.

\begin{wrapfigure}{r}{0.25\linewidth}
\vspace{-5pt}
    \centering   \includegraphics[width=0.95\linewidth]{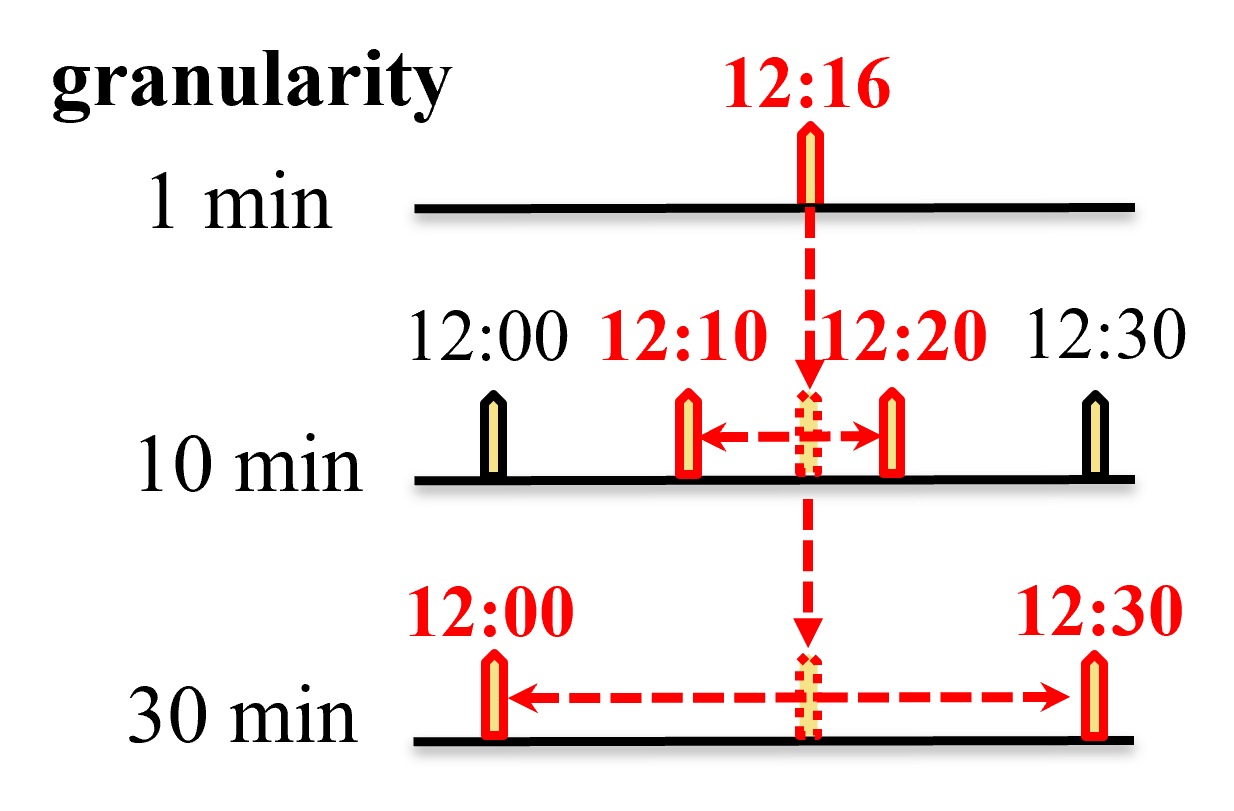}
    \vspace{-5pt}
    \caption{Temporal Granularity 
    % Raw spatial coordinates are discretized into fixed time slots, assigning each coordinate of activity to its corresponding slot. 
    % A binary temporal mask marks slots with at least one activity.
    } 
    \label{fig:granularity}
\end{wrapfigure}

\noindent \textbf{Definition 3. (Temporal Granularity).} The granularity of a HAT is defined as the smallest time unit used to record the trace. Coarse-grained traces employ relatively large time units (e.g., thirty minutes) to record data, whereas fine-grained traces use smaller time units (e.g., one minute) to capture more detailed temporal variations. In this work, we consider one minute-level time unit for fine-grained HATs to enable a more accurate representation of their spatio-temporal patterns. A toy example for different temporal granularity visualization is shown in Figure~\ref{fig:granularity}. 
% For an event occurring at 12:16, a fine-grained temporal granularity of 1 minute can represent it precisely as 12:16, whereas coarse-grained granularity, such as 10-minute or 30-minute intervals, can only describe it as a time range (e.g., 12:10–12:20 or 12:00–12:30).
An event at 12:16 can be represented exactly under a 1-minute temporal granularity, whereas coarser settings (e.g., 10-minute or 30-minute bins) can only localize it to an interval, such as 12:10–12:20 or 12:00–12:30.

\noindent \textbf{Definition 4. (Fine-grained HAT Data Generation).}
Given a real-world dataset containing $m$ fine-grained HATs, denoted as $S = [s_1, s_2, \ldots, s_m]$, the fine-grained HAT data generation task aims to synthesize a new dataset $\hat{S} = [\hat{s}_1, \hat{s}_2, \ldots, \hat{s}_k]$ that preserves the spatio-temporal characteristics and statistical distributions of $S$, while maintaining high utility and protecting data privacy for downstream applications such as next location recommendation.

\subsection{Diffusion Probabilistic Model}
\label{DPM}

In recent years, diffusion probabilistic models, a class of deep generative methods based on deep neural networks, have demonstrated remarkable capabilities in generating diverse types of data such as images and speech~\cite{dhariwal2021diffusion, huang2022prodiff, kulikov2023sinddm}.
Generally, a diffusion model consists of two fundamental processes:
(i) a \textit{forward diffusion process} that gradually corrupts data with Gaussian noise, and
(ii) a \textit{reverse diffusion process} that reconstructs the original data from its noisy version. Let $\mathbf{x}^0 \sim q(\mathbf{x}^0)$ denote a real data sample drawn from the data distribution $q(\mathbf{x}^0)$, and let $p_{\theta}(\mathbf{x}^0)$ be the model distribution that aims to approximate $q(\mathbf{x}^0)$.
Similar to hierarchical variational autoencoders, diffusion models can be formulated as latent variable models in the form
$p_{\theta}(\mathbf{x}^0) := \int p_{\theta}(\mathbf{x}^{0:N}) d\mathbf{x}^{1:N}$,
where $\mathbf{x}^1, \ldots, \mathbf{x}^N$ are latent variables of the same dimension $D$.

\textbf{(i) Forward diffusion process.}
The forward process is a non-trainable Markov chain that gradually adds Gaussian noise to transform $\mathbf{x}^0$ into a sequence of latent variables $\mathbf{x}^1, \ldots, \mathbf{x}^N$, defined as:
\begin{equation}
q(\mathbf{x}^{1:N} | \mathbf{x}^0) = \prod_{n=1}^{N} q(\mathbf{x}^n | \mathbf{x}^{n-1}),
\end{equation}
\begin{equation}
q(\mathbf{x}^n|\mathbf{x}^{n-1}) = \mathcal{N}(\mathbf{x}^n; \sqrt{1 - \beta_n}\mathbf{x}^{n-1}, \beta_n \mathbf{I}),
\end{equation}
where $\beta_1, \ldots, \beta_N \in (0, 1)$ form a fixed variance schedule.
By marginalizing intermediate steps, $\mathbf{x}^n$ can be directly sampled from $\mathbf{x}^0$ via
$q(\mathbf{x}^n|\mathbf{x}^0) = \mathcal{N}(\mathbf{x}^n; \sqrt{\overline{\alpha}_n}\mathbf{x}^0, (1 - \overline{\alpha}_n)\mathbf{I})$,
where $\alpha_n = 1 - \beta_n$ and \textbf{$\overline{\alpha}_n = \prod_{i=1}^{n} \alpha_i
$}.
Using the reparameterization trick, the noisy variable can be expressed as
$\mathbf{x}^n = \sqrt{\overline{\alpha}_n}\mathbf{x}^0 + \sqrt{1 - \overline{\alpha}_n}\boldsymbol{\epsilon}$,
where $\boldsymbol{\epsilon} \sim \mathcal{N}(0, \mathbf{I})$.

\textbf{(ii) The backward diffusion process} is a trainable Markov chain that aims to recover $\mathbf{x}^0$ from $\mathbf{x}^N$, which can be formulated as:
\begin{equation}
\label{formula-3}
    p_{\theta}(\mathbf{x}^{0:N}) = p(\mathbf{x}^N) \prod_{n=1}^{N} p_{\theta}(\mathbf{x}^{n-1}|\mathbf{x}^n), 
\end{equation}
\begin{equation}
\label{formula-4}
    p_{\theta}(\mathbf{x}^{n-1}|\mathbf{x}^n) = \mathcal{N}(\mathbf{x}^{n-1}; \mu_{\theta}(\mathbf{x}^n, n), \sigma_{\theta}(\mathbf{x}^n, n)\mathbf{I}),
\end{equation}
where $\mu_{\theta}(\mathbf{x}^n, n)$ and $\sigma_{\theta}(\mathbf{x}^n, n)$ denote the mean and variance that are normally predicted by a neural network parameterized by $\theta$. The parameter $\theta$ can be optimized by minimizing the negative log-likelihood via a variational lower bound in a format of the KL-divergence between distributions \cite{ho2020denoising}:
% \begin{equation}
%     \min_{\theta} \mathbb{E}_{q(\mathbf{x}^{0:N})} \left[
%     - \log p(\mathbf{x}^N) - \sum_{n=1}^{N} \log \frac{p_{\theta}(\mathbf{x}^{n-1}|\mathbf{x}^n)}{q(\mathbf{x}^n|\mathbf{x}^{n-1})} 
%     \right],
% \end{equation}
\begin{align}
- \log p_{\theta}(\mathbf{x}^0|\mathbf{x}^1) &+ D_{KL}(q(\mathbf{x}^N|\mathbf{x}^0) | p(\mathbf{x}^N)) \nonumber \\
&+ \sum_{n=2}^{N} D_{KL}(q(\mathbf{x}^{n-1}|\mathbf{x}^n, \mathbf{x}^0) | p_{\theta}(\mathbf{x}^{n-1}|\mathbf{x}^n)),
\end{align}
where $q(\mathbf{x}^{n-1}|\mathbf{x}^n, \mathbf{x}^0) = \mathcal{N}(\mathbf{x}^{n-1}; \tilde{\mu}_{\theta}(\mathbf{x}^n, \mathbf{x}^0), \tilde{\beta}_n \mathbf{I})$ and $\tilde{\mu}_{\theta}(\mathbf{x}^n, n) = \frac{1}{\sqrt{\alpha_n}} \left( \mathbf{x}^n - \frac{\beta_n}{\sqrt{1 - \alpha_n}} \epsilon_{\theta}(\mathbf{x}^n, n) \right)$, $\tilde{\beta}_n = \frac{1 - \overline{\alpha}_{n-1}}{1 - \overline{\alpha}_n} \beta_n$. In consequence, the optimization objective can be transformed to:
\begin{equation}
    \mathbb{E}_{\mathbf{x}^0, \boldsymbol{\epsilon}} \left[ 
    \frac{\beta_n^2}{2\Sigma_{\theta}\alpha_n(1 - \overline{\alpha}_n)} \left\| \boldsymbol{\epsilon} - \epsilon_{\theta} \left( \sqrt{\overline{\alpha}_n}\mathbf{x}^0 + \sqrt{1 - \overline{\alpha}_n}\boldsymbol{\epsilon}, n \right) \right\|^2 
    \right],
\label{diff:loss}
\end{equation}
where $\epsilon_{\theta}$ is a neural network for predicting sampled $\boldsymbol{\epsilon} \sim \mathcal{N}(0, \mathbf{I})$. After trained, trace generation is conducted by progressively sampling $\mathbf{x}^{n-1}$ from distribution $p_{\theta}(\mathbf{x}^{n-1}|\mathbf{x}^n)$ until reach $\mathbf{x}^{0}$ by computing:
\begin{equation}
    \mathbf{x}^{n-1} = \frac{1}{\sqrt{\alpha_n}} \left( \mathbf{x}^n - \frac{\beta_n}{\sqrt{1-\alpha_n}}\epsilon_{\theta}(\mathbf{x}^n, n) \right) + \sqrt{\Sigma_{\theta}}\mathbf{z},
\label{diff:infer}
\end{equation}
where \( \mathbf{z} \sim \mathcal{N}(\mathbf{0},\mathbf{I}) \) for \( n = N, \ldots, 2 \) and \( \mathbf{z} = \mathbf{0} \) when \( n = 1 \).

\section{Methodology}
In this paper, we propose \N, an efficient two-stage coarse-to-fine granular human activity trace synthesis framework. Figure~\ref{fig:framework} shows the overall pipeline and key components of \N. In the following part, we will introduce the details of the two stages, respectively.

% , and formalize the diffusion model Coarse-HADIFF and Fine-HADIFF designed in the last section.

\begin{wrapfigure}{r}{0.6\linewidth}
    \centering   \includegraphics[width=1\linewidth]{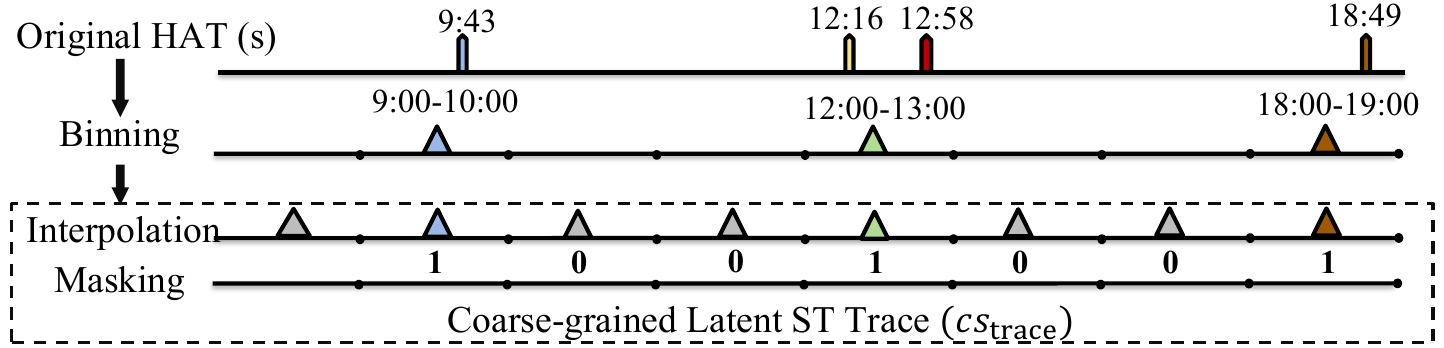}
    \caption{\textbf{Coarse-grained Latent Spatio-Temporal Trace Construction.} 
    } 
    \label{fig:framework31}
\end{wrapfigure}

\begin{figure*}[htpb]
    \centering
\includegraphics[width=\linewidth]{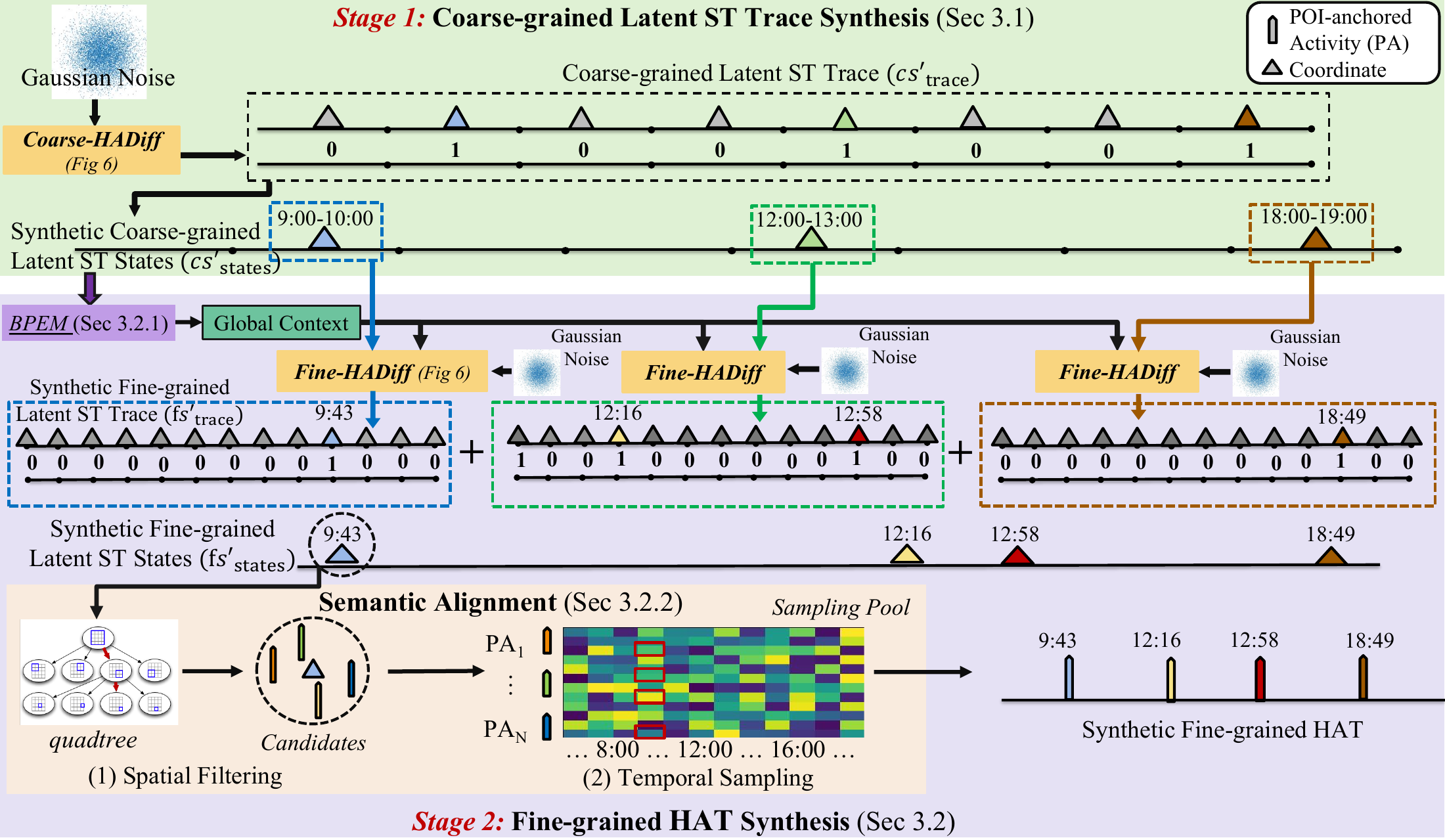}
  \caption{An overall pipeline of the inference process of \N. In the first stage, we input Gaussian noise into the trained Coarse-HADiff in the inference process, and output synthetic coarse-grained latent ST traces. To address the issues of missing activities and misaligned timestamps in coarse-grained traces, the following stage 2 comprises three modules, namely the \textit{Behavior Pattern Extraction Module (BPEM)}, the \textit{Fine-HADiff}, and \textit{Semantic Alignment}, to produce synthetic fine-grained HATs.
  }
  \label{fig:framework}
\end{figure*}

% \vspace{-5pt}
\subsection{Coarse-grained Latent Spatiotemporal Trace Synthesis (Stage 1)}
\label{coarse-grained-trajGen}

% \subsubsection{Coarse-grained Latent Spatio-temporal Trace Formulation}
% We first reconstruct each spatially-discrete, temporally-irregular, fine-grained original HATs into spatially-continuous, temporally-irregular, coarse-grained latent spatio-temporal (ST) trace, which comprises an interpolated trace and its corresponding temporal mask.
\subsubsection{Coarse-grained Latent Spatio-temporal Trace Construction}\label{coarse-grained-Construction}
As shown in Fig.~\ref{fig:framework31}, given a predefined coarse-grained time granularity $Int$ (e.g., 60 minutes) and a duration $D$ (e.g., two weeks), we employ a \textbf{binning} technique to construct a spatially-continuous, temporally-regular, coarse-grained latent spatio-temporal (ST) trace $cs_{trace} = [(l_1, m_1), \ldots, (l_{\frac{D}{Int}}, m_{\frac{D}{Int}})]$ from each original fine-grained, spatially discrete, and temporally irregular HAT $s = [(a_1, t_1), \ldots, (a_{N_s}, t_{N_s})]$ in the training set, where $N_s$ denotes the number of events in HAT $s$. The coarse-grained latent ST trace $cs_{trace}$ comprises an interpolated coordinate sequence $[l_1, \ldots, l_{\frac{D}{Int}}]$ for latent spatial movement modeling and a corresponding temporal mask $[m_1, \ldots, m_{\frac{D}{Int}}]$ for denoting when activities happen.
% Given a predefined coarse-grained time interval $Int$ (e.g., 60 minutes) and a duration $D$ (e.g., two weeks), we employ the \textbf{binning} technique to obtain the trace $s' = [l_1, l_2, \ldots, l_L]$ from the original HAT sample $s = [(a_1, t_1), (a_2, t_2), \ldots, (a_n, t_n)]$.
Each element $l_i$ represents the average location within a specific time slot and is computed as:
\begin{equation}
l_i = \frac{1}{N_i} \sum_{j=1}^{N_i} \mathrm{co}(\hat{a_j}),
\quad \text{where } \hat{a_j} \in A_i= \left\{ a_j \mid t_j \in T_i \right\}, \quad N_i = |A_i|.
\end{equation}
Here, $T_i = [\mathrm{Int}\cdot i,\ \mathrm{Int}\cdot (i+1))$ denotes the $i$-th time slot, and the overall time range is $T=\{T_1, T_2, \ldots, T_\frac{D}{Int}\}$.
If no locations are recorded before the first activity, we create a dummy location $a_s$ to fill the missing positions and ensure temporal continuity. 
In practice, we assign the most frequent POI-anchored activity to $a_s$ and we then linearly interpolate the slots with missing coordinates, yielding spatially smoother, temporally regular sequences that the diffusion model can learn more effectively.
To preserve original temporal information, we introduce a temporal masking to indicate whether any real activity occurs within each time slot of an HAT, as defined below:

\begin{equation}
m_i =
\begin{cases}
1, & \text{if } N_i \ge 1,\\
0, & \text{otherwise}.
\end{cases}
\end{equation}

In this way, we can compress the coarse-grained latent ST trace into meaningful coarse-grained latent ST states $cs_{states}=\left\{ 
    (l_i, T_i) | m_i=1, 0 \leq i \leq L
    \right\}$ by filtering these points with a temporal mask equal to 0. 
% Afterwards, HA-UNet is used to learn the unconditional noise from the latent ST trace, which will output the coarse-grained
% latent ST states.
% The details are in section \ref{diffusionmodel}.

\subsubsection{Coarse-grained Human Activity Diffusion Model (Coarse-HADiff)} \label{HADiff}

% The constructed coarse-grained latent spatio-temporal (ST) trace $cs_{trace}$ and 

% Based on the constructed coarse-grained latent spatio-temporal (ST) trace $cs_{trace}$, we then propose a novel diffusion model called Coarse-HADiff to learn to generate the coarse-grained latent spatio-temporal trace $\hat{s}$ by modeling complex spatio-temporal dependencies in HATs.
% % which is presented in Section~\ref{diffusionmodel}.

Our target next is to 
generate synthetic coarse-grained latent ST traces by learning the complicated behavioral patterns from the constructed coarse-grained latent ST trace to enable HAT synthesis.
Unlike spatially-continuous GPS trajectories, generating latent ST traces is more challenging as \textbf{(i) the temporal mask is highly sparse and has a strong correlation with coordinate sequence}, \textbf{(ii) spatial dynamics after spatial linear interpolation are not as stable as GPS trajectories.} 
To overcome these challenges, we propose a novel \textbf{H}uman \textbf{A}ctivity \textbf{Diff}usion architecture (HADiff) to effectively generate synthetic coarse-grained latent ST traces at different granularities through simulating the underlying distributions.
Moreover, to lower the model complexity and improve the generalizability, we design a flexible denoising network, \textbf{L}atent \textbf{S}patio-\textbf{T}emporal \textbf{UNet} (LST-UNet) within HADiff that can support both unconditional generation and conditional generation at different granularities, which will be adapted to both the two stages for coarse-grained latent ST trace synthesis (Coarse-HADiff) at stage 1 and fine-grained latent ST trace synthesis (Fine-HADiff) at stage 2, respectively.
In this section, we illustrate how Coarse-HADiff generates coarse-grained latent ST traces.

\begin{figure*}[htpb]
    \centering
\includegraphics[width=\linewidth]{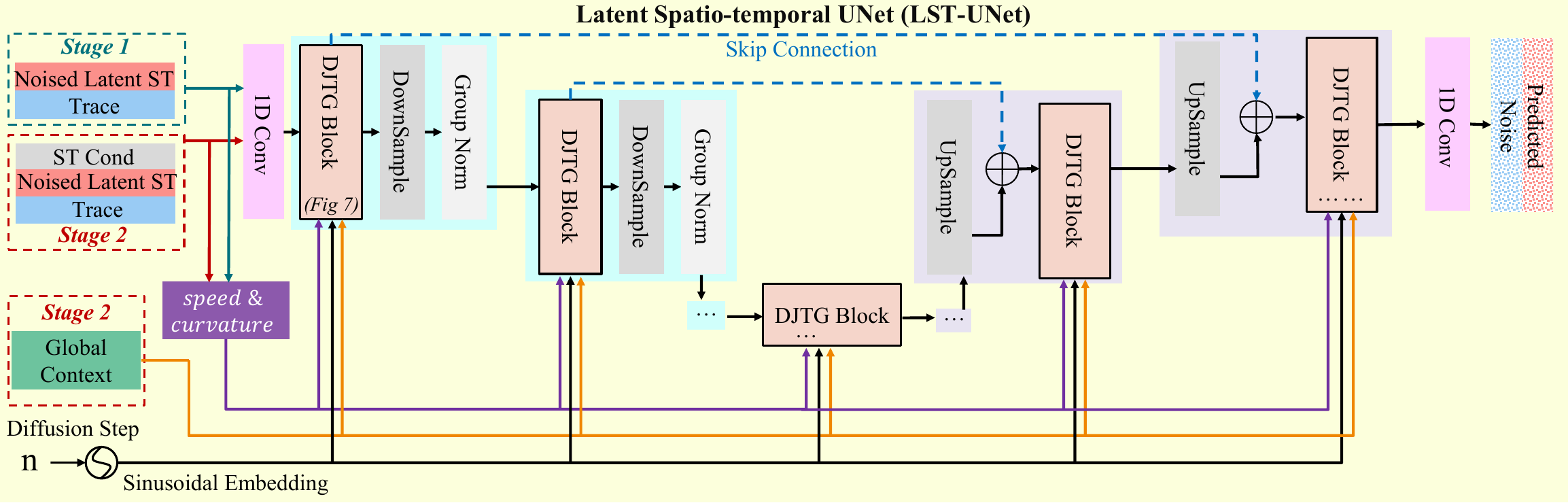}
  \caption{Denoising Network in Coarse-grained HADiff and Fine-grained HADiff of \N. 
  }
  \label{fig:LST-UNet}
\end{figure*}

\begin{figure*}[htpb]
    \centering
\includegraphics[width=\linewidth]{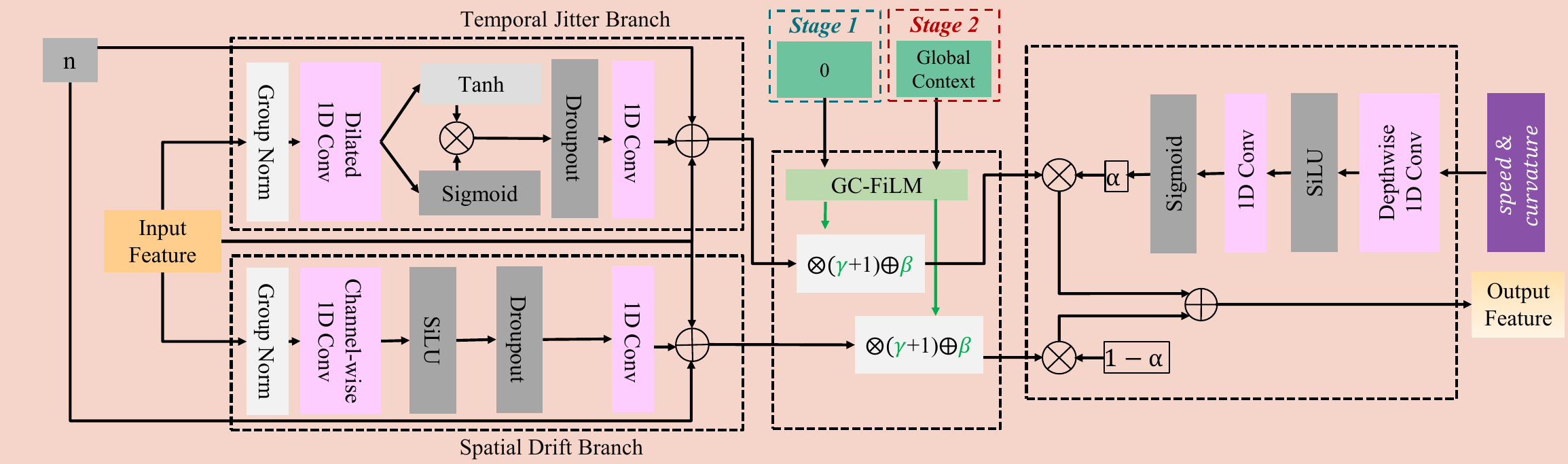}
  \caption{Dual-branch Drift-Jitter TempoGate Block (DJTG Block) in LST-UNet}
  \label{fig:DJTG}
\end{figure*}

The detailed framework of LST-UNet is shown in Figure~\ref{fig:LST-UNet}.
In Coarse-HADiff, LST-UNet takes the noisy coarse-grained latent ST trace $cs_{\text{trace}}^{(n)} \in \mathbb{R}^{\text{Int}\times 3}$ at diffusion step $n$ as input and \emph{unconditionally} predicts the noise added at this step.
We adopt the widely used U-Net as the denoising backbone, where an encoder–decoder with skip connections captures multi-scale patterns while preserving fine details. 
In particular, the architecture comprises resolution-changing stages, i.e., a downsampling encoder that successively halves the spatio-temporal resolution (\(1\times \to \tfrac{1}{2}\times \to \tfrac{1}{4}\times\ \to \dots\)) while widening channels, a bottleneck, and a symmetric upsampling decoder that restores the resolution (\(\dots  \to \tfrac{1}{4}\times \to \tfrac{1}{2}\times \to 1\times\)) via skip connections.
In practice, we adopt three downsampling encoders and upsampling decoders.
At each downsampling and upsampling stage, we propose and insert a dual-branch \textbf{D}rift–\textbf{J}itter \textbf{T}empo\textbf{G}ate blocks (DJTG block) to better model the intrinsic characteristics of latent ST traces and enrich the feature representations.
The detailed framework of the DJTG block is represented in Figure~\ref{fig:DJTG}.
To mitigate sparsity in the temporal mask (challenge (i)), we introduce a \emph{temporal jitter} branch to capture high-frequency variations and noise.
Given an input feature map $ft \in \mathbb{R}^{W \times C}$ (with temporal length $W$ and channels $C$), we first expand the channel capacity via a 1D \emph{dilated} convolution to obtain $M \in \mathbb{R}^{W \times 2C}$.
Splitting along the channel dimension yields $M_1, M_2 \in \mathbb{R}^{W \times C}$ (the first and last $C$ channels of $M$), and a gated activation is applied:
\[
M' \;=\; \tanh(M_1)\;\odot\;\sigma(M_2),
\]
where $\sigma(\cdot)$ is the sigmoid gate that modulates the pass-through. 
This gating efficiently models rapid temporal fluctuations (“jitter”) that complement the coarse trends learned elsewhere.
Alongside the jitter branch for capturing local fluctuations, we introduce a \emph{drift} branch that models smooth, long-term spatial dynamics (challenge (ii)) by incorporating movement-momentum awareness.
We employ depthwise separable 1D convolutions that process each feature channel independently, achieving substantial parameter reduction compared to standard convolutions, and apply SiLU activation to ensure smooth, stable gradient flow during training.
As LST-UNet performs denoising unconditionally at stage 1, the Feature-wise Linear Modulation (details will be introduced in Sec.\ref{FineHADiff}) is skipped by fixing $\gamma, \beta$ to zero.
We further dynamically fuse both branches via learned content-aware weights.
It computes trace motion features—velocity $\|\nabla_t x\|$, curvature $\|\nabla^2_t x\|$, and variance $\text{Var}(ft^{s'}_m)$—and feeds them to a lightweight CNN that predicts timestep-specific fusion weights $\alpha(t)$, yielding $h_{\text{fused}} = \alpha \odot f^{s'}_m + (1-\alpha) \odot f^{t'}_m$, where \(f^{s'}_m, f^{t'}_m\) denote the output features from the Temporal Jitter Branch and Spatial Drift Branch respectively. 
This content-aware mechanism enables automatic weighting between detail-preserving (jitter-dominant) and smoothness-preserving (drift-dominant) processing based on the instantaneous motion characteristics that latent ST trace features.

\subsection{Fine-grained HAT Synthesis (Stage 2)}
\label{fine-grained-trajgen}
There are two main limitations of the generated synthetic coarse-grained latent ST traces at the first stage. (i) States in the original data will be smoothed after reconstruction when multiple activities occur within the same time slot. (ii) The timestamps of points in the generated coarse-grained traces are always integer multiples of the predefined interval \( Int \), e.g., several hours, which introduces significant temporal inaccuracy. 
To address these issues, we design a \textit{Fine-grained Latent ST Trace Learning (FLSTL)} mechanism that generates fine-grained HATs with fine-grained temporal granularity. 
For each activity point \( (l_i, T_i)\) in the coarse-grained latent ST states $cs_{state}$, the FLSTL produces a short-term fine-grained trace segment 
\(fs_{trace} = [(\tilde{l_1}, \tilde{m_1}), \ldots, (\tilde{l_{Int}}, \tilde{m_{Int}})]\) 
within the time slot \( T_i = [\mathrm{Int}\cdot i,\ \mathrm{Int}\cdot (i+1))\). 
All short-term traces are transformed into states and then concatenated to form the complete fine-grained latent ST states $fs_{states}=\left\{ 
    (l_i, t_i) | m_i=1, 0 \leq i \leq \tilde{n}
    \right\},$  where \( \tilde{n} \) denotes the number of states.
% To be more specific, we further perform interpolation to obtain $\tilde{s}$ from $\bar{s}$. 
For each time slot $T_i$, the spatial interpolation range is defined by the head and tail points 
$(l_{i-1},\, \mathrm{Int}\cdot (i - 0.5))$ and $(l_{i+1},\, \mathrm{Int}\cdot (i + 1.5))$, 
respectively, along with the original segments that fall within the corresponding time slot. 
After interpolation, only the segment within the interval $T_i = [\mathrm{Int}\cdot i,\ \mathrm{Int}\cdot (i+1))$ is retained.
The interpolation range is extended to prior and post slots, as we assume that it takes approximately half of the slot duration from the previous activity to the next activity across adjacent slots.

The proposed FLSTL mechanism consists of three main components: 
\textit{(i) Behavior Pattern Extraction Module (BPEM)}, which extracts contextual information from previous activity points to guide the generation of short-term traces by capturing spatio-temporal behavior patterns; \textit{(ii) Fine-HADiff}, which generates short-term traces based on the learned patterns from the BPEM. These traces not only fill in missing points but also include calibrated timestamps. The conditional Fine-HADiff shares the same structure as the unconditional Coarse-HADiff introduced in Sec.~\ref{HADiff}; and \textit{(iii) Semantic Alignment}, which transforms the fine-grained latent ST states to the final fine-grained HATs with spatial filtering and temporal sampling.

\subsubsection{Behavior Pattern Extraction Module (BPEM)}
\label{sec:bpem}
Taking the generated coarse-grained latent ST states from Stage 1 as input, the BPEM extracts global contextual behavior patterns that guide the generation in the following.
% of \( \tilde{s} \), where \(\ddot{n}\) is the length of \(\ddot{c}\).
% HAT \( \ddot{c} = [(a_1, t_1), \ldots, (a_L, t_{\ddot{n}})] \) that formed by concatenating \( [\tilde{s_1}, \dots, \tilde{s_{i-1}}] \) as input, the BPEM extracts global contextual spatio-temporal behavior patterns that guide the generation of \( \tilde{s} \)., where \(\ddot{n}\) is the length of \(\ddot{c}\).
% , which can be expressed as: 
% \begin{equation}
%     G_{context}=\text{Transformer}(\ddot{c}).
% \end{equation}
Initially, each coordinate \( l_i \) is represented by a \( d \)-dimensional spatial embedding \( e_{s_i} \), and each time slot \( T_i \) by a \( d \)-dimensional temporal embedding \( e_{T_i} \).
% For each $a_i$, we define embedding \( e_{s_i} \) as the output of a linear transformation applied to the GPS coordinates \( \mathbf{c}_i \) and the visiting distribution vector \( \mathbf{v}_i \in \mathbb{R}^{L} \) of the activity \(a_i\), 
The embedding \( e_{s_i} \) is obtained via a linear transformation applied to the GPS coordinates \( \mathbf{c}_i \) and the visiting distribution vector \( \mathbf{v}_i \in \mathbb{R}^{L} \), 
where $L$ denotes the number of coarse-grained time steps used to reconstruct the activity trace. The transformation can be mathematically represented as:
\begin{equation}
e_{s_i} = W_c \mathbf{c}_i + W_v \mathbf{v}_i + \mathbf{b}, \quad e_{s_i} \in \mathbb{R}^d
\end{equation}
where \( W_c \in \mathbb{R}^{d \times 2} \) and \( W_v \in \mathbb{R}^{d \times L} \) are the weight matrices for the GPS coordinates and visiting distribution vector respectively, and \( \mathbf{b} \in \mathbb{R}^d \) is a bias vector. 
For each $T_i$, we utilize a positional encoding function $POS(T_i)$ that is commonly used in timestamp embedding to obtain the temporal embedding $e_{T_i}$:
\begin{equation}
\begin{aligned}
e_{T_i, j}= POS(T_i, j) = \left\{\begin{matrix}
\sin\left(\frac{T_i}{10000^{(j - 1)/d}}\right) & \text{if } j \text{ is even}\\ 
 & \\ 
\cos\left(\frac{T_i}{10000^{(j - 1)/d}}\right) & \text{if } j \text{ is odd}
\end{matrix}\right.
\end{aligned}
\end{equation}
\noindent where \( j \) ranges from 1 to \( d \) and \(e_{T_i, j}\) represents the $j$-th value in the \( d \)-dimensional temporal embedding vector \(e_{T_i}\). 

Next, we design a Transformer encoder to process the embeddings \( E = [e_1, e_2, \ldots, e_{\ddot{n}}] \), where each \( e_i \) is formed by combining spatial (\( e_{s_i} \)) and temporal (\( e_{T_i} \)) embeddings, defined as \( e_i = e_{s_i} + e_{T_i} \). This process derives latent representations \( H = [h_1, h_2, \ldots, h_{\ddot{n}}] \), with each \( h_i \) encapsulating the global contextual information of the states in traces.
Through the integration of the attention mechanism, the Transformer encoder captures the essential ST patterns for global states in traces.
% Therefore, the diffusion process can be processed in parallel. 
It is worth mentioning that each diffusion process uses the global context rather than the local context from the previous diffusion, so all the diffusion processes are able to \textbf{execute in parallel} in order to obtain the fine-grained latent ST trace.

\subsubsection{Fine-grained Human Activity Diffusion Model (Fine-HADiff)} \label{FineHADiff}
It is worth noting that though the LST-UNets in the Coarse-HADiff and Fine-HADiff share the same architecture, they are trained separately, and the network parameters are also different.
There are two main differences between the Coarse-HADiff and Fine-HADiff.
Firstly, Fine-HADiff, to generate fine-grained latent ST trace from each coarse state $(l_i, t_i)$, LST-UNet targets to predict the noise with the global context $H$ extracted through BPEM (refer to Sec. \ref{sec:bpem}) in a conditional generation way. 
To adhere to the coarse-grained state, we broadcast $(l_i, t_i)$ to the same size of the noisy fine-grained latent ST trace $fs_{\text{s}}^{(n)}$ and utilize the concatenated tensor of size $\mathbb{R}^{\text{Int}\times 6}$ as the input for denoising.
To enable a single denoiser to support both the presence (Stage 2) and absence (Stage 1) of global context, we adopt Global Context Feature-wise Linear Modulation (GC-FiLM), a lightweight FFN that maps the context vector to per-channel \emph{scale} $\gamma$ and \emph{shift} $\beta$. 
% In the unconditional Stage 1, conditioning is disabled by fixing $(\gamma,\beta)=(0,0)$, 
In the conditional Stage 2, GC-FiLM is activated to generate $(\gamma,\beta)$ from the context.
% In Stage~2 (conditional), C-FiLM is enabled and $(\gamma,\beta)$ are produced from the context; in Stage~1 (unconditional), conditioning is disabled by fixing $(\gamma,\beta)=(0,0)$.
Secondly, to further utilize the global context extracted by BPEM as a condition, we optionally employ GC-FiLM. It performs learned affine transformations on both branches, formulated as $ft^{s'}_m = (1 + \gamma) \odot ft^s_m + \beta$, $ft^{t'}_m = (1 + \gamma) \odot ft^t_m + \beta$ , where $ft^s_m$ and $ft^t_m$ denote the output features from temporal jitter and spatial drift branch, respectively. The transformations enable both branches to adaptively rescale and reshift features based on the diffusion timestep.
This multiplicative conditioning complements the additive approach, providing enhanced expressiveness for complex conditional generation scenarios.

\subsubsection{Semantic Alignment}
Given the generated synthetic fine-grained latent ST states, we propose a semantic alignment with joint spatio-temporal awareness that efficiently maps spatially-continuous, temporally-irregular, fine-grained coordinates to spatially-discrete, temporally-irregular, fine-grained POI-anchored activities.
% discrete activities in the area’s finite set $P$.
First, the quadtree algorithm is applied to efficiently identify nearby activities within a given search radius $r$. 
Given a coordinate $l_i$, the quadtree partitions the spatial region hierarchically, allowing rapid retrieval of candidate activities:
\begin{equation}
    P = \{ a_k \mid \text{dist}(l_i, co(a_k)) \le r, \quad 1\leq k\leq|A| \},
\end{equation}
where $\text{dist}(\cdot)$ denotes the Euclidean distance between two coordinates.

Next, for each candidate activity $a_k \in P$, we compute its temporal visit frequency to capture how often this location is visited over time.
% --- Activity-location inference (HAT-consistent) ---
Let $v_i(t)$ denote the number of visits to activity location $a_i$ at time $t$.
The visit density within a time window $[t-\Delta,\, t+\Delta]$ is defined as
\[
d_i(t) \;=\; \sum_{\tau = t-\Delta}^{t+\Delta} K(t-\tau)\, v_i(\tau),
\]
where $K(\cdot)$ is a temporal kernel (e.g., Gaussian or exponential decay) that assigns higher weights to activities occurring closer to $t$.
After computing temporal densities, we normalize them over the candidate activity set $\mathcal{A}_t$ to obtain a probability distribution:
\[
P(a_i \mid t) \;=\; \frac{d_i(t)}{\sum_{a_j \in A} d_j(t)}.
\]
Finally, we obtain the POI-anchored activity according to $P(a_i \mid t)$ from sampling.
% we select the most likely activity location within the window:
% \[
% a^* \;=\; \arg\max_{a_i \in \mathcal{A}_t} \; P(a_i \mid t).
% \]
This procedure jointly considers spatial proximity (via quadtree-based search) and temporal activity dynamics (via normalized density estimation), yielding more reliable inference of activity locations in HATs.

\subsection{HADiff Training and Inference} \label{training}
In this section, we describe the two-stage training of HADiff and explain how the two models are used during inference.
In stage one, we execute the Coarse-HADiff training following the Eq.\ref{diff:loss} (where $x^0$ is replaced with each original latent ST trace $cs^0_{trace}$ without noise added) and generate a synthetic coarse-grained latent ST trace through auto-regressively recovering a sampled noise \( \epsilon \sim \mathcal{N}(0, I) \) in an unconditional manner according to the following equation:
\begin{align}
    cs^{n-1}_{trace} &= \frac{1}{\sqrt{\alpha_n}} \left({cs}^n_{trace} - \sqrt{1 - \alpha_n} \cdot \epsilon^c_\theta(cs^n_{trace}, n)\right), 
\end{align}
where $n$ is the diffusion step and \( \alpha_n \) are variance-preserving coefficients and $\epsilon^c_\theta(cs^n_{trace}, n)$ denote the estimated noise from the LST-UNet in Coarse-HADiff.
In stage two, we train the conditional LST-UNet $\epsilon^f_\theta$ in Fine-HADiff and BPEM \(e_\xi\) simultaneously to model each fine-grained latent ST trace \({fs}^0_{trace,\hat{t}_i}\) at timestamp \(\hat{t}_i\) in the corresponding latent ST states $cs_{state}=\left\{ (\hat{s}_i, \hat{t}_i)\right\}$ by optimizing: 
\begin{equation}
    \mathbb{E}_{{fs}^0_{trace,\hat{t}_i}, \boldsymbol{\epsilon}} \left[ 
    \frac{\beta_n^2}{2\Sigma_{\theta}\alpha_n(1 - \overline{\alpha}_n)} \left\| \boldsymbol{\epsilon} - \epsilon^f_{\theta} \left( \sqrt{\overline{\alpha}_n} \times{fs}^0_{trace,\hat{t}_i} + \sqrt{1 - \overline{\alpha}_n}\times\boldsymbol{\epsilon}, n, e_\xi(cs^0_{state}), \hat{s}_i, \hat{t}_i \right) \right\|^2 
    \right].
\label{s2diff:loss}
\end{equation}
To improve the robustness and avoid overfitting to seen conditions in the training set and model collapse when taking unseen coarse-grained spatial and temporal conditions $\hat{s}_i, \hat{t}_i$ in the inference, we utilize a spatio-temporal condition perturbation mechanism that adds small random noise to the conditioning signals during training.
With a probability $p_{\text{perturb}}$ during the training, we apply Gaussian perturbations $\hat{s}'_i = \hat{s}_i + \epsilon_s$ and $\hat{t}'_i = \hat{t}_i + \epsilon_t$, where $\epsilon_s, \sim \mathcal{N}(0, \sigma_s^2)$, $\epsilon_t \sim \mathcal{N}(0, \sigma_t^2)$ and $\sigma_s, \sigma_t$ are variance to modulate the spatio-temporal condition deviations. 
We schedule $p_{\text{perturb}}$ during training: it starts small so the model first learns the basic structure, then gradually increases so the model must handle slightly noisy conditions. This encourages a smoother condition-to-trace mapping, reduces memorization of exact condition values, and improves generalization to out-of-distribution spatial and temporal inputs.
After LST-UNet $\epsilon^f_\theta$ and BPEM \(e_\xi\) are trained, given a synthetic latent ST state $cs'_{state}=\left\{ (\hat{s'}_i, \hat{t'}_i)\right\}$ obtained in stage one, we first extract the global context $ct = e_\xi(cs'_{state}) $ once and then generate each synthetic fine-grained latent ST trace auto-regressively as following:
\begin{align}
    fs^{n-1}_{trace,i} &= \frac{1}{\sqrt{\alpha_n}} \left({fs}^n_{trace,i} - \sqrt{1 - \alpha_n} \cdot \epsilon^f_\theta(fs^n_{trace,i}, n, ct, \hat{s'}_i, \hat{t'}_i)\right).
\end{align}

\section{Evaluation}

% In this section, we perform comprehensive experiments to evaluate the performance of our \N from different perspectives, including data fidelity, utility, and privacy. Particularly, we utilize datasets collected from different countries from different sources such as Foursquare and Gowalla. We generate synthetic HATs for different sizes of cities to further verify the generalizability of our design. 

In this section, we first introduce the experimental setup, including
datasets, implementation details, baselines, and evaluation metrics.
We then perform comprehensive experiments to evaluate the performance of our \N\ framework from three dimensions, including 
\textbf{effectiveness}, 
\textbf{robustness}, and \textbf{scalability}. 
Specifically, for effectiveness, we measure the quality of the generated synthetic data from three aspects, i.e.,  \textbf{fidelity}, \textbf{utility}, and \textbf{privacy}, as well as the contributions of each key technical component. For robustness, we evaluate the ability of \N\ to adapt across diverse geographies, dataset scales, and parameter configurations. For scalability, we assess the computational efficiency of \N\ when generating large-scale synthetic HATs.
We aim to address the following research questions:

\begin{itemize}[leftmargin=*]
    \item \textbf{RQ1:} How does the proposed \N\ perform compared with state-of-the-art HAT synthesis baselines in terms of data fidelity from \textit{quantitative} and \textit{qualitative} aspects? Can \N\ generalize to different countries and cities of varying sizes? 
    \item \textbf{RQ2:} Are the synthetic HATs generated by \N\ effective in supporting real-world applications?
    \item \textbf{RQ3:} How effective is \N\ at preserving privacy?
    \item \textbf{RQ4:} What are the contributions of the key technical components within \N\ to the overall performance?   
    \item \textbf{RQ5:} How do parameter choices in \N\ affect overall performance? 
   \item \textbf{RQ6:} Is \N\ computationally efficient to generate large-scale synthetic HATs?

\end{itemize}
\subsection{Experimental Settings}
\subsubsection{Datasets}

To evaluate the effectiveness and generalizability of our proposed \N, we use diverse publicly available datasets including HATs from three different countries (i.e., U.S., Japan, and Sweden) with four different sizes of cities (i.e., New York City (NYC), Tokyo (TKY), Austin (ATX), and Stockholm (STO)). The HATs of NYC, TKY, and ATX are drawn from Foursquare \cite{yang2014modeling,yang2016participatory}, and the STO traces are drawn from Gowalla \cite{cho2011friendship}.
To comprehensively assess HAT generation across time horizons, we set the HAT durations to 7, 14, 28, and 56 days for ATX, NYC, TKY, and STO, respectively.
For preprocessing, we truncate each individual's trace to a fixed time window and retain only traces with more than five visits to ensure data quality.
To reduce temporal drift and keep mobility patterns comparable, we restrict Foursquare traces to April 2012--June 2012 and Gowalla traces to February 2009--October 2010.
Each dataset is randomly split into training, validation, and test sets in a 7:1:2 ratio. Detailed statistics are provided in Table~\ref{tab:dataset_stats}.
Note that, unlike NYC, TKY, and STO, the ATX HAT dataset is much smaller (314 traces vs. thousands), providing a stringent testbed for evaluating the generalizability of \N.

\begin{table}[h]
\caption{Statistics of four datasets. }
\centering
\setlength{\tabcolsep}{4pt}
\begin{tabular*}{0.88\textwidth}{@{\extracolsep{\fill}} lcccc}
\toprule
 & New York City (NYC) & Tokyo (TKY) & Austin (ATX) & Stockholm (STO) \\
\midrule
\# of POIs         & 18207  & 24689  & 859   & 11803  \\
\# of Activities    & 76194  & 143944 & 3828  & 62820  \\
\# of HATs & 4482   & 7576   & 319    & 2094   \\
Duration            & 14 days & 28 days & 7 days & 56 days \\
Average HAT Length & 17   & 19   & 12    & 30   \\
% Unique POIs per HAT & 11   & 12   & 9    & 20   \\
\bottomrule
\end{tabular*}
\label{tab:dataset_stats}
\end{table}

\subsubsection{Baselines}
\label{app:baseline}
We utilize the following HAT synthesis baselines to evaluate the effectiveness of our proposed \N:
\begin{itemize} 
    \item SMM \cite{maglaras2015social}: Semi-Markov process models the interval between two adjacent visits by the exponential distribution. The spatial domain is modeled by the transition matrix and the intensity of the interval. 
    \item  TimeGEO \cite{jiang2016timegeo}: TimeGEO captures the temporal domain by home-based context information and the spatial domain by the explore and preferential return (EPR) model. 
    \item Hawkes \cite{laub2015hawkes}: Hawkes process is a temporal point process model. The partially observed sequence will influence the intensity function of the points to be predicted. 
    \item LSTM \cite{rossi2021vehicle}: LSTM generates a trace by predicting the time interval and location of visits sequentially. 
    \item SeqGAN \cite{yu2017seqgan}: SeqGAN is the state-of-the-art GAN-based discrete sequence generation model. 
    \item MoveSim \cite{feng2020learning}: MoveSim combines the advantages of model-based and model-free methods by introducing the prior knowledge of human mobility regularities into a GAN-based framework. 
    \item DiffTraj \cite{zhu2023difftraj}: DiffTraj is a state-of-the-art diffusion-based model for generating continuous GPS trajectories with a fixed time interval. 
    \item ControlTraj \cite{zhu2024controltraj}: ControlTraj is a diffusion-based framework that generates controllable continuous traces by integrating structural constraints from road network topology. 
    \item Geo\mbox{-}Llama ~\cite{li2025geo}: Geo-Llama is a foundation model that adapts Large Language Models to generate human activity traces by treating locations as a sequence of semantic tokens.
\end{itemize}
SMM, TimeGEO, and Hawkes are model-based methods that utilize physical rules to model spatio-temporal data. In contrast, the remaining methods are learning-based, which employ deep neural networks to learn the complicated patterns of HATs.

% \subsubsection{Implementation Details}
% \label{app:Implementation}
% We implement our \N\ and other
% baselines with XX to guarantee the fairness of performance comparison. \textcolor{red}{Add more details} 

\subsubsection{ Settings and Metrics} \label{sec:metrics}
Following common practice in previous trajectory and trace data synthesis studies~\cite{zhu2023difftraj,feng2020learning}, we evaluate the performance of HAT data synthesis from both spatial and temporal perspectives using a set of task-specific metrics. We comprehensively assess the effectiveness of our method from three perspectives: fidelity, utility, and privacy. The detailed metric definitions are introduced as follows.

\textbf{Fidelity Evaluations:} We collect the following five statistics to evaluate the fidelity of synthetic HATs by comparing the distributions of generated HATs with the original HATs:
\begin{itemize}
    \item \textbf{Distance}: The moving distance between activities in individuals’ traces.
    \item \textbf{Radius}: The radius of gyration is the root mean square distance of all activity locations from the central one, which measures the spatial range.
    \item \textbf{Interval}: The time interval between two consecutive activities in a trace.
    \item \textbf{Length}: It means the length of traces (i.e., the number of activities in each trace).
    % \item G-Rank: The number of visits per location, which is calculated as the visiting frequency of top-100 locations.
    \item \textbf{Average}: The average of the four metrics for an overall comparison.
\end{itemize}
Next, we utilize \textbf{Jensen-Shannon Divergence (JSD)}~\cite{JSD} to measure the similarity between the spatio-temporal distributions of synthetic HATs and real HATs based on the above metrics. For two distributions $p$ and $q$ of the first four metrics, the JSD between them is represented by:
\begin{equation}
    JSD(p,q) = \frac{1}{2} KL\left(p \middle\| \frac{p+q}{2} \right) + \frac{1}{2} KL\left(q \middle\| \frac{p+q}{2} \right),
\end{equation}
where \(KL(\cdot\|\cdot)\) is the Kullback-Leibler divergence \cite{van2014renyi}. A smaller JSD value indicates a higher similarity between the two distributions, corresponding to better fidelity of the generated data. 

\textbf{Utility Evaluations:}
We utilized Spatio-Temporal Gated Network (STGN) \cite{zhao2020go} and Deep Adversarial Model \cite{taymouri2021deep} for the next Points-of-Interest (POIs) recommendation and next activity time prediction, respectively. 
The LSTM-based STGN incorporates two specialized pairs of Time Gates and Distance Gates to explicitly model the non-uniform intervals between events in HAT.  
The model acts as a sequential classifier that outputs a probability distribution over all possible next Points-of-Interest (POIs) in the area based on the hidden state of the gated network.
We define the following five metrics to evaluate the utility of synthetic HATs. They respectively measure the performance of the generated data when applied to two downstream tasks: next POI recommendation and next activity time prediction.
\begin{itemize}
    \item  \textbf{Next POI Recommendation:} 
    This task evaluates the spatial accuracy of the generated data by predicting the next POI that a user will visit. 
    We adopt three widely used ranking-based metrics to measure the recommendation performance:
    \begin{itemize}
        \item \textbf{HR@5 (Hit Ratio@5)} measures whether the true next POI appears in the top-5 predicted list. 
        A higher HR@5 indicates better accuracy for the most relevant recommendations.
        
        \item \textbf{HR@10 (Hit Ratio@10)} measures whether the true next POI appears in the top-10 predicted list.
        
        \item \textbf{MRR (Mean Reciprocal Rank)} computes the average reciprocal rank of the true next POI in the prediction list, capturing how highly the correct POI is ranked on average. A larger MRR corresponds to better ranking quality.
    \end{itemize}

    \item \textbf{Next Activity Time Prediction:}
    This task evaluates the temporal accuracy of the generated data by predicting the occurrence time of the next activity. We adopt two commonly used regression-based metrics to measure the prediction error:
    \begin{itemize}
        \item \textbf{MAE (Mean Absolute Error)} measures the average absolute difference between the predicted and actual activity times. A smaller MAE indicates more accurate overall predictions.
        
        \item \textbf{RMSE (Root Mean Square Error)} calculates the square root of the average squared differences between the predicted and actual activity times. It penalizes larger errors more heavily, thus reflecting the stability of temporal prediction performance.
    \end{itemize}
\end{itemize}

\textbf{Privacy-preserving Evaluations:}
HATs are highly identifiable, as prior work shows that as few as four spatio-temporal points can uniquely re-identify 95\% of individuals in large-scale human mobility activity datasets~\cite{deMontjoye2013unique}, underscoring the risk of memorization in synthetic HAT generation.
Motivated by this risk, we first test whether synthetic HATs inadvertently reveal training data.
To evaluate the privacy-preserving effectiveness of our proposed \N, we test whether synthetic HATs memorize training HAT data by checking how much of each synthetic HAT can be covered from any single training HAT under small spatial and temporal tolerances.

Formally, we define \textbf{similarity} $sim(s_1,s_2)$ to quantify the overlap between a synthetic HAT and real HATs in the training set.
Formally, given two HATs $s_1=[x_1,\ldots,x_{N_1}]$ and $s_2=[x'_1,\ldots,x'_{N_2}]$, 
where each event $x_i=(a_i,t_i)$ denotes a POI-anchored activity $a_i$ and its timestamp $t_i$, 
two events $x_i$ and $x'_j$ are considered a \textit{match} if their spatial and temporal distances 
are within predefined thresholds, i.e., $|co(a_i)-co(a'_j)|\le tr_s$ and $|t_i-t'_j|\le tr_t$, 
where $tr_s$ and $tr_t$ are the spatial and temporal tolerances, respectively. In our privacy-preserving evaluation, we consider both short-range and long-range matching, using (\(tr_s = 0.2\) km, \(tr_t = 30\) minutes) and (\(tr_s = 2\) km, \(tr_t = 2\) hours), respectively, for each city.
The similarity between two HATs is then defined as:
\[
sim(s_1,s_2) = \frac{N_{\text{matched}}}{N_1 + N_2 - N_{\text{matched}}},
\]
where $N_{\text{matched}}$ denotes the number of matched activities between $s_1$ and $s_2$. 
A smaller similarity indicates lower memorization of training samples and thus stronger privacy preservation.

% \begin{itemize}
%     \item Distance: The moving distance between activities in individuals’ traces.
%     \item Radius: The radius of gyration is the root mean square distance of all activity locations from the central one, which measures the spatial range.
%     \item Interval: The time interval between two consecutive activities in a trace.
%     \item Length: It means the length of traces (i.e., the number of activities in each trace).
%     % \item G-Rank: The number of visits per location, which is calculated as the visiting frequency of top-100 locations.
%     \item Average: The average of the four metrics for an overall comparison.
% \end{itemize}
% We utilize Jensen-Shannon Divergence (JSD)~\cite{JSD} to measure the similarity between the spatio-temporal distributions of generated synthetic HATs and real HATs based on the above metrics. For two distributions $p$ and $q$ of the first four metrics, the JSD between them is represented by:
% \begin{equation}
%     JSD(p,q) = \frac{1}{2} KL\left(p \middle\| \frac{p+q}{2} \right) + \frac{1}{2} KL\left(q \middle\| \frac{p+q}{2} \right),
% \end{equation}
% where \(KL(\cdot\|\cdot)\) is the Kullback-Leibler divergence \cite{van2014renyi}. 

% Meanwhile, 

\subsection{RQ1: Fidelity Evaluation}
\begin{table*}[t]
\caption{Fidelity Evaluation. The best results on each dataset are in \textbf{bold}, and the second-best results are \underline{underlined}.}
\centering
\small
\begin{adjustbox}{max width=\textwidth, max totalheight=\textheight, keepaspectratio}
\begin{tabular}{|c|c|c|c|c|c|c|c|c|c|c|}
\hline
\diagbox{Method}{Metric} & Distance & Radius & Interval & Length & Average & Distance & Radius & Interval & Length & Average\\ \hline \hline
\multicolumn{1}{|c|}{} & \multicolumn{5}{c|}{\textbf{TKY}} & \multicolumn{5}{c|}{\textbf{NYC}} \\ \hline
SMM~\cite{maglaras2015social} & 0.178 & 0.277 & 0.217 & 0.329 & 0.250 & 0.196 & 0.259 & 0.381 & 0.358 & 0.298 \\
TimeGEO~\cite{jiang2016timegeo} & 0.277 & 0.524 & 0.232 & 0.216 & 0.312 & 0.251 & 0.625 & 0.278 & 0.386 & 0.385 \\
Hawkes~\cite{laub2015hawkes} & 0.157 & 0.315 & 0.269 & 0.281 & 0.256 & 0.190 & 0.517 & 0.339 & 0.253 & 0.325 \\
LSTM~\cite{rossi2021vehicle} & 0.184 & 0.292 & 0.230 & 0.257 & 0.241 & 0.166 & 0.441 & 0.175 & 0.235 & 0.254 \\
SeqGAN~\cite{yu2017seqgan} & 0.081 & 0.256 & 0.114 & 0.352 & 0.201 & 0.079 & 0.290 & 0.133 & 0.389 & 0.223 \\
MoveSim~\cite{feng2020learning} & 0.101 & 0.287 & 0.212 & 0.275 & 0.219 & 0.105 & 0.352 & 0.161 & 0.303 & 0.230 \\
DiffTraj~\cite{zhu2023difftraj} & 0.277 & 0.201 & 0.117 & 0.744 & 0.335 & \underline{0.052} & \underline{0.085} & 0.104 & 0.321 & 0.140 \\
ControlTraj~\cite{zhu2024controltraj} & \underline{0.050} & \underline{0.097} & 0.191 & 0.223 & 0.140 & 0.193 & 0.356 & 0.160 & 0.337 & 0.262 \\
Geo\mbox{-}Llama~\cite{li2025geo} & 0.260 & 0.246 & \underline{0.006} & \textbf{0.043} & \underline{0.139} & 0.153 & 0.184 & \textbf{0.005} & \underline{0.044} & \underline{0.096} \\
\textbf{SynHAT (ours)} & \textbf{0.032} & \textbf{0.047} & \textbf{0.004} & \underline{0.051} & \textbf{0.034} & \textbf{0.047} & \textbf{0.064} & \underline{0.007} & \textbf{0.041} & \textbf{0.040} \\ \hline \hline

\multicolumn{1}{|c|}{} & \multicolumn{5}{c|}{\textbf{ATX}} & \multicolumn{5}{c|}{\textbf{STO}} \\ \hline
SMM~\cite{maglaras2015social} & 0.603 & 0.625 & 0.139 & 0.347 & 0.428 & 0.637 & 0.599 & 0.697 & 0.428 & 0.590 \\
TimeGEO~\cite{jiang2016timegeo} & 0.416 & 0.501 & 0.148 & 0.401 & 0.366 & 0.579 & 0.489 & 0.727 & 0.397 & 0.548 \\
Hawkes~\cite{laub2015hawkes} & 0.517 & 0.556 & 0.261 & 0.521 & 0.464 & 0.621 & 0.496 & 0.579 & 0.433 & 0.532 \\
LSTM~\cite{rossi2021vehicle} & 0.589 & 0.602 & 0.187 & 0.339 & 0.429 & 0.618 & 0.527 & 0.699 & 0.369 & 0.553 \\
SeqGAN~\cite{yu2017seqgan} & 0.654 & 0.639 & 0.069 & 0.292 & 0.413 & 0.507 & 0.454 & 0.242 & 0.448 & 0.413 \\
MoveSim~\cite{feng2020learning} & 0.529 & 0.549 & 0.179 & 0.297 & 0.389 & 0.397 & 0.431 & 0.127 & \underline{0.218} & 0.293 \\
DiffTraj~\cite{zhu2023difftraj} & \textbf{0.207} & \underline{0.324} & 0.382 & 0.702 & 0.404 & \underline{0.122} & 0.352 & 0.099 & 0.806 & 0.345 \\
ControlTraj~\cite{zhu2024controltraj} & 0.326 & 0.525 & 0.195 & 0.287 & 0.333 & 0.208 & 0.268 & \textbf{0.035} & 0.234 & 0.186 \\
Geo\mbox{-}Llama~\cite{li2025geo} & 0.232 & 0.344 & \underline{0.053} & \underline{0.150} & \underline{0.195} & 0.123 & \underline{0.140} & 0.086 & 0.347 & \underline{0.174} \\
\textbf{SynHAT (ours)} & \underline{0.224} & \textbf{0.251} & \textbf{0.047} & \textbf{0.099} & \textbf{0.155} & \textbf{0.061} & \textbf{0.082} & \underline{0.057} & \textbf{0.159} & \textbf{0.090} \\ \hline
\end{tabular}
\end{adjustbox}
\label{tab:fidelity} 
\end{table*}

\begin{figure*}[t]
\centering

% --- Row 1: FS-TKY Dataset ---
\begin{subfigure}[t]{0.24\textwidth}
    \includegraphics[width=\linewidth]{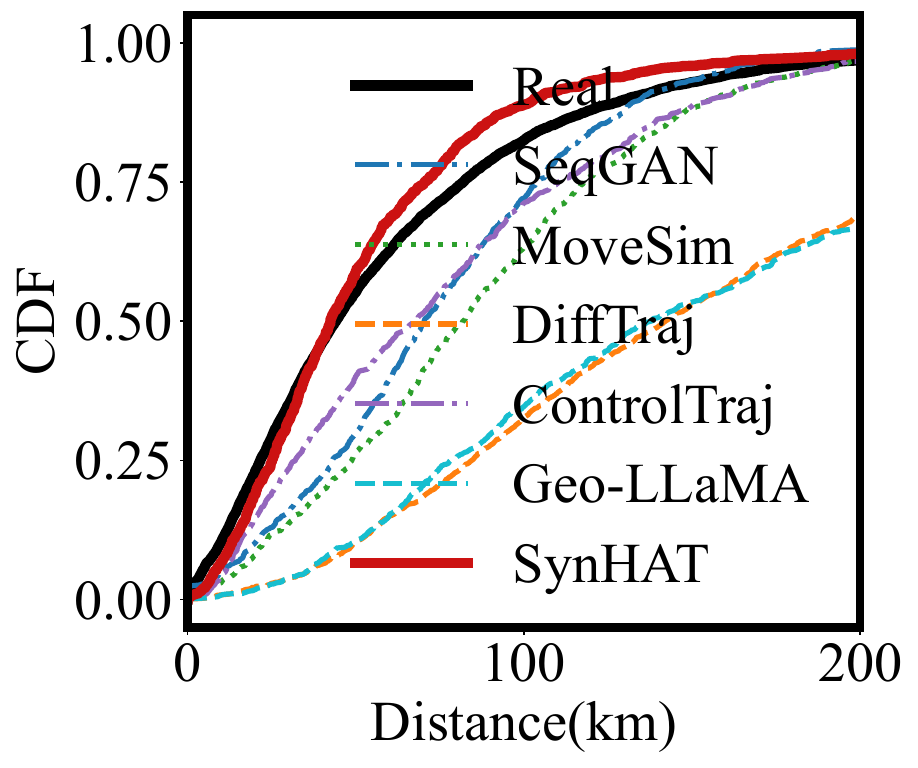}
    \caption{CDF of distance in TKY}
\end{subfigure}
\hfill % Adds horizontal space
\begin{subfigure}[t]{0.24\textwidth}
    \includegraphics[width=\linewidth]{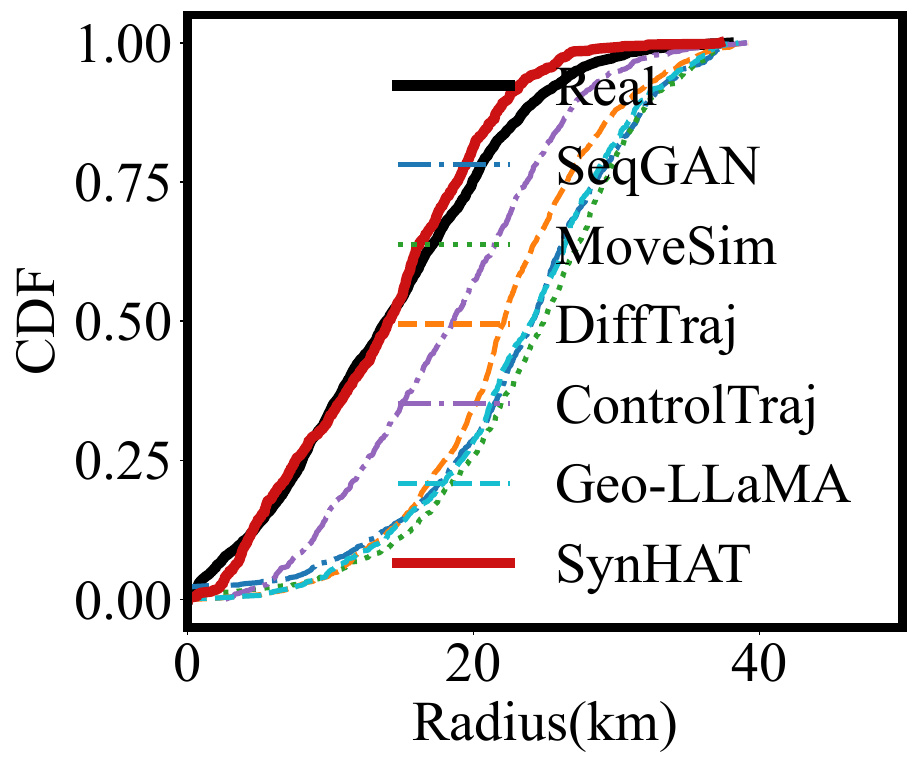}
    \caption{CDF of radius in TKY}
\end{subfigure}
\hfill
\begin{subfigure}[t]{0.24\textwidth}
    \includegraphics[width=\linewidth]{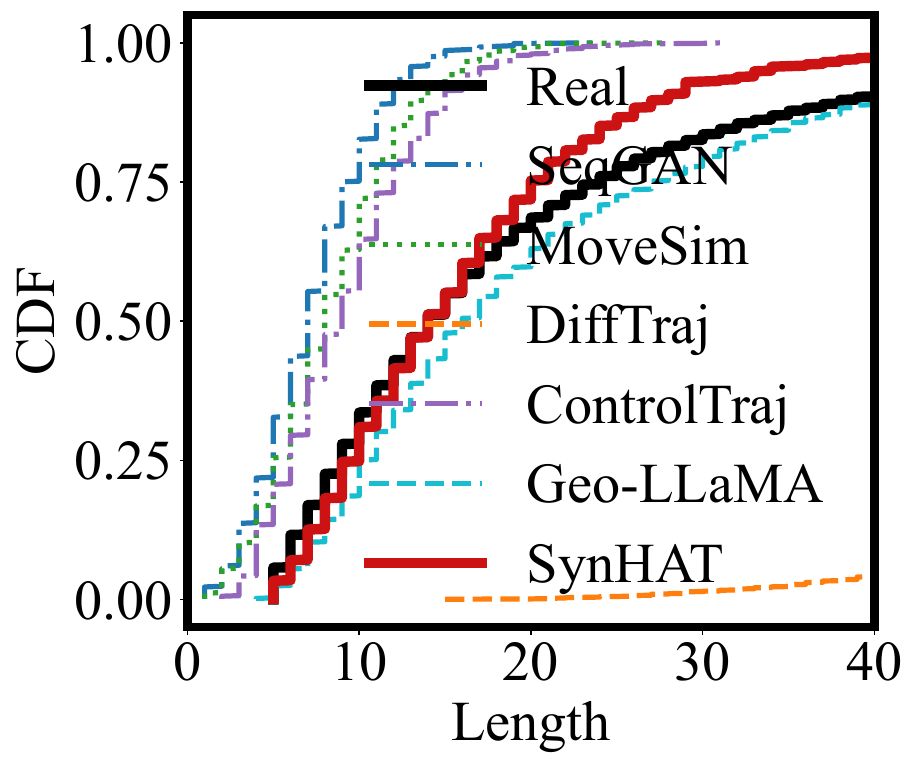}
    \caption{CDF of length in TKY}
\end{subfigure}
\hfill
\begin{subfigure}[t]{0.24\textwidth}
    \includegraphics[width=\linewidth]{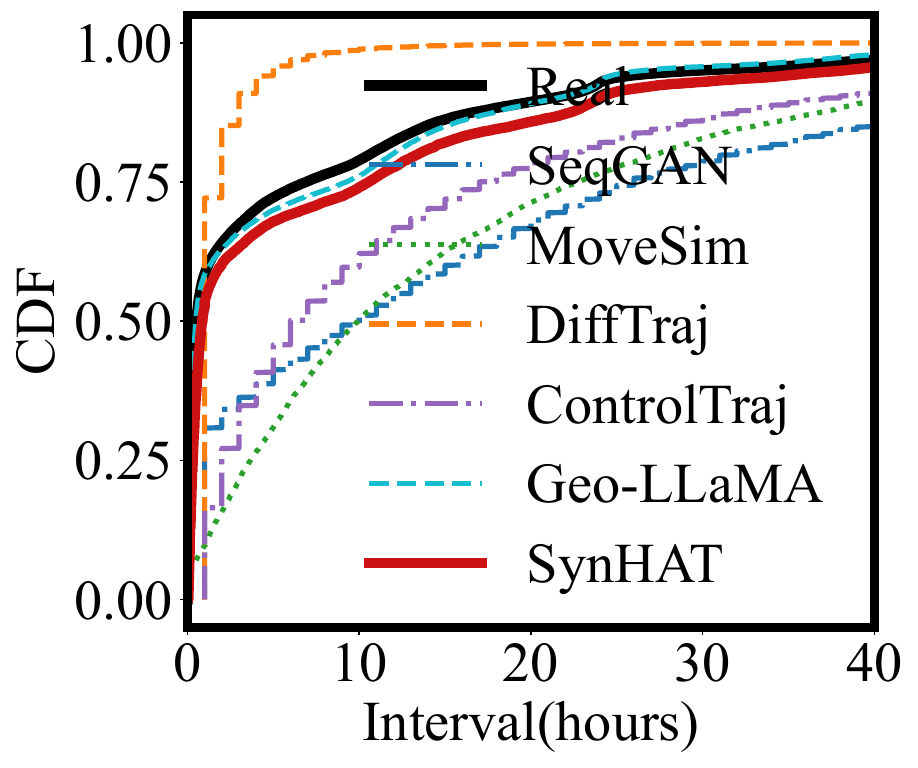}
    \caption{CDF of interval in TKY}
\end{subfigure}

\vspace{1ex} 

% --- Row 1: FS-TKY Dataset ---
\begin{subfigure}[t]{0.24\textwidth}
    \includegraphics[width=\linewidth]{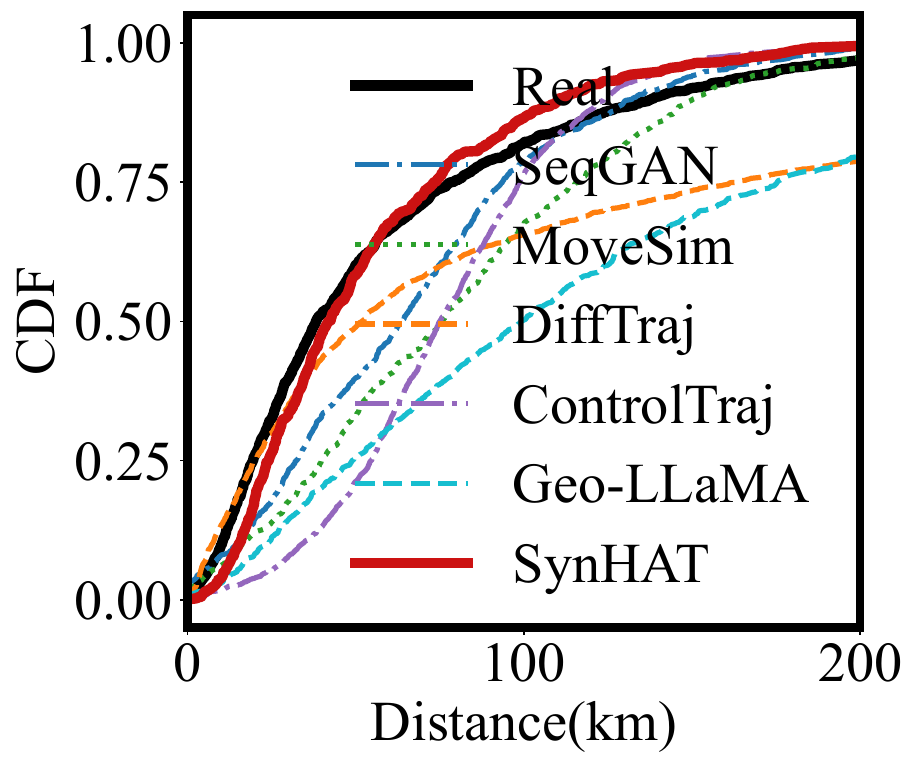}
    \caption{CDF of distance in NYC}
\end{subfigure}
\hfill % Adds horizontal space
\begin{subfigure}[t]{0.24\textwidth}
    \includegraphics[width=\linewidth]{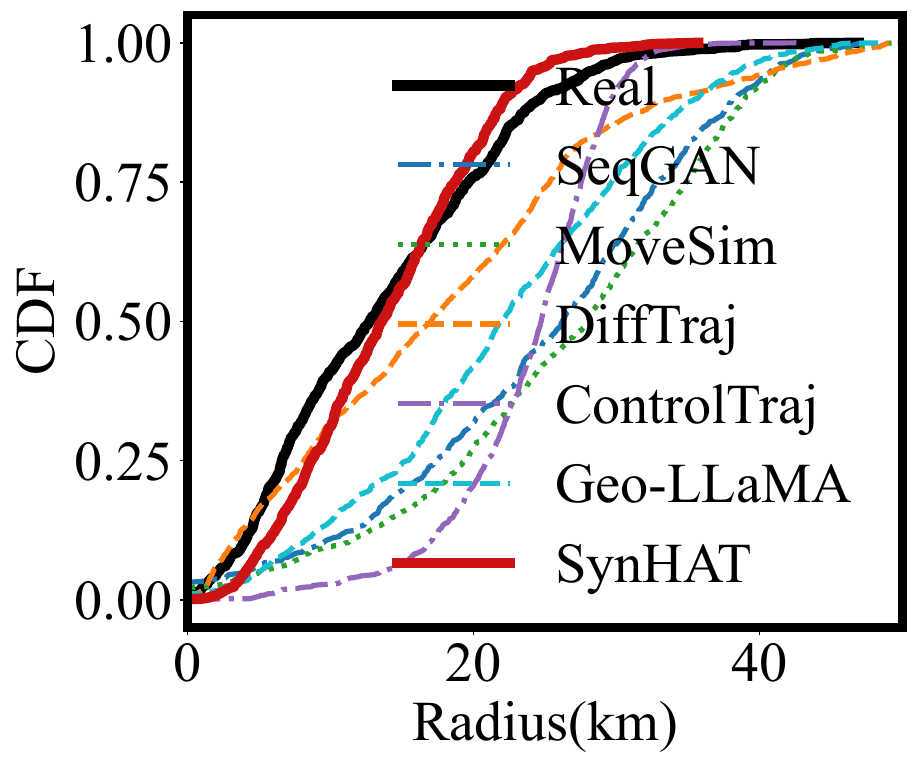}
    \caption{CDF of radius in NYC}
\end{subfigure}
\hfill
\begin{subfigure}[t]{0.24\textwidth}
    \includegraphics[width=\linewidth]{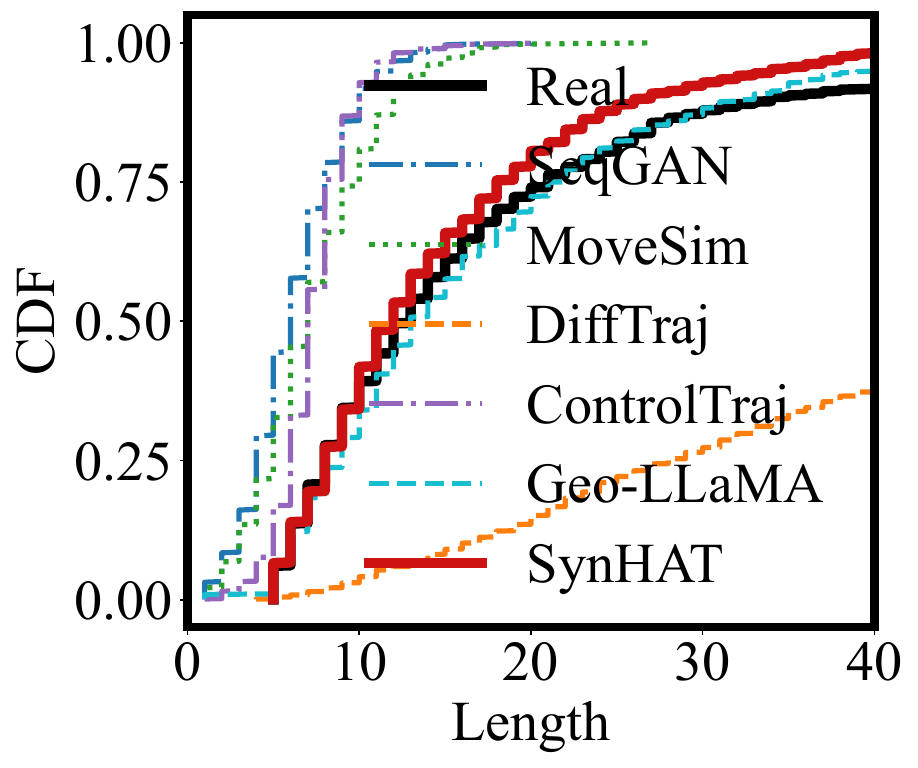}
    \caption{CDF of length in NYC}
\end{subfigure}
\hfill
\begin{subfigure}[t]{0.24\textwidth}
    \includegraphics[width=\linewidth]{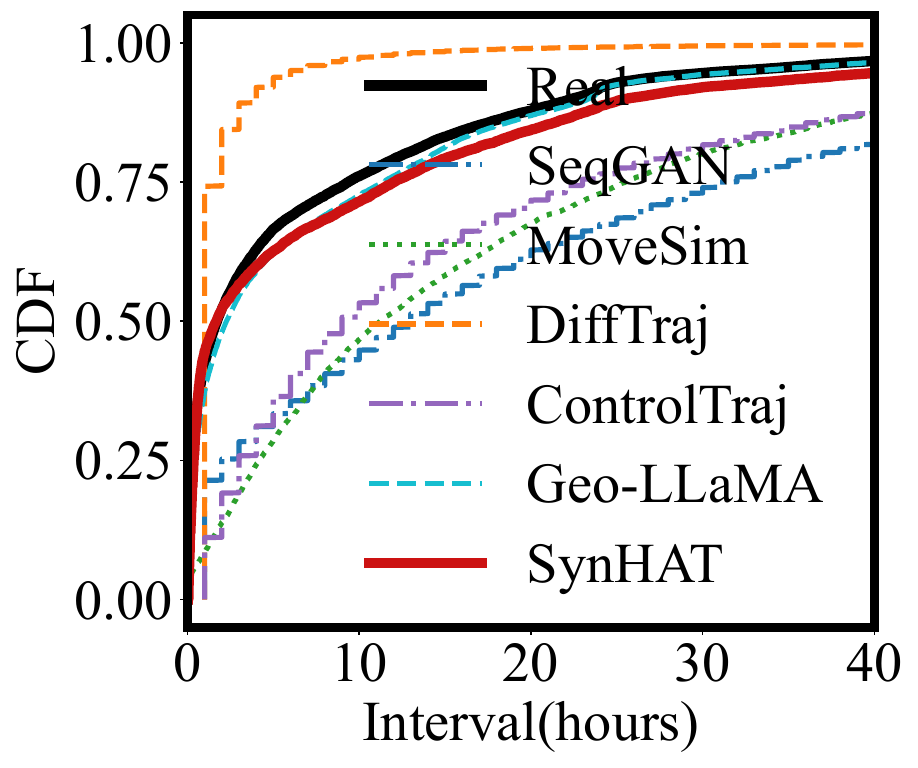}
    \caption{CDF of interval in NYC}
\end{subfigure}

\caption{HAT generation fidelity performance comparison in NYC and TKY.}
\label{fig:fidelity_ext}
% \vspace{-15pt}
\end{figure*}

To evaluate the quality of the generated HATs, we calculate their spatio-temporal pattern distributions using the above-defined five fidelity metrics.
The results are presented in Table~\ref{tab:fidelity}. 
Physics-based methods (i.e., SMM, TimeGEO, and Hawkes) typically underperform on either the spatial or temporal dimension because their hand-crafted statistical/physics assumptions struggle to capture behavioral uncertainty and complex spatio-temporal dependencies.
In contrast, deep learning-based methods demonstrate a more balanced and superior performance compared to traditional model-based approaches. 
Among the four deep learning-based approaches, MoveSim, which utilizes the spatio-temporal characteristics of the original traces, excels above the other three in most metrics. Lacking spatio-temporal context, SeqGAN relies on its proficiency in simulating discrete sequence data, employing the generative adversarial network framework to grasp the nuanced dynamics of mobility sequences. Conversely, adapting DiffTraj, initially designed for continuous GPS trajectories, to generate discrete HATs has led to suboptimal performance in both spatial and temporal dimensions. Additionally, LSTM, which generates traces sequentially, suffers from cumulative errors, resulting in poor performance on generation tasks. 
Our \N\ stands out significantly, demonstrating superb efficacy across most evaluation metrics. For example, our model achieves an improvement exceeding 10\% $(i.e., (0.052 - 0.047) / 0.052)$ and 25\% over the second-best method for the distance and radius metric, respectively, on the NYC dataset, underscoring its advanced capability to model and replicate complex spatio-temporal patterns in spatially discrete, temporally irregular HATs. 

For a more detailed performance evaluation, we utilized Cumulative Distribution Function (CDF) curves to illustrate the distributions of various metrics in NYC and TKY, as shown in Fig.~\ref{fig:fidelity_ext}. Across Fig.~\ref{fig:fidelity_ext} (a) to (h), the CDFs associated with \N\ closely align with those of the original data. This alignment suggests that \N\ is capable of replicating the spatio-temporal patterns found in the actual data with high fidelity.
% To provide an intuitive visualization of these quantitative results, Fig.~\ref{fig:utility_radar} presents radar plots of the five fidelity metrics (i.e., Distance, Radius, Interval, Length, and Average) across four cities. Each metric is normalized within its respective city using inverted min–max scaling, ensuring that \textit{\textbf{higher values indicate smaller JSDs}}. Specifically, this normalization maps the minimum error to $1$ and the maximum error to $0$, effectively inverting the scale so that larger values indicate better performance. As shown in the figure, \N\ consistently forms the largest and most outward-spanning polygons in all four cities, highlighting its robust and balanced capability to capture both spatial and temporal characteristics of HATs. In contrast, other models such as Geo-Llama or DiffTraj exhibit strong performance in specific aspects (e.g., interval or length) but fall short in other metrics, resulting in uneven polygon shapes. These results further confirm that \N\ achieves the highest fidelity in reproducing complex spatio-temporal activity patterns.

\subsection{RQ2: Utility Evaluation (Downstream Applications)} \label{sec:Utility}
We further assess the utility, which reflects the effectiveness of the generated data when applied to downstream tasks.
We consider two evaluation settings and utilize the downstream models and metrics as illustrated in Section~\ref{sec:metrics}.
The first is data publishing, where models are trained entirely on 100\% synthetic data and evaluated on real test data. The second is data augmentation, where models are trained on a mixed dataset consisting of 50\% real data and 50\% synthetic data, and likewise evaluated on real data.
\begin{table*}[t]
\caption{Use Case 1: \textbf{Data Publishing} (100\% synthesis HATs for Training).}% Columns: $^{\mathrm{A}}$ = actvity-level; $^{\mathrm{T}}$ = temporal-level.}
\centering
\small
\renewcommand{\arraystretch}{1.08}
\setlength{\tabcolsep}{4.5pt}
\begin{adjustbox}{max width=\textwidth, keepaspectratio}
\begin{tabular}{|l|ccccc|ccccc|}
\hline
\diagbox[dir=NW]{Training Set}{Metric} &
\multicolumn{5}{c|}{\textbf{NYC}} & \multicolumn{5}{c|}{\textbf{ATX}} \\ \cline{2-11}
& HR@5 & HR@10 & MRR & MAE & RMSE
& HR@5 & HR@10 & MRR & MAE & RMSE \\ \hline \hline
Real  & 0.463 & 0.511 & 0.375 & 0.202 & 0.254 & 0.325 & 0.335 & 0.333 & 1.147 & 1.244 \\ \hline
SeqGAN                      & 0.371 & 0.444 & 0.298 & \underline{0.212} & \underline{0.255} & \textbf{0.291} & \underline{0.301} & \underline{0.308} & 1.230 & 1.451 \\
MoveSim                     & \underline{0.381} & \underline{0.488} & \underline{0.307} & 0.217 & 0.258 & 0.256 & 0.277 & 0.290 & \underline{1.188} & \underline{1.230} \\
DiffTraj                    & 0.066 & 0.077 & 0.056 & 0.879 & 0.928 & 0.048 & 0.065 & 0.054 & 3.211 & 3.679 \\
ControlTraj                 & 0.128 & 0.134 & 0.102 & 0.525 & 0.590 & 0.066 & 0.099 & 0.091 & 2.781 & 2.995 \\ \hline
\textbf{SynHAT (ours)}      & \textbf{0.408} & \textbf{0.508} & \textbf{0.315} & \textbf{0.191} & \textbf{0.233} & \underline{0.288} & \textbf{0.302} & \textbf{0.314} & \textbf{1.150} & \textbf{1.159} \\ \hline
\end{tabular}
\end{adjustbox}
\label{tab:utility_publishing}
\vspace{-8pt}
\end{table*}

% ============== TABLE 2: 50/50 — Data Augmentation =====================
\begin{table*}[t]
\caption{Use Case 2: \textbf{Data Augmentation} (50\% real HATs and 50\% synthesis HATs for Training) }% Columns: $^{\mathrm{A}}$ = actvity-level; $^{\mathrm{T}}$ = temporal-level.}
\centering
\small
\renewcommand{\arraystretch}{1.08}
\setlength{\tabcolsep}{4.5pt}
\begin{adjustbox}{max width=\textwidth, keepaspectratio}
\begin{tabular}{|l|ccccc|ccccc|}
\hline
\diagbox[dir=NW]{Training Set}{Metric} &
\multicolumn{5}{c|}{\textbf{NYC}} & \multicolumn{5}{c|}{\textbf{ATX}} \\ \cline{2-11}
& HR@5 & HR@10 & MRR & MAE & RMSE
& HR@5 & HR@10 & MRR & MAE & RMSE \\ \hline \hline
Real                        & 0.463 & 0.511 & 0.375 & 0.202 & 0.254 & 0.325 & 0.335 & 0.333 & 1.147 & 1.244 \\ \hline
SeqGAN                      & 0.461 & 0.501 & 0.346 & \underline{0.199} & \underline{0.259} & 0.330 & 0.336 & 0.351 & 1.114 & \underline{1.187} \\
MoveSim                     & \underline{0.468} & \textbf{0.538} & \underline{0.383} & 0.214 & 0.260 & \underline{0.333} & \underline{0.348} & \underline{0.355} & \underline{1.101} & 1.771 \\
DiffTraj                    & 0.327 & 0.421 & 0.211 & 0.429 & 0.493 & 0.167 & 0.181 & 0.200 & 1.277 & 1.420 \\
ControlTraj                 & 0.388 & 0.488 & 0.302 & 0.411 & 0.459 & 0.201 & 0.213 & 0.274 & 1.312 & 1.249 \\ \hline \hline
\textbf{SynHAT (ours)}      & \textbf{0.469} & \underline{0.526} & \textbf{0.385} & \textbf{0.195} & \textbf{0.245} & \textbf{0.361} & \textbf{0.377} & \textbf{0.365} & \textbf{0.977} & \textbf{1.004} \\ \hline
\end{tabular}
\end{adjustbox}
\label{tab:utility_augmentation}
% \vspace{-8pt}
\end{table*}

In Table~\ref{tab:utility_publishing}, we present Use Case One for Data Publishing. The results show that \N\ achieves performance close to that of the real dataset, demonstrating notably superior results on both next POI recommendation and next activity time prediction tasks (e.g., the highest HR@5 = 0.408 and the lowest MAE = 0.191 on the NYC dataset) compared with the baselines. 
This improvement can be attributed to \N's ability to leverage the strong generative power of the diffusion model and effectively capture the complex characteristics of real-world HATs. 
Meanwhile, \N's semantic consistency---e.g., its activity-transition patterns (Appendix Sec.~\ref{ext_fid})---is also crucial for downstream spatio-temporal prediction models.
Among the baseline methods, MoveSim ranks second in most cases because it considers spatial factors such as the physical distance between POIs when modeling user movements. In contrast, SeqGAN and DiffTraj perform relatively poorly, as SeqGAN ignores spatial semantics by treating POIs as categorical identifiers, while DiffTraj lacks sufficient ability to capture spatio-temporal dependencies between transitions.

In Table~\ref{tab:utility_augmentation}, we present Use Case Two for Data Augmentation. It is observed that \N\ also provides improvements, surpassing almost all baseline methods and even enhancing downstream model performance compared to using only real data in all the ten cases. Specifically, \N\ achieves the best performance in next POI recommendation (e.g., HR@5 = 0.469 on NYC and 0.361 on ATX) while obtaining the lowest errors in next activity time prediction (MAE = 0.195 on NYC and 0.977 on ATX). These results demonstrate that \N\ is not only realistic but also complementary to real-world data, effectively enriching the data distribution and improving the generalization capability of downstream models. These findings highlight the strong potential of \N\ as an effective data augmentation strategy, particularly for small-scale datasets where limited data hinders model generalization.

\subsection{RQ3: Privacy-Preserving Performance}
We follow the experimental settings for privacy-preserving illustrated in Sec.~\ref{sec:metrics}.
We randomly sample \(1{,}500\) synthetic HATs from each method trained on the same original HAT dataset \(S\). For each synthetic HAT $s_1$, we record its maximum similarity to any single original HAT $s_2$ in \(S\).

\begin{figure*}[t]
\centering
\begin{subfigure}[t]{0.24\textwidth}
    \includegraphics[width=\linewidth]{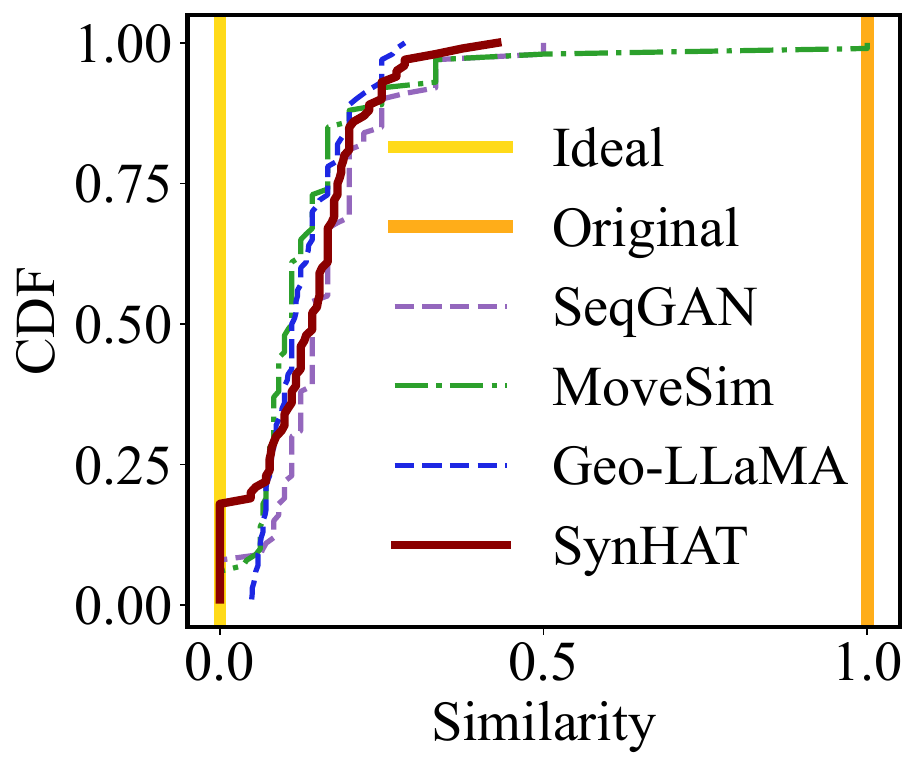}
    \caption{TKY ($tr_s$ = .2km, $tr_t$ = .5h)}
\end{subfigure}
\hfill
\begin{subfigure}[t]{0.24\textwidth}
    \includegraphics[width=\linewidth]{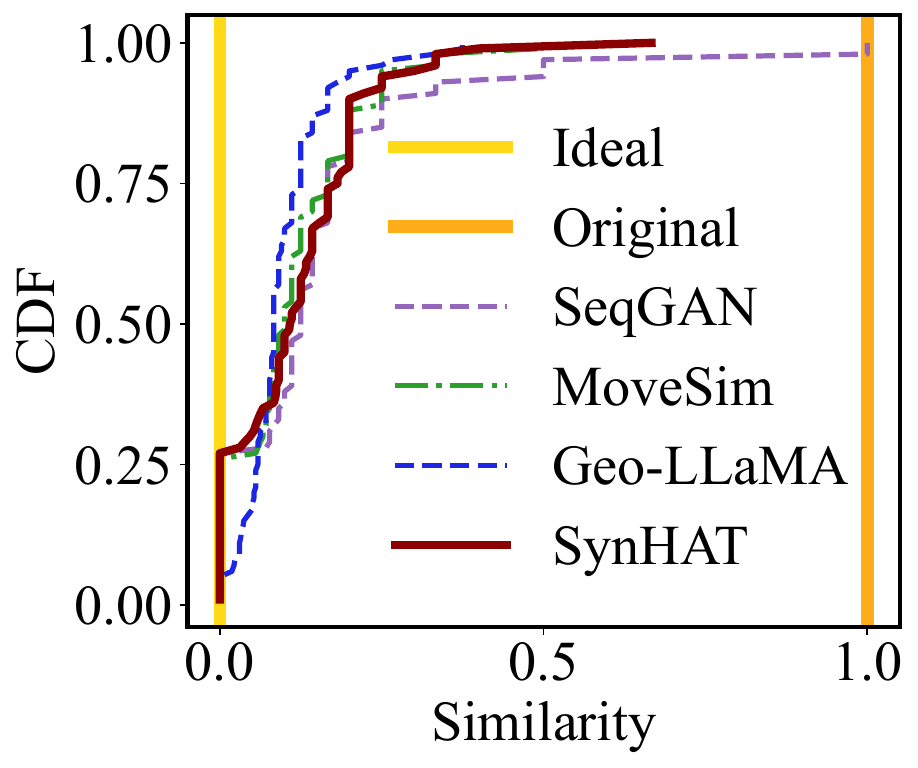}
    \caption{NYC ($tr_s$ = .2km, $tr_t$ = .5h)}
\end{subfigure}
\hfill
\begin{subfigure}[t]{0.24\textwidth}
    \includegraphics[width=\linewidth]{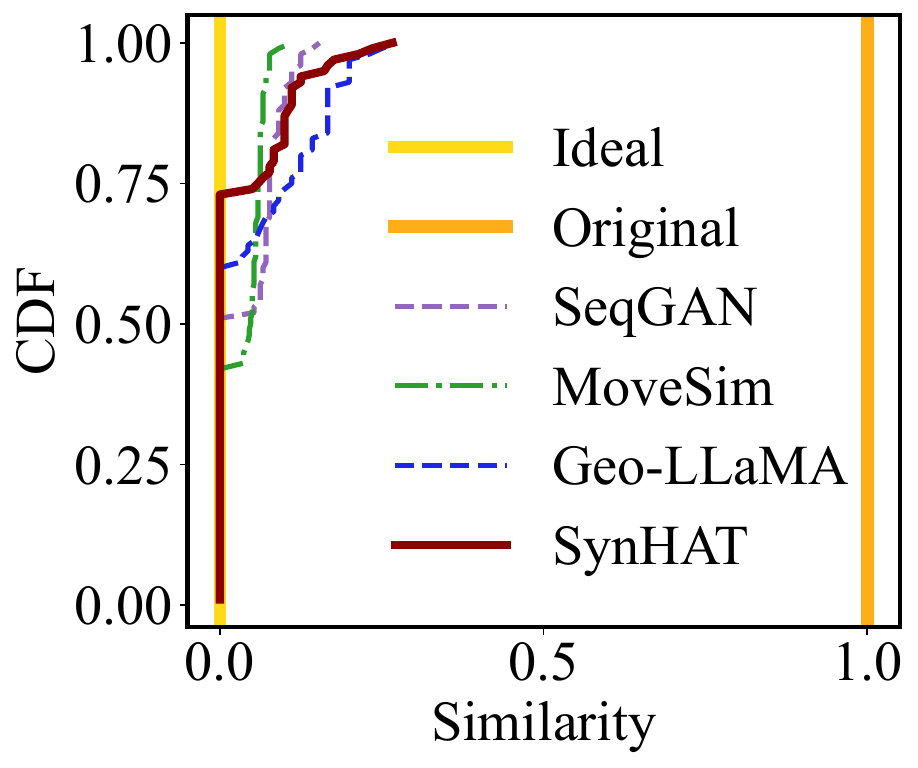}
    \caption{ATX ($tr_s$ = .2km, $tr_t$ = .5h)}
\end{subfigure}
\hfill
\begin{subfigure}[t]{0.24\textwidth}
    \includegraphics[width=\linewidth]
    {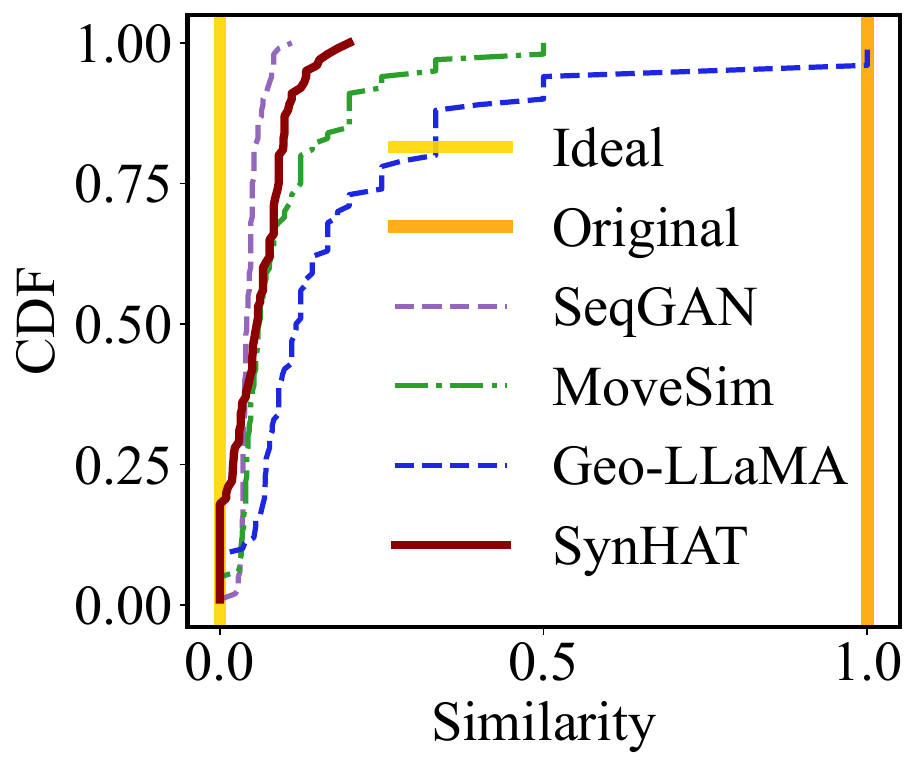}
    \caption{STO ($tr_s$ = .2km, $tr_t$ = .5h)}
\end{subfigure}
% \vspace{1ex}

\begin{subfigure}[t]{0.24\textwidth}
    \includegraphics[width=\linewidth]{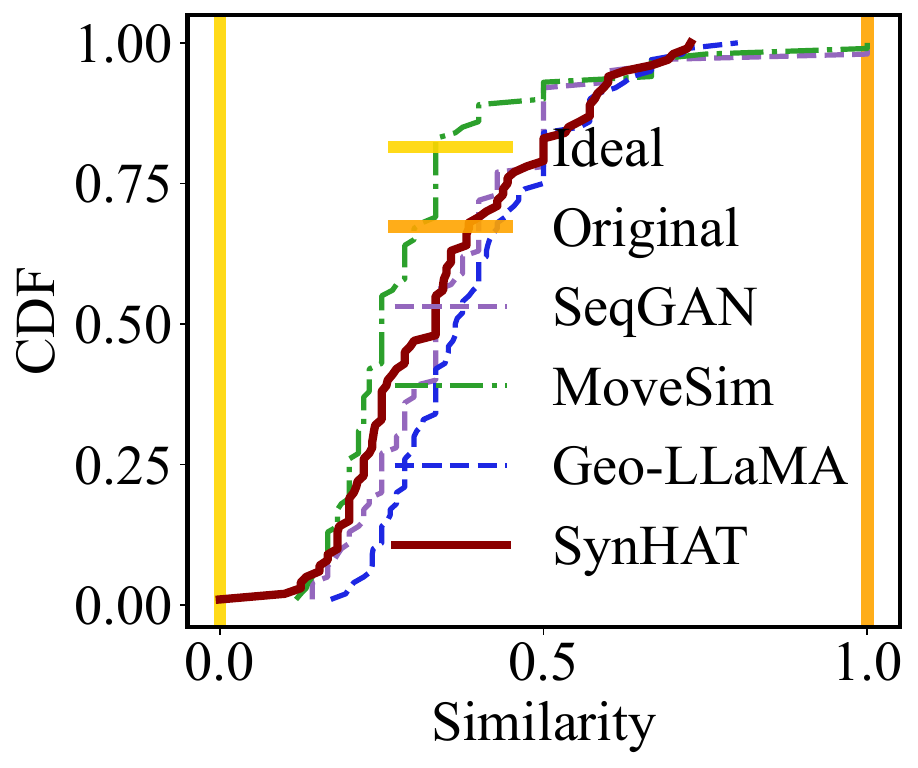}
    \caption{TKY ($tr_s$ = 2km, $tr_t$ = 2h)}
\end{subfigure}
\hfill
\begin{subfigure}[t]{0.24\textwidth}
    \includegraphics[width=\linewidth]{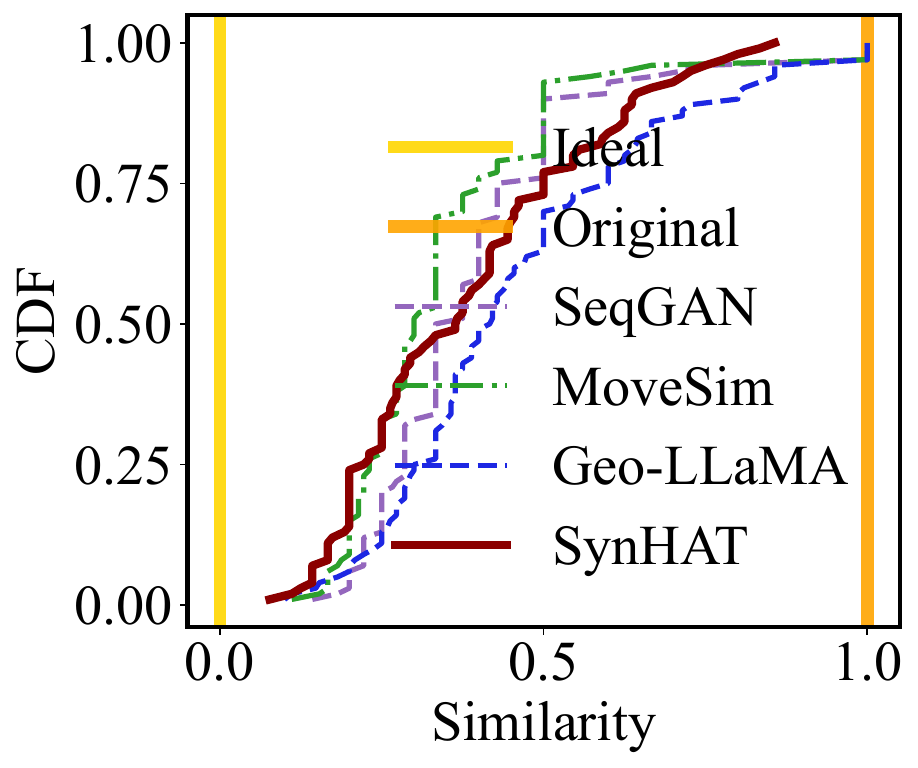}
    \caption{NYC ($tr_s$ = 2km, $tr_t$ = 2h)}
\end{subfigure}
\hfill
\begin{subfigure}[t]{0.24\textwidth}
    \includegraphics[width=\linewidth]{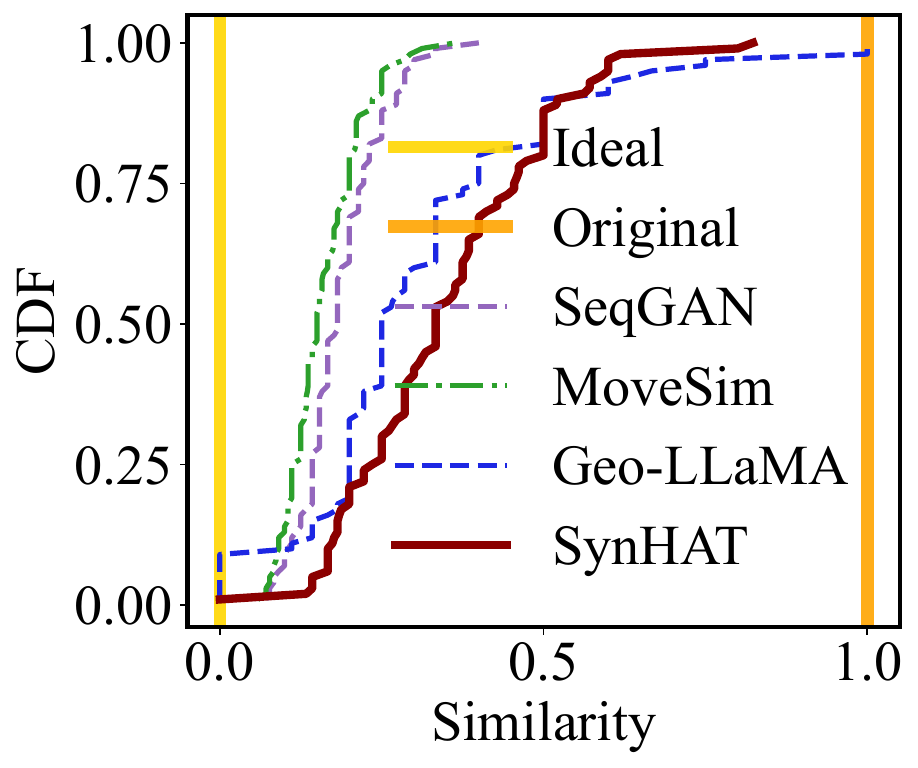}
    \caption{ATX ($tr_s$ = 2km, $tr_t$ = 2h)}
\end{subfigure}
\hfill
\begin{subfigure}[t]{0.24\textwidth}
    \includegraphics[width=\linewidth]
    {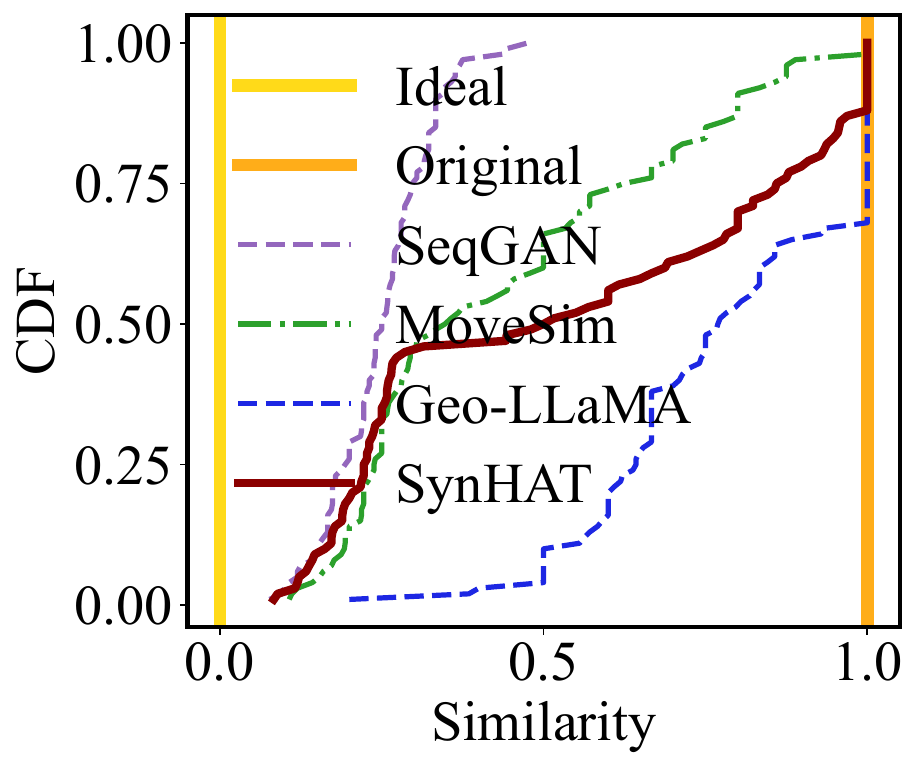}
    \caption{STO ($tr_s$ = 2km, $tr_t$ = 2h)}
\end{subfigure}

\caption{Privacy-preserving performance of HAT generation across four cities. The left vertical yellow line denotes the ideal privacy target (best-case non-memorization), and the right vertical orange line denotes the similarity CDF of the original training HATs. $tr_s$ and $tr_t$ denote the spatial and temporal thresholds for similarity computation. The closer a model’s similarity CDF is to the yellow line and the farther from the orange line, the stronger its privacy-preserving effectiveness.}
% \vspace{-6mm}
\label{fig:privacy_cdf}
\end{figure*}

The CDF curves of similarity distribution for different settings and cities are shown in Figure~\ref{fig:privacy_cdf}.
An ideal generator produces synthetic HATs with zero similarity to any training HAT in \(S\) (solid yellow line at \(x=0\)). On the other hand, a model that copies training HATs provides no privacy protection (solid orange line at \(x=1\)).
Overall, \N\ achieves a level of privacy protection comparable to strong HAT-generation baselines, even though none of the methods (including ours) employ explicit privacy mechanisms such as differential privacy (DP).
Under tight spatio-temporal matching thresholds, \N’s curve tracks the ideal line more closely than competing methods and yields a substantially larger fraction of zero-similarity cases.
\N\ exhibits strong privacy-preserving performance due to its two-stage generation pipeline. 
Specifically, it decomposes HAT synthesis into coarse- and fine-grained latent spatio-temporal state generation, followed by Semantic Alignment, using two diffusion models that learn distinct distributions.
This separation reduces direct memorization of individual training traces, leading to lower privacy leakage in practice.

\subsection{RQ4: Ablation Study}
\label{exp:ablation}

\begin{table*}[t]
\caption{Ablation studies on NYC and ATX. Reported values are Jensen–Shannon divergence (JSD); lower is better.}
\centering
\small
\begin{adjustbox}{max width=\textwidth, max totalheight=\textheight, keepaspectratio}
\begin{tabular}{|c|c|c|c|c|c|c|c|c|c|c|}
\hline
\diagbox{Method}{Metric} & Distance & Radius & Interval & Length & Average & Distance & Radius & Interval & Length & Average\\ \hline \hline
\multicolumn{1}{|c|}{} & \multicolumn{5}{c|}{\textbf{NYC}} & \multicolumn{5}{c|}{\textbf{ATX}} \\ \hline
\textit{S1} LST-UNet \textit{S-Rep}                      & 0.107 & 0.099 & 0.010 & 0.048 & 0.066 & 0.296 & 0.270 & \textbf{0.047} & 0.105 & 0.180 \\
\textit{S1} LST-UNet \textit{S-Mean}                     & 0.129 & 0.134 & 0.021 & 0.051 & 0.084 & 0.329 & 0.299 & 0.058 & 0.111 & 0.199 \\
\textit{S1} LST-UNet \textit{T-Tog}                      & 0.091 & 0.081 & 0.059 & 0.101 & 0.083 & 0.319 & 0.295 & 0.088 & 0.189 & 0.223 \\
\textit{S1} LST-UNet \textit{S-Rep}\&\textit{T-Tog}      & 0.101 & 0.079 & 0.057 & 0.099 & 0.084 & 0.321 & 0.291 & 0.087 & 0.187 & 0.222 \\
\textit{S2} LST-UNet \textit{S-Rep}                      & 0.051 & 0.069 & 0.009 & 0.049 & 0.045 & 0.247 & \underline{0.266} & 0.051 & 0.102 & \underline{0.167} \\
\textit{S2} LST-UNet \textit{S-Mean}                     & 0.107 & 0.099 & 0.010 & 0.048 & 0.066 & 0.296 & 0.270 & \textbf{0.047} & 0.105 & 0.180 \\
\textit{S2} LST-UNet \textit{T-Rep}                      & 0.061 & \underline{0.066} & 0.041 & 0.094 & 0.066 & \underline{0.241} & 0.278 & 0.059 & 0.135 & 0.178 \\
\textit{S2} LST-UNet \textit{S-Rep}\&\textit{T-Tog}      & 0.068 & 0.068 & 0.038 & 0.094 & 0.067 & 0.249 & 0.280 & 0.059 & 0.137 & 0.181 \\
\hline
S1\ LST-UNet\ w/o\ DJTG                                  & 0.069 & 0.071 & 0.034 & 0.068 & 0.061 & 0.280 & 0.313 & 0.079 & 0.111 & 0.196 \\
S2\ LST-UNet\ w/o\ DJTG                                  & 0.057 & 0.067 & 0.030 & 0.061 & 0.054 & 0.266 & 0.291 & 0.055 & 0.103 & 0.179 \\
S1+S2\ LST-UNet\ w/o\ DJTG                               & 0.070 & 0.084 & 0.040 & 0.081 & 0.069 & 0.289 & 0.328 & 0.078 & 0.115 & 0.203 \\
S2 LST-UNet\ w/o\ GC                                     & \underline{0.050} & 0.071 & 0.010 & \underline{0.044} & \underline{0.044} & 0.256 & \underline{0.266} & 0.050 & \underline{0.100} & 0.168 \\
S1+S2 LST-UNet\ w/o\ Spd\&Cur                            & 0.052 & 0.071 & \underline{0.008} & 0.051 & 0.046 & 0.280 & 0.279 & 0.048 & 0.105 & 0.178 \\
S2 LST-UNet\ w/o\ G-Context                                & 0.317 & 0.299 & 0.017 & 0.059 & 0.173 & 0.691 & 0.587 & 0.059 & 0.107 & 0.361 \\ \hline
S2\ w/o\ ST\ Perturb & 0.097 & 0.088 & 0.019 & 0.069 & 0.068 & 0.288 & 0.290 & 0.057 & 0.101 & 0.184 \\
\hline
\textbf{Complete SynHAT}                                 & \textbf{0.047} & \textbf{0.064} & \textbf{0.007} & \textbf{0.041} & \textbf{0.040} & \textbf{0.224} & \textbf{0.251} & \textbf{0.047} & \textbf{0.099} & \textbf{0.155} \\ \hline 
\end{tabular}
\end{adjustbox}
\label{tab:ablation} \vspace{-8pt}
\end{table*}

We conduct a series of ablation studies to showcase the effectiveness of the key components of our proposed \N.
The results are shown in Table~\ref{tab:ablation}. 
First, we evaluate whether the latent ST trace construction is effective in both stages.
For the coordinate sequence, instead of utilizing linear interpolation, we test two substitutes for time slots that are empty after spatial binning (Sec.~\ref{coarse-grained-Construction}): (i) \textbf{Replication (\textit{S-Rep})}—set the \(i\)-th element to the \((i\!-\!1)\)-th value; and (ii) \textbf{Mean filling (\textit{S-Mean})}—impute with the city-level mean of all POI-anchored activity coordinates. For the temporal mask, we also examine a \textbf{Temporal Toggling (\textit{T-Tog})} scheme in which the indicator flips at each real point; e.g., \([0,0,1,0,0,1,0]\) becomes \([0,0,1,1,1,0,1]\), where in every consecutive \((0,1)\) or \((1,0)\) pair the second bit marks the existence of event in the coarse-grained latent ST trace.
It is evident that our latent ST–trace construction outperforms all listed variants across every metric on both the normal-scale (NYC) and small-scale (ATX) datasets. 
In Stage 1, replacing linear interpolation in the coarse latent ST construction with \textit{S-Rep} or \textit{S-Mean} (i.e., \textit{S1} LST-UNet \textit{S-Rep}/\textit{S-Mean}) leads to a pronounced drop in spatial metrics (distance, radius), demonstrating that linear interpolation provides richer and more learnable spatial cues for Coarse-HADIff.
Moreover, although diffusion models are well-suited to signal-like time series~\citep{kong2020diffwave,chen2021wavegrad,rasul2021autoregressive,tashiro2021csdi,yang2024survey}, converting the temporal mask into a signal-like toggle within the latent ST trace markedly degrades temporal modeling and lengthens the synthesized latent ST traces, which in turn further harms spatial performance.

Secondly, we verify the effectiveness of the core denoising network LST-UNet design in both stages by replacing or eliminating core components. 
To assess the effectiveness of the DJTG core blocks in LST-UNet, we ablate them by replacing each with a standard 1D convolution layer (kernel size \(3\), stride \(1\)) at the same positions.
The replacement (S1 LST-UNet\ w/o\ DJTG, S2 LST-UNet\ w/o\ DJTG, S1+S2 LST-UNet\ w/o\ DJTG) in both stages causes LST-UNet to have a weaker capability for modeling both coarse-grained and fine-grained latent ST traces.
Meanwhile, removing the auxiliary speed and curvature inputs (S1+S2 LST-UNet w/o Spd\&Cur) in both stages degrades all four metrics, indicating that these signals that align with the intrinsic characteristics of latent ST traces can provide strong guidance.
Ablating the global context and GC-FiLM in stage 2 (S2 LST-UNet w/o G-Context) significantly degrades the performance.
This is intuitive because without the global context, the LST-UNet in stage two is only provided with the coarse-grained time slot and coordinate from stage one.  
Without the coarse-grained HAT structure, the S2 LST-UNet has no awareness of the distribution of the number of activities in each time slot among the whole HAT, which results in more short intervals (temporal) and mismatched geo-movements (spatial) compared with real HATs.

Furthermore, we assess the spatio-temporal condition perturbation (refer to Sec.\ref{training}) used to train Fine-HADiff in Stage~2 by replacing it with the unperturbed, real ST conditions from the original training set. 
Although coarse-grained latent ST traces from Stage~1 provide strong guidance, the model trained without perturbations performs worse than the variant trained with the spatio-temporal condition perturbation. 
This indicates that perturbing the ST conditions acts as an effective regularizer—improving robustness to misalignment and noise and reducing overfitting to seen conditions in the original HAT datasets.

\subsection{RQ5: Parameter Sensitivity Analysis} \label{RQ5}
Meanwhile, we study the influence of different coarse-grained time granularities $Int$ in Stage 1 (refer to Sec.\ref{coarse-grained-Construction}) on the performance of the proposed \N. 
\begin{table*}[t]
\caption{Performance comparison with different time intervals in the first stage, where lower results are better.}
\centering 
\resizebox{\textwidth}{!}{
\begin{tabular}{|c|c|c|c|c|c|c|c|c|c|c|}
\hline
\multicolumn{1}{|c|}{City} & \multicolumn{5}{c|}{NYC} & \multicolumn{5}{c|}{STO} \\ \hline
\diagbox{Int}{Metric} & Distance & Radius & Interval & Length & Average & Distance & Radius & Interval & Length & Average \\ \hline
60 mins   & 0.038  & 0.072  & 0.005 & 0.060 & 0.040 & 0.078 & 0.089 & 0.090 & \underline{0.187} & 0.114 \\ 
120 mins & \textbf{0.032} & \textbf{0.047} & \textbf{0.004}& \underline{0.051}& \textbf{0.034}& \textbf{0.061}&\textbf{0.082} &\textbf{0.057}& \textbf{0.159} & \textbf{0.090}\\ 
240 mins  & \underline{0.035}  & \underline{0.051}  & \textbf{0.004} & \textbf{0.049} & \underline{0.035} & \underline{0.069} & \underline{0.089} & \underline{0.084} & 0.208 & \underline{0.108} \\ 
480 mins  & 0.039  & 0.074  & 0.006 & 0.073 & 0.050 & 0.095 & 0.105 & 0.093 & 0.229 & 0.129 \\ 
960 mins  & 0.068 & 0.090 & 0.018 & 0.087 & 0.066 & 0.100 & 0.144 & 0.099 & 0.280 & 0.156 \\ 
 \hline
\end{tabular}
}
\label{tab:table-5}
\end{table*}

The results are shown in Table~\ref{tab:table-5}.
In general, \N\ shows robustness to the adaptation of different granularities for coarse-grained latent ST traces in Stage 1. For example, in NYC, the average JSD decreases from 0.040 at 60 minutes to 0.034 at 120 minutes, and remains competitive at 0.035 for 240 minutes;
in STO, the average JSD reaches 0.090 at 120 minutes and 0.108 at 240 minutes.
When finer $Int$ is chosen (60 mins), Stage 2 becomes easier due to shorter fine-grained sequences with points to model, but Stage 1 becomes significantly harder as it must generate longer coarse-grained sequences.
This imbalance shifts the computational burden to Stage 1, where the extended sequence length makes the model optimal harder to learn the underlying ST distribution, resulting in suboptimal overall performance (NYC: 0.040 average JSD, STO: 0.114). Conversely, extremely coarse granularities (960 mins) make Stage 2 harder by requiring each fine-grained block to model excessively long time periods with high event density, while simultaneously making Stage 1 easier due to shorter coarse-grained latent ST trace, but at the cost of losing critical temporal structure, leading to degraded quality (NYC: 0.066, STO: 0.156).
In practice, from a performance standpoint, keeping the data sizes of the \emph{coarse-grained} and \emph{fine-grained} latent ST traces in Stage~1 and Stage~2 roughly balanced is key to achieving optimal performance.

\subsection{RQ6: Computational Efficiency Analysis}

\begin{table*}[t]
\caption{Efficiency versus granularity (Stage~1 variable) for \textbf{TKY} (28 days), \textbf{NYC} (14 days), \textbf{ATX} (7 days), and \textbf{STO} (56 days). Batch size = 8. \emph{Units shown in cells:} G = GFLOPs, MB = megabytes. \emph{S1/S2 FLOPs are per-batch}; \textbf{FLOPs/HAT} = \((\text{S1}+\text{S2})/8\).}
\centering
\small
\begin{adjustbox}{max width=\textwidth, max totalheight=\textheight, keepaspectratio}
\begin{tabular}{|c|c|c|c|c|c|c|c|c|}
\hline
\diagbox{Granularity}{Metric} & S1 FLOPs & S2 FLOPs & FLOPs/HAT & Total Memory & S1 FLOPs & S2 FLOPs & FLOPs/HAT & Total Memory \\ \hline \hline
\multicolumn{1}{|c|}{} & \multicolumn{4}{c|}{\textbf{TKY}} & \multicolumn{4}{c|}{\textbf{NYC}} \\ \hline
60 mins  & 11.501G & 23.553G & 4.382G & 862MB & 11.501G & 10.381G & 2.735G & 370MB \\
120 mins  & 5.750G  & 9.819G  & 1.946G & 664MB & 5.750G  & 10.871G & 2.078G & 675MB \\
240 mins & 2.825G  & 21.941G & 3.096G & 591MB & 2.825G  & 8.769G  & 1.449G & 649MB \\
480 mins & 1.387G  & 21.672G & 2.882G & 496MB & 1.387G  & 8.501G  & 1.236G & 807MB \\
960 mins & 0.637G  & 13.096G & 1.717G & 579MB & 0.637G  & 8.366G  & 1.125G & 717MB \\ \hline \hline

\multicolumn{1}{|c|}{} & \multicolumn{4}{c|}{\textbf{ATX}} & \multicolumn{4}{c|}{\textbf{STO}} \\ \hline
60 mins  & 11.501G & 6.859G  & 2.295G & 541MB & 69.004G & 23.006G & 11.501G & 947MB \\
120 mins  & 5.750G  & 9.187G  & 1.867G & 678MB & 34.502G & 16.559G & 6.383G  & 642MB \\
240 mins & 2.825G  & 5.247G  & 1.009G & 576MB & 17.251G & 13.335G & 3.823G  & 666MB \\
480 mins & 1.387G  & 4.978G  & 0.796G & 587MB & 8.575G  & 11.723G & 2.537G  & 529MB \\
960 mins & 0.637G  & 4.844G  & 0.685G & 495MB & 4.212G  & 10.917G & 1.891G  & 596MB \\ \hline
\end{tabular}
\end{adjustbox}
\label{tab:efficiency-2x2}
\end{table*}
We evaluate the computational efficiency of \N\ during training and inference on Human Activity Traces (HATs). Table~\ref{tab:efficiency-2x2} reports results across time-slot granularities for four cities with different durations: ATX (7 days), NYC (14 days), TKY (28 days), and STO (56 days). The batch size is set to $8$.
A consistent trade-off emerges between stage one and stage two. 
As the coarse-grained time granularity $Int$ becomes longer, Coarse-HADiff’s per-batch computational cost (FLOPs) drops sharply because the coarse-grained trace length scales inversely with the interval ($L_{\text{coarse}}=\text{duration}/\text{interval}$). 
For example, in NYC the per-batch FLOPs decrease by $94.5\%$ from 11.501G to 0.637G when moving from 30-minute to 480-minute slots.
By contrast, Stage two FLOPs are mainly affected by the length of the synthetic latent ST states as the output in stage one. 
For shorter HATs (NYC, ATX), Stage~2 FLOPs change little across interval choices.
For longer HATs (TKY, STO), Stage~2 FLOPs decline as $Int$ increases, but the rate of decline quickly diminishes; beyond a certain granularity, they stabilize at an approximately constant level.
Per-HAT FLOPs \big(\((\text{S1}+\text{S2})/8\)\big) significantly drop at coarse granularities: NYC decreases from 2.735G to 1.125G ($58.9\%$), TKY from 4.382G to 1.717G ($60.8\%$), ATX from 2.295G to 0.685G ($70.2\%$), and STO from 11.501G to 1.891G ($83.6\%$). The improvements scale with HAT duration—STO (56 days) benefits the most—highlighting a key advantage of our coarse-to-fine latent ST–trace generation strategy. 
Across all cities and granularities, total memory remains under 1\, GB—370–947\, MB—even for STO’s 56-day setting, indicating a low deployment threshold for \N.
By contrast, LLM-based and graph-based generators typically demand substantially more memory for both training and inference.
However, according to RQ5 in Sec.\ref{RQ5}, although efficiency improves as the granularity gets coarser, the performance also drops.
% This is because, as the granularity gets coarser, the fine-grained latent ST traces to model in stage two get more complex, which increases the difficulties of modeling. 
This is because coarser granularity leads to more complex fine-grained latent ST traces to be modeled in stage two, thereby increasing the modeling difficulty.

In practice, keeping the latent ST–trace length in both stages within the range of $[100, 500]$ time steps generally yields strong performance.

\section{Related Work}
% As defined in Section~\ref{sec:problem statement}, a HAT represents a sequence of POI-anchored activities associated with specific times and locations.
% Existing studies on HATs can be broadly divided into two categories: those focusing on collection and analysis and those focusing on synthesis.

In this section, we summarize existing works on HAT collection based on different types of sensing devices such as GPS, Wi-Fi, and Bluetooth, as well as techniques for HAT synthesis.

% \vspace{-10pt}
\subsection{Human Activity Trace Collection and Analysis}
We organize existing studies on HAT collection and analysis into three main aspects: data acquisition and sensing modalities, activity construction and semantic enrichment, and representation and modeling for analysis. 
% Building on these discussions, we summarize the key characteristics of HATs and the corresponding challenges.

\textbf{Data Acquisition and Sensing Modalities:} The collection of HATs relies on heterogeneous sensing signals captured from personal mobile devices and urban infrastructures. On the personal side, modern smartphones integrate GPS, Wi-Fi, Bluetooth, and inertial sensors to record fine-grained spatio-temporal data, often fused with app-level contextual information such as location sharing or background activity recognition \cite{vaid2021ubiquitous}. GPS provides high spatial precision but are weakly linked to human activities, requiring further processing to infer behavioral patterns \cite{zheng2008understanding,zhang2021passive,yu2026trustenergy}. Wi-Fi positioning passively senses indoor presence and mobility by analyzing device connections and signal strengths \cite{guo2022wepos}. Bluetooth enables short-range proximity detection; for example, Eleme deployed 12,109 beacons across Shanghai stores to track couriers’ arrivals in the wild \cite{ding2021conception, ding2021nationwide}. Inertial sensors such as accelerometers and gyroscopes capture continuous motion signals for activity and mobility recognition, as demonstrated in the UK Biobank study with over 100,000 participants \cite{doherty2017large}. Moreover, manual check-ins and activity annotations collected from mobile devices provide precise and semantically rich information for constructing HAT data \cite{chang2019tourgether}. In SynHAT, POI-anchored activities are derived from these heterogeneous sensing modalities through location-based check-in detection.

\textbf{Activity Construction and Semantic Enrichment:}
Raw mobility and sensing data are often noisy, irregular, and lack semantic meaning, necessitating a transformation process to construct structured and interpretable activity traces \cite{vaizman2017recognizing}. Raw GPS or Wi-Fi traces are first processed to identify stay points, which are locations where users remain for a meaningful duration, thereby filtering out transient movements and reducing noise. This process converts dense position records into representative candidate locations for subsequent semantic anchoring \cite{nair2019understanding,liao2022wheels}. The identified stay points are then matched to POI databases or contextual maps to anchor them to real-world locations such as home, workplace, or public venues, providing the semantic foundation for activity inference \cite{nishida2014probabilistic,jiang2021transfer}. Temporal and contextual factors such as visit duration, time of day, and surrounding POIs are then used to infer activity semantics (e.g., working, dining, or commuting), and interpolation techniques reconstruct temporally continuous sequences from irregular samples \cite{yan2013semantic,hu2020mathsf}. In SynHAT, the framework relies on datasets with predefined activity labels that can be derived through the methods described above.

\textbf{Representation and Modeling for Analysis:}
Once structured activity traces are obtained, a variety of representation and modeling techniques have been developed to characterize human behavior patterns \cite{vaizman2017recognizing}. For human activity encoding, three mainstream paradigms are commonly adopted. Event-based representations treat POI-anchored activities as ordered temporal events, enabling analyses such as mobility entropy, home–work inference, and transition modeling \cite{ma2021unsupervised}. Grid-based representations discretize the spatial domain into regular regions or grids to capture aggregated mobility flows and spatial dynamics, supporting large-scale urban analysis and population behavior modeling \cite{schestakov2025trajectory}. Graph-based representations model POIs and their transitions as nodes and edges in a spatial–temporal graph, allowing graph embeddings and neural architectures such as GCNs or ST-GNNs to capture complex dependencies and behavioral patterns \cite{yang2021spatio,tabatabaie2023cross}. More recently, large language models have been employed to encode HAT sequences through semantic alignment and contextual reasoning \cite{wei2025one}. In SynHAT, we do not adopt a specific HAT representation; instead, the synthesized data can flexibly integrate into diverse representation frameworks and support downstream analytical tasks.

\subsection{Diffusion-based Discrete Sequence Generation}
Recent diffusion-based \textit{GenAI} has been extended beyond continuous signals to a wide range of \emph{discrete} sequence generation problems. 
In \textit{text generation}, discrete-state diffusion models define categorical noising and denoising directly on token spaces via state-transition processes~\cite{austin2021structured}, while alternative approaches diffuse in continuous embedding spaces to enable controllable generation and flexible guidance~\cite{li2022diffusion}. 
In \textit{image generation}, images can be represented as sequences of discrete visual tokens, and generated through masked parallel decoding with iterative refinement, offering a non-autoregressive alternative to left-to-right decoding~\cite{chang2022maskgit}. 
In \textit{graph generation}, diffusion has been adapted to structured discrete objects by modeling categorical node/edge attributes and iteratively denoising graph structures~\cite{vignac2022digress}.
Moreover, coarse-to-fine hierarchical diffusion further improves global-to-local consistency and scalability by generating coarse structures before refining fine details (e.g., for molecular generation)~\cite{qiang2023coarse}. 
Across these domains, a recurring design trade-off emerges between \emph{semantic richness} and \emph{efficiency}: rich continuous latents or large pretrained representations can capture nuanced dependencies but often incur higher inference cost, whereas purely discrete transitions are efficient yet may lack expressive semantics without additional structure. 
Motivated by this perspective, SynHAT adopts a coarse-to-fine decomposition that first generates an efficient spatio-temporal ``skeleton'' of latent movement states and then aligns these states with fine-grained spatio-temporal semantics, enabling both faithful activity-transition modeling and practical generation efficiency.

\vspace{-10pt}
\subsection{Human Activity Trace Synthesis}
As HAT data becomes increasingly important yet often limited by privacy, coverage, and accessibility constraints, researchers have explored various approaches to synthesize it for simulation, data augmentation, and privacy-preserving analytics. HAT synthesis aims to generate realistic and behaviorally consistent activity sequences that capture both spatial regularities and temporal dependencies observed in the data \cite{kong2023mobility}. Existing studies on HAT synthesis can be broadly categorized into four paradigms: rule-based and statistical models; GAN-based models; diffusion-based models; and LLM-based synthesis.

\textbf{Rule-based and Statistical Model for Synthesis:}
Early efforts on HAT synthesis primarily relied on rule-based and statistical frameworks that simulate human mobility using behavioral heuristics and aggregated statistics. Rule-based simulation frameworks generate HATs by explicitly modeling behavioral rules and mobility routines, often through agent-based or heuristic simulations that incorporate temporal and contextual constraints \cite{jiang2016timegeo,cai2021simulating}. Markovian and statistical transition models describe human mobility as stochastic transitions between locations or activity states, relying on conditional probabilities or statistical distributions to efficiently reproduce aggregated movement patterns \cite{yin2017generative}. Probabilistic and Bayesian generative frameworks employ probabilistic graphical models, such as hierarchical Bayesian or topic-based formulations, to infer latent activity semantics and generate realistic spatio-temporal sequences \cite{uugurel2024learning}. Despite their interpretability and simplicity, these methods struggle to capture complex temporal dependencies, heterogeneous behaviors, and fine-grained individual dynamics present in real-world HATs.

\textbf{GAN-based Synthesis:}
Generative adversarial networks (GANs) have shown strong capability in synthesizing realistic and diverse HATs by learning from large-scale authentic data. Early approaches employed sequence-based GANs such as SeqGAN to model POI visit sequences through adversarial training with policy gradients \cite{yu2017seqgan}. Subsequent works extended this framework to capture fine-grained spatio-temporal dependencies and population-level mobility dynamics, enabling the generation of both individual traces and aggregated urban flow patterns \cite{ouyang2018non,feng2020learning,yuan2022activity}.
Moreover, several studies in the pervasive computing domain investigate GAN-based HAT synthesis from a system perspective, emphasizing practical data generation, robustness, and evaluation in real-world sensing environments. For example, SynthCAT \cite{lyu2024synthcat} synthesizes cellular association traces through a fusion of model-driven and data-driven approaches, while ActivityGAN \cite{li2020activitygan} and CrossHAR \cite{hong2024crosshar} leverage adversarial frameworks to augment sensor-based human activity datasets and enhance cross-dataset generalization. Despite their ability to capture complex spatio–temporal dependencies, GAN-based approaches often suffer from unstable training dynamics and mode collapse, which hinder their scalability, reproducibility, and consistency in modeling fine-grained, long-term human behaviors.

\textbf{Diffusion-based Synthesis:}
Diffusion probabilistic models have recently emerged as a powerful class of deep generative methods that model complex data distributions through iterative denoising~\cite{dhariwal2021diffusion}. Building on these advances, several studies have explored their potential for mobility data generation. Representative works such as DiffTraj~\cite{zhu2023difftraj} and ControlTraj~\cite{zhu2024controltraj}, employ diffusion architectures to synthesize GPS trajectories with fixed sampling intervals and continuous spatial coordinates, effectively capturing spatial continuity and temporal regularity in human movement. These models demonstrate stable training dynamics and strong controllability compared with GAN-based models.
However, current diffusion-based approaches remain limited to continuous trajectories like GPS trajectories and cannot be extended to HATs, which are POI-anchored with semantics, spatially discrete, and temporally irregular. Adapting diffusion models to HAT synthesis, therefore, requires rethinking temporal conditioning and semantic representation to handle heterogeneous activity types and irregular event intervals.

\textbf{LLM-based Synthesis:}
Large Language Models (LLMs) have recently demonstrated strong capabilities in structured sequence generation and contextual reasoning, motivating their adaptation for spatio-temporal data modeling~\cite{guo2025language}. Although no existing studies have directly applied LLMs to model semantic correlations within HATs for downstream tasks, recent works such as UrbanGPT~\cite{li2024urbangpt}, MobGLM~\cite{zhang2024mobglm}, and Geo-Llama~\cite{li2024geo} have explored their extensions to spatio-temporal domains. However, these models still struggle to capture complex spatial dependencies effectively. More importantly, training LLMs requires intensive computational resources and long processing times, making them currently impractical and unscalable for large-scale HAT generation.
Under our current implementation and comparison, SynHAT outperforms the evaluated LLM-based baseline GeoLLaMa.

\section{Discussions}
In this section, we provide some lessons learned, limitations, and future directions, and ethical considerations. 

\subsection{Lessons Learned}
\subsubsection{Significance of our work.} 
(i) \textbf{For Data Accessibility}. With growing concerns over data privacy, obtaining large-scale, user-level activity data has become increasingly difficult. Furthermore, deploying extensive sensing infrastructures for high-resolution data collection is prohibitively expensive. For example, the instant delivery company Ele.me deployed the aBeacon system in Shanghai to accurately detect couriers’ arrival activities, involving more than 12,000 devices and an annual cost exceeding \$3.6 million \cite{ding2021nationwide}. As a result, many studies in the pervasive computing community are limited to small-scale user experiments. For instance, AuraRing \cite{parizi2019auraring} evaluated its system with only 12 participants, while HomeIoT-Wellbeing \cite{koh2025harnessing} conducted a four-week study with 20 participants. Even in these small-scale settings, privacy concerns often prevent researchers from sharing their collected data, which significantly hinders the reproducibility, validation, and broader applicability of their findings. To address this limitation, our proposed SynHAT provides a practical and privacy-preserving solution that enables researchers to share and reproduce human activity data research without exposing sensitive user information. 
(ii) \textbf{For Data Augmentation}. Even when human activity data are available, they are often limited in scale and completeness due to the high cost of data collection and the high variability of human behaviors. The sparse and irregular sampling of real-world data, caused by incomplete sensing coverage, inconsistent user engagement, and missing records, leads to fragmented and biased behavioral representations, making it difficult for deep learning models to capture continuous mobility patterns and contextual dependencies~\cite{zhou2021stuanet,zhu2023difftraj}. To mitigate this limitation, our proposed SynHAT serves as an effective data augmentation framework that generates realistic, fine-grained synthetic activity traces to enrich small or incomplete datasets. By augmenting real-world data with synthetic samples, SynHAT helps to alleviate data sparsity, improve model robustness, and enable data-intensive learning in human activity research.
Large-scale synthetic HATs can be efficiently compressed and stored using existing data compression methods~\cite{zhang2025lcp, yang2025ipcomp}.

\subsubsection{Effectiveness of our approach.}
(i) \textbf{Problem Reformulation}.
Our method achieves good performance through an effective problem transformation that aligns with the strengths of diffusion models.
Specifically, we reformulate the spatially discrete and temporally irregular HAT synthesis problem into two spatially continuous and temporally regular latent ST trace synthesis subproblems. 
By decomposing the problem into two stages, in both stages, Coarse-HADiff and Fine-HADiff generate smoother spatial movements within regular temporal windows, which are well-suited to the diffusion model's continuous denoising process.

(ii) \textbf{Novel Denoising Network Design}.
We fully take advantage of the unique characteristics of our constructed latent ST traces. Our proposed flexible HADiff with a deliberately-designed LST-UNet further enhances HAT synthesis quality. 
We design several effective designs within LST-UNet, which facilitate better latent ST trace modeling in both stages, as proved in the ablation studies in Sec.~\ref{exp:ablation}.
For example, the introduction of the DJTG block significantly improves the LST-UNet's modeling capacity, and the Spatio-temporal condition plays an important role in guiding the conditional generation in Stage 2. 
In real practice, it is self-evident that a well-suited neural network design in the deep learning framework is crucial.

\subsubsection{Evaluation of our approach} Evaluation is a critical component of HAT synthesis research. In this work, we conduct a comprehensive assessment of our model from multiple perspectives, including generation fidelity, downstream utility, model generalizability, computational efficiency, and privacy-preserving capability. An effective model should not excel in only a single aspect but instead achieve a balanced trade-off across these complementary dimensions. The experimental results show that our proposed two-stage coarse-to-fine framework, SynHAT, consistently delivers strong performance across all evaluation criteria,  demonstrating its robustness, scalability, and practicality for real-world HAT synthesis.

\subsection{Limitations and Future Work}
While our model can effectively generate high-quality synthetic traces, several aspects can be further improved in future studies.
(i) Although synthetic HATs can alleviate privacy concerns, potential privacy risks may still exist in both the synthesis models and the generated data~\cite{kunlin}, as such models might memorize portions of the original data during training.
In Sec. 4.4, we evaluate the privacy-preserving performance of SynHAT and the baselines by computing similarity between synthetic HATs and authentic HATs. However, this similarity-based metric doesn't have a direct quantitative privacy-preserving effectiveness with solid theory and proof.
To our knowledge, there is no widely recognized privacy evaluation mechanism for this task.
Existing membership inference attack (MIA) evaluations are mostly developed for tabular data, images, or generic generative models, and directly transferring them to check-in trajectories is non-trivial because (i) trajectories are variable-length spatio-temporal sequences, (ii) privacy risks depend on trajectory-specific distance functions and alignment, and (iii) realistic adversarial knowledge often consists of partial spatio-temporal subsequences, which differs from typical IID record settings. 
A promising direction for our future work is to integrate advanced privacy-preserving mechanisms, such as Differential Privacy (DP), into the model training process. This approach could further enhance privacy protection but may also introduce additional computational overhead and longer training times.
(ii) Under our current implementation and comparison, SynHAT outperforms the evaluated LLM-based baseline GeoLLaMa. As shown in Table~\ref{tab:fidelity}, LLM-based methods currently demonstrate relatively weak performance in spatial generation, though they perform well in the temporal dimension. Meanwhile, large-scale LLMs still face high computational costs and limited scalability, making them impractical for efficient HAT synthesis. Nevertheless, their strong temporal reasoning ability suggests potential for future research once their spatial modeling capability and training efficiency are improved. Hence, LLM-based synthesis could serve as a complementary direction when efficiency is not the primary concern, whereas our work focuses on achieving efficient and scalable modeling.
Recent studies show LLMs can benefit from explicit geo-spatial encodings (e.g., map-/graph-aware or spatio-temporal dependency encoders) and from large-scale pretraining and continued pretraining on diverse mobility tasks, suggesting clear headroom for improving LLM baselines \cite{manvi2023geollm, gong2024mobility, li2024urbangpt}.

\subsection{Ethical Use of Data}
The data used in this paper are anonymized public datasets, which have been extensively used by existing papers \cite{xie2016learning,zhao2020go,yin2023next}. In this work, we are not trying to identify individual users' private information. Instead, our work aims to generate synthetic HATs, which can further help protect user privacy if data owners plan to release HATs for social good or along with their publications. Therefore, this study has no ethical concerns.

\section{Conclusion}
In this paper, we design a two-stage coarse-to-fine HAT synthesis framework called \N\ to address two fundamental research challenges, i.e., (i) capturing complicated spatio-temporal behavior patterns from spatially discrete and temporally irregular HATs, and (ii) achieving high synthetic data quality with low computational costs. An innovative human activity diffusion model with an effective Latent Spatio-Temporal UNet (LST-UNet) is designed for data synthesis. We extensively evaluate the proposed \N\ based on real-world HAT data from four cities in three countries for generalizability. Experiments demonstrate superior performance on synthesis fidelity, data utility for downstream applications, privacy-preserving capability, and computational efficiency of the proposed \N. For example, \N\ outperforms the best baseline by 52\% and 33\% on spatial and temporal metrics on the TKY dataset, respectively.

\section{Acknowledgments}
We thank all the reviewers for their insightful feedback to improve this paper.
This work is partially supported by the National Science Foundation under Grant 2411152, National Artificial Intelligence Research Resource (NAIRR) 240332, and
FSU/AWS Computer Support Seed Fund.
%000278

% In this paper, we designed a two-stage spatio-temporal conditional denoising diffusion model called \N\ for LBSN trace generation. In \N, a multi-step spatio-temporal interpolated trace simulator is first designed to capture spatio-temporal distributions from interpolated asynchronous traces and generate synthetic coarse-grained interpolated traces. 
% A sequential continuous trace filling module is then proposed to generate missing trace records between transitions in the generated coarse-grained interpolated trace.
% We conducted comprehensive experiments to evaluate the performance of \N\ based on two real-world datasets. 
% Experiments for both fidelity and utility verification show \N\ outperforms state-of-the-art baselines in terms of spatial and temporal metrics by 60\% and 38\%, respectively. 
% An in-depth analysis was also conducted to show the strengths of our design.

\bibliographystyle{ACM-Reference-Format}
\bibliography{main}

% \clearpage 
\appendix
\section*{Appendix}

\section{Implementation Details}
We implemented SynHAT in PyTorch and trained it on a single  NVIDIA A10 GPU with 24 GB memory. For both stages, we set the model use a base channel dimension of 128, 4 resolution scales with multipliers [1, 2, 4, 8], and 2 residual blocks per scale with group normalization. All embeddings are 32-dimensional.
In Stage 1, we generate coarse-grained latent ST traces with sequence length $D/Int$ determined by HAT duration $D$ and coarse-grained time granularity $Int$.
For Coarse-HADiff training uses $T = 1000$ diffusion timesteps with a cosine schedule, AdamW optimizer ($\alpha = 2 \times 10^{-4}$), batch size 64, gradient clipping at 1.0, and EMA decay 0.999 for 2000 epochs.
In Stage 2, we utilize the same hyperparameter settings as Stage 1 for Fine-HADiff. 
We set the  $\alpha_0 = 0.15$, maximum event weight $\alpha_1^{\max} = 6.0$, neighbor coefficient $\alpha_{\text{near}} = 0.6$, and Gaussian blur radius $r = 2$. Event emphasis uses cosine warm-up over 15\% of training with per-sequence normalization (mean=1). Training follows the same configuration as Stage 1.
For inference of both stages, we use DDIM sampling with 50 steps ($\eta = 0.0$). 
Stage 1 identifies activated slots via a stay threshold of 0.5.
Stage 2 generates blocks only for activated slots with an event detection threshold of 0.7.
Stage 3 performs POI ranking with learned embeddings.

For our baseline comparisons, we report the key hyperparameters selected to ensure fair and competitive performance. The SMM \cite{maglaras2015social} model's number of states was set to the total number of unique POIs in each respective dataset. For TimeGEO \cite{jiang2016timegeo}, following the original author's suggestions, we set the return probability $\rho = 0.6$ and the exploration probability $\gamma = 0.2$. The parameters for the Hawkes \cite{laub2015hawkes} process, which utilizes an exponential kernel, were optimized via maximum likelihood estimation on the training data. The deep learning methods, LSTM \cite{rossi2021vehicle}, SeqGAN \cite{yu2017seqgan}, and MoveSim \cite{feng2020learning}, were configured with a hidden dimension and embedding size of 128 and trained with a batch size of 64 using the Adam optimizer with a learning rate of $1 \times 10^{-3}$. 
For SeqGAN, the number of Monte Carlo search rollouts was set to 16. 

To adapt DiffTraj \cite{zhu2023difftraj} and ControlTraj \cite{zhu2024controltraj}, originally designed for spatially continuous, temporally regular trajectory generation, for fair comparison, we follow the same latent movement sequence reconstruction process and extract the two spatial dimensions as the training and target data for them. 
The continuous trajectory is originally formatted as a sequence $\{(c_1, 0), (c_2, \tau), (c_3, 2\tau) \dots\}$, where each coordinate $c_i$ can be sampled in the continuous spatial space and the time interval $\tau$ between $c_i$ and $c_{i+1}$ is pre-determined. 
In practice, we assign $\tau$ equal to one hour to strike a balance between efficiency and granularity.
We transform each HAT $s = [(a_1,t_1), (a_2,t_2), \ldots, (a_n,t_n)]$ in the original dataset to the continuous trajectory according to the same process of constructing an \textbf{interpolated coordinate sequence} mentioned in Sec. 3.1.1. 
These continuous trajectory synthesis baselines aim to simulate the distribution of the continuous trajectory set and generate synthetic continuous trajectories. 
% We do not change the format of continuous trajectories as the input of these baselines, considering their deliberate design for the continuous trajectory characteristics. 
After obtaining each synthetic continuous trajectory $\{(c_1', 0), (c_2', \tau), (c_3', 2\tau) \dots\}$ generated by baselines, we identify each turning point index $i$ through determining whether the moving angle change exceeds a pre-determined threshold $\theta_{thr}$ through:
\begin{equation}
\vec{u} = co(c'_{i-1}) - co(c'_i), \quad \vec{v} = co(c'_i) - co(c'_{i+1})
\end{equation}

\begin{equation}
\theta = \arccos \left( \frac{\vec{u} \cdot \vec{v}}{\|\vec{u}\| \|\vec{v}\|} \right)
\end{equation}
$(c'_i, i\tau)$ is selected if $\theta > \theta_{thr}$. 
We proceed following this because intermediate points between actual activity locations are \textit{linearly} interpolated.  
Finally, we obtain each corresponding POI-anchored activity $a'_i$ according to the same process of \textbf{Semantic Alignment} illustrated in Sec. 3.2.3.
Since these continuous trajectory generation baselines are not deliberately designed for HAT generation, we also tried to use other adaptation strategies and found that their results are extremely poorer than the above adaptation strategy: 
a) filling a dummy geo-coordinate in the points where no event happens: it makes baselines fail to capture spatio-temporal dependency because the trajectories to model are no longer continuous;
b) getting the spatially-closest POI-anchored activity to each selected $c'_i$: this makes baselines have no capability to model the POI-anchored activity density distribution in the area. 
In terms of the training, we utilize $T=1000$ diffusion steps and a linear noise schedule from $\beta_{start}=10^{-4}$ to $\beta_{end}=0.02$. For ControlTraj, we set the constraint guidance scale to 1.5. The backbone network for both diffusion baselines is a standard UNet architecture.

We implemented Geo-LLaMA~\cite{li2024geo} following its “trajectory-as-text” serialization pipeline, which consisted of discretized spatial IDs and time bins, and next-text-token training. 
We utilize the pretrained LLaMA-7B backbone with LoRA fine-tuning (rank 8, alpha 16) for trajectory generation, training for 20 epochs with batch size 4 and learning rate $2 \times 10^{-4}$. 
We used comparable training budgets across baselines to ensure fairness. 
In practice, we fine-tuned Geo-LLaMa for 10 epochs in each city in our experiments.

% \clearpage
\section{Extented Evaluations}

\subsection{Extended Fidelity Evaluations} \label{ext_fid}

% \begin{figure*}[t]
% \centering
% \begin{subfigure}[t]{0.43\textwidth}
%     \includegraphics[width=\linewidth]{Figures/Radars/radar_TKY_times.pdf}
%     \caption{TKY}
% \end{subfigure}
% \hfill
% \begin{subfigure}[t]{0.43\textwidth}
%     \includegraphics[width=\linewidth]{Figures/Radars/radar_NYC_times.pdf}
%     \caption{NYC}
% \end{subfigure}
% \hfill

% \vspace{1ex} 

% \begin{subfigure}[t]{0.43\textwidth}
%     \includegraphics[width=\linewidth]{Figures/Radars/radar_ATX_times.pdf}
%     \caption{ATX}
% \end{subfigure}
% \hfill
% \begin{subfigure}[t]{0.43\textwidth}
%     \includegraphics[width=\linewidth]
%     {Figures/Radars/radar_STO_times.pdf}
%     \caption{STO}
% \end{subfigure}

% \caption{\textcolor{red}{Fidelity results shown in radar plots.}}
% \label{fig:utility_radar}
% \end{figure*}   

% \vspace{-10pt}
\begin{figure*}[t]
    \centering
\includegraphics[width=\linewidth]{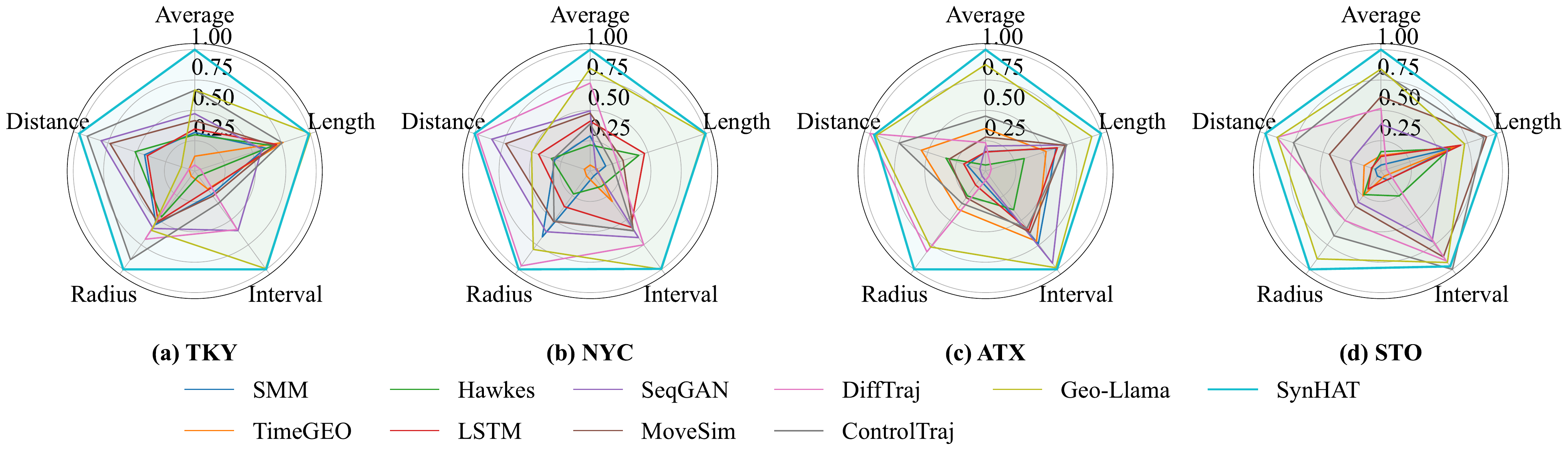}
% \vspace{-25pt}
  \caption{Radar plots of trajectory fidelity metrics across four cities (TKY, NYC, ATX, STO). Each subplot illustrates model performance on five evaluation dimensions. To enable fair cross-metric comparison, all metric values are linearly rescaled within each city using inverted min–max normalization so that higher values correspond to better fidelity (i.e., smaller JSDs). The closer a value is to 1, the better the performance; therefore, larger polygon areas indicate superior overall performance in reproducing real-world mobility patterns.}
  \label{fig:utility_radar}\vspace{-10pt}
  % \vspace{-5pt}
\end{figure*}
\begin{figure*}[h]
\centering

% --- Row 2: FS-ATX Dataset ---
\begin{subfigure}[t]{0.22\textwidth}
    \includegraphics[width=\linewidth]{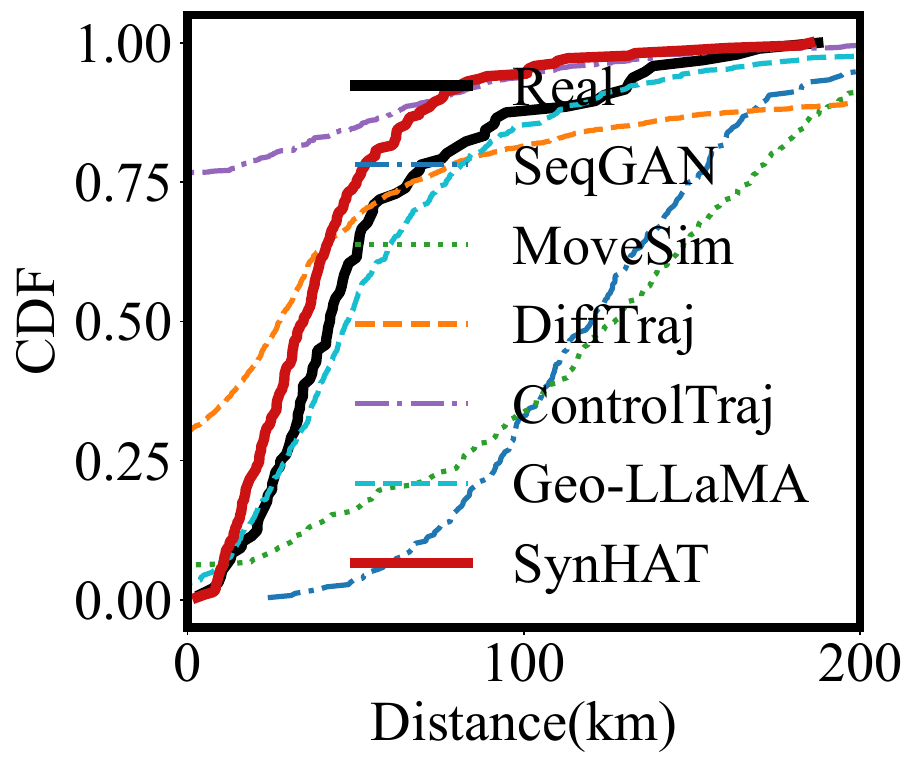}
    \caption{CDF of distance in ATX}
\end{subfigure}
\hfill
\begin{subfigure}[t]{0.22\textwidth}
    \includegraphics[width=\linewidth]{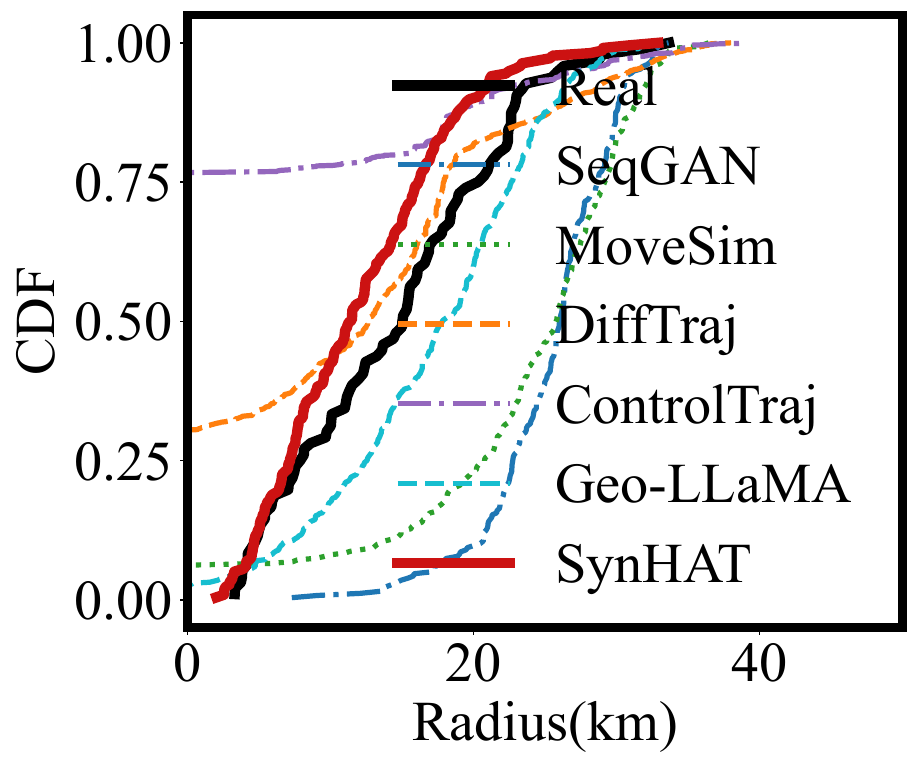}
    \caption{CDF of radius in ATX}
\end{subfigure}
\hfill
\begin{subfigure}[t]{0.22\textwidth}
    \includegraphics[width=\linewidth]{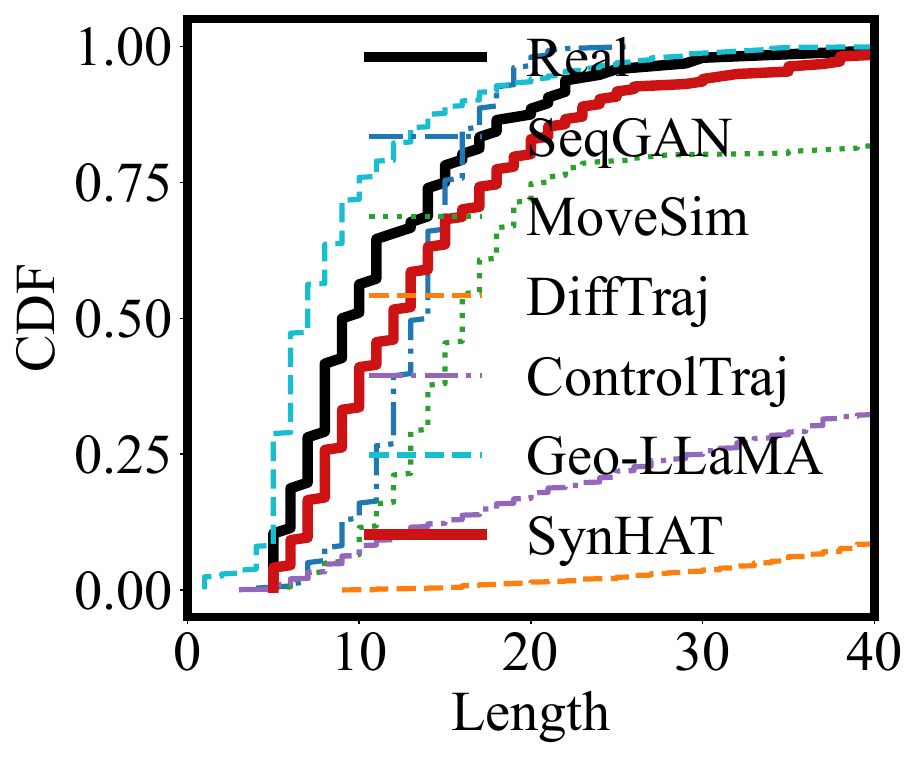}
    \caption{CDF of length in ATX}
\end{subfigure}
\hfill
\begin{subfigure}[t]{0.22\textwidth}
    \includegraphics[width=\linewidth]{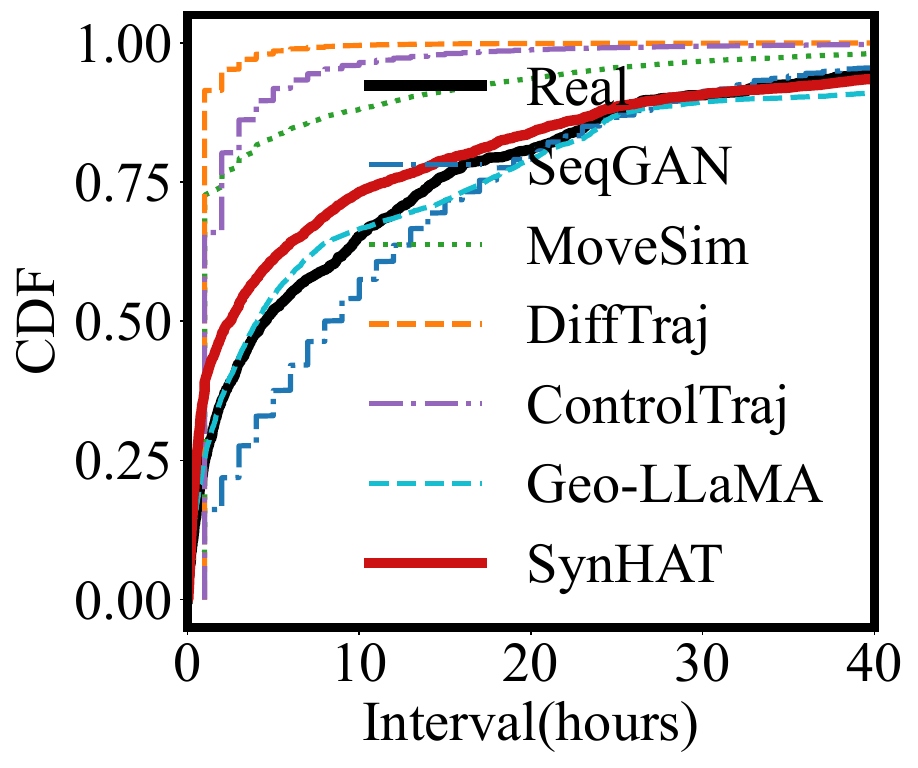}
    \caption{CDF of interval in ATX}
\end{subfigure}

\vspace{1ex} 

% --- Row 3: GW-STO Dataset ---
\begin{subfigure}[t]{0.22\textwidth}
    \includegraphics[width=\linewidth]{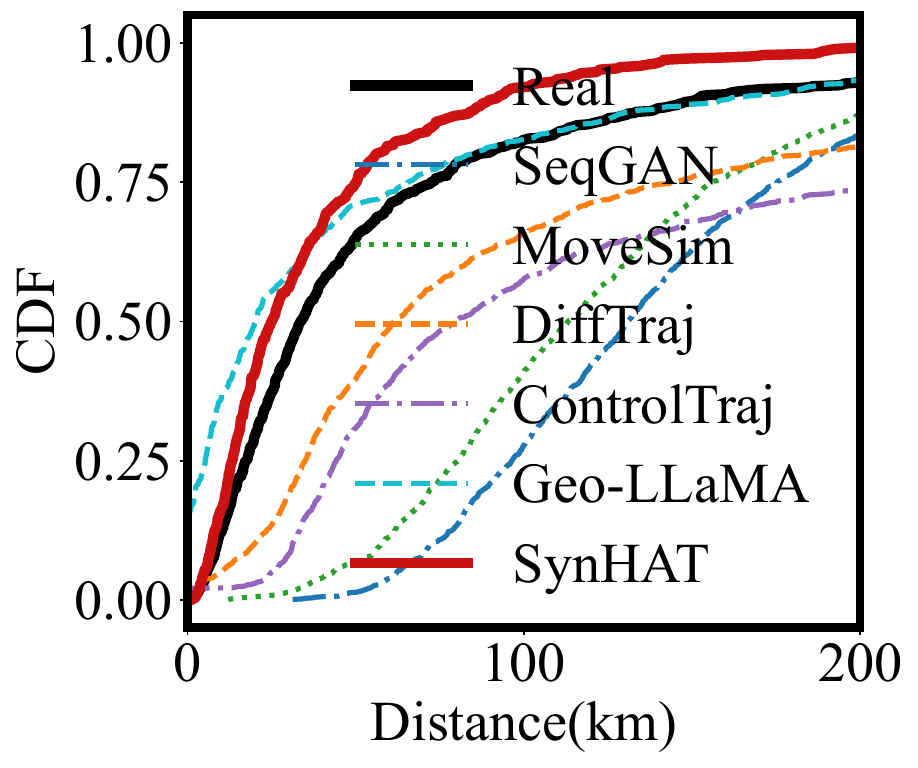}
    \caption{CDF of distance in STO}
\end{subfigure}
\hfill
\begin{subfigure}[t]{0.22\textwidth}
    \includegraphics[width=\linewidth]{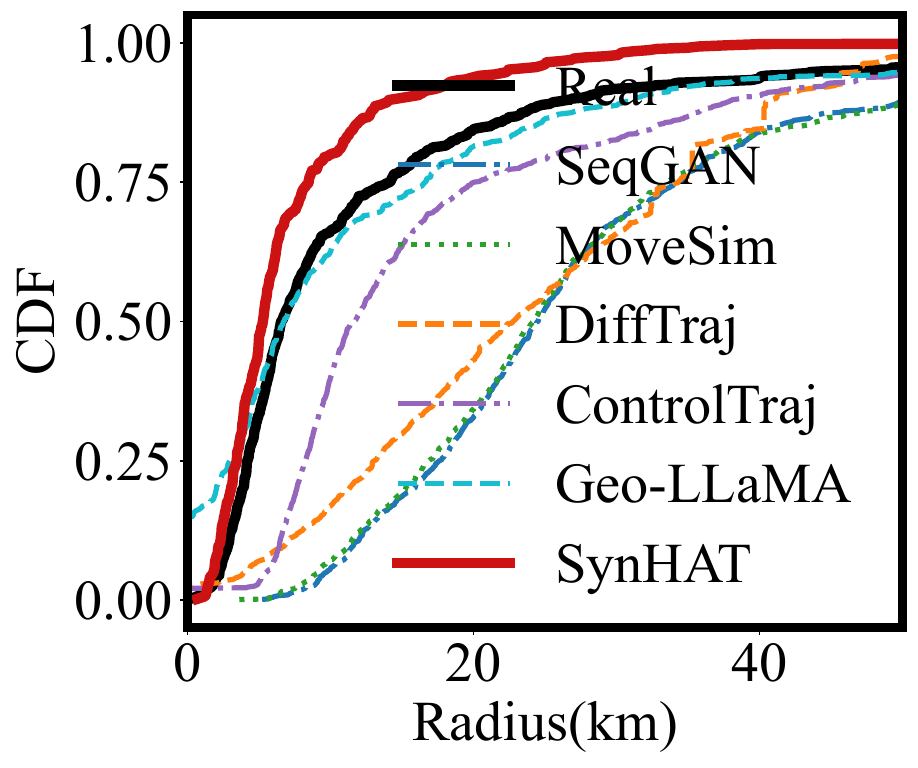}
    \caption{CDF of radius in STO}
\end{subfigure}
\hfill
\begin{subfigure}[t]{0.22\textwidth}
    \includegraphics[width=\linewidth]{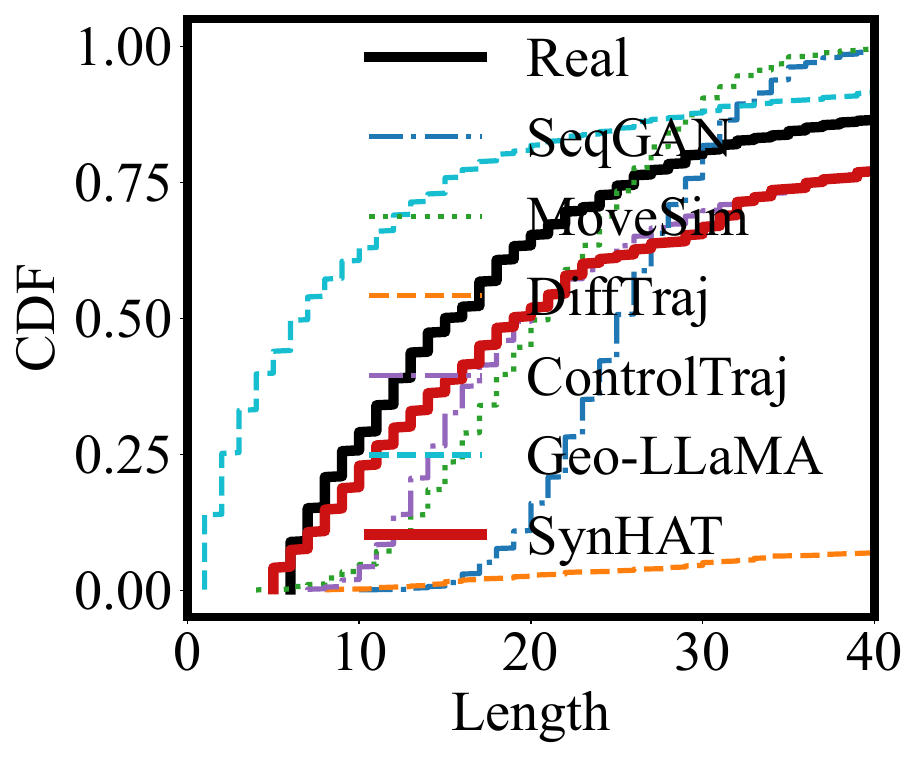}
    \caption{CDF of length in STO}
\end{subfigure}
\hfill
\begin{subfigure}[t]{0.22\textwidth}
    \includegraphics[width=\linewidth]{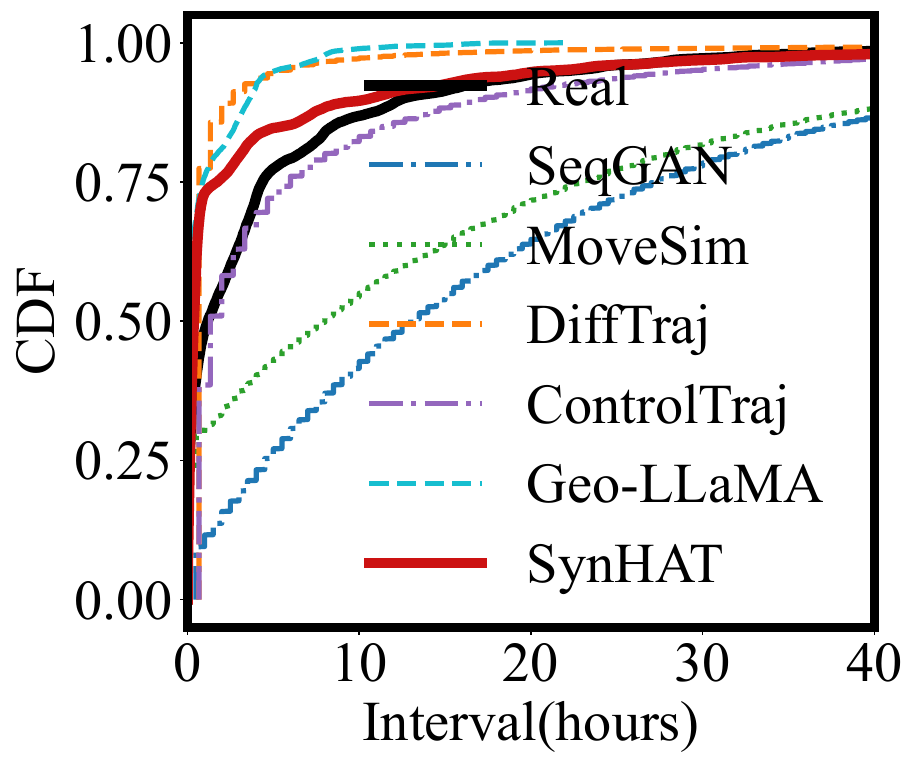}
    \caption{CDF of interval in STO}
\end{subfigure}
\vspace{-10pt}
\caption{HAT generation fidelity performance comparison in ATX and STO.}\vspace{-10pt}
\label{fig:app_fidelity_ext}
% \vspace{-5pt}
\end{figure*}
\begin{table*}[b] \small
\centering \vspace{-12pt}
\caption{Transition Evaluation. The best results on each dataset are in \textbf{bold}, and the second-best results are \underline{underlined}.}
\label{tab:your_label}\vspace{-10pt}
\begin{tabular}{lccc}
\toprule
\diagbox{Method}{City} & NYC & TKY & ATX \\
\midrule
ControlTraj & 0.048 & 0.225 & 0.973 \\
DiffTraj    & 0.263 & 0.296 & 0.843 \\
Geo-LLaMa   & 0.049 & \textbf{0.023} & 0.207 \\
SeqGAN     & 0.059 & 0.168 & 0.510 \\
MoveSim     & \underline{0.038} & 0.053 & \underline{0.147} \\
SynHAT (ours)     & \textbf{0.017} & \underline{0.035} & \textbf{0.145} \\
\bottomrule
\end{tabular}\vspace{-10pt}
\label{table:transition}
\end{table*}

To facilitate interpretation of the fidelity metrics, we visualize the results as radar plots in Figure~\ref{fig:utility_radar}, based on the values reported in Table~\ref{tab:fidelity}.
We present Cumulative Distribution Function (CDF) curves to illustrate the distributions of various metrics, as shown in Fig.~\ref{fig:app_fidelity_ext} in ATX and STO. 
Compared with the NYC and TKY datasets, the scale of the ATX and STO datasets is smaller. 
Across Fig.~\ref{fig:app_fidelity_ext} (a) to (h), the CDFs associated with \N\ closely align with those of the original data.
The results suggest that \N\ is capable of replicating the spatio-temporal patterns found in the actual data with high fidelity for various scales of datasets. 

\begin{figure*}[t]
\centering

% Row 1: NYC (Real, DiffTraj, SeqGAN)
\begin{subfigure}[t]{0.30\textwidth}
    \centering
    \includegraphics[width=\linewidth]{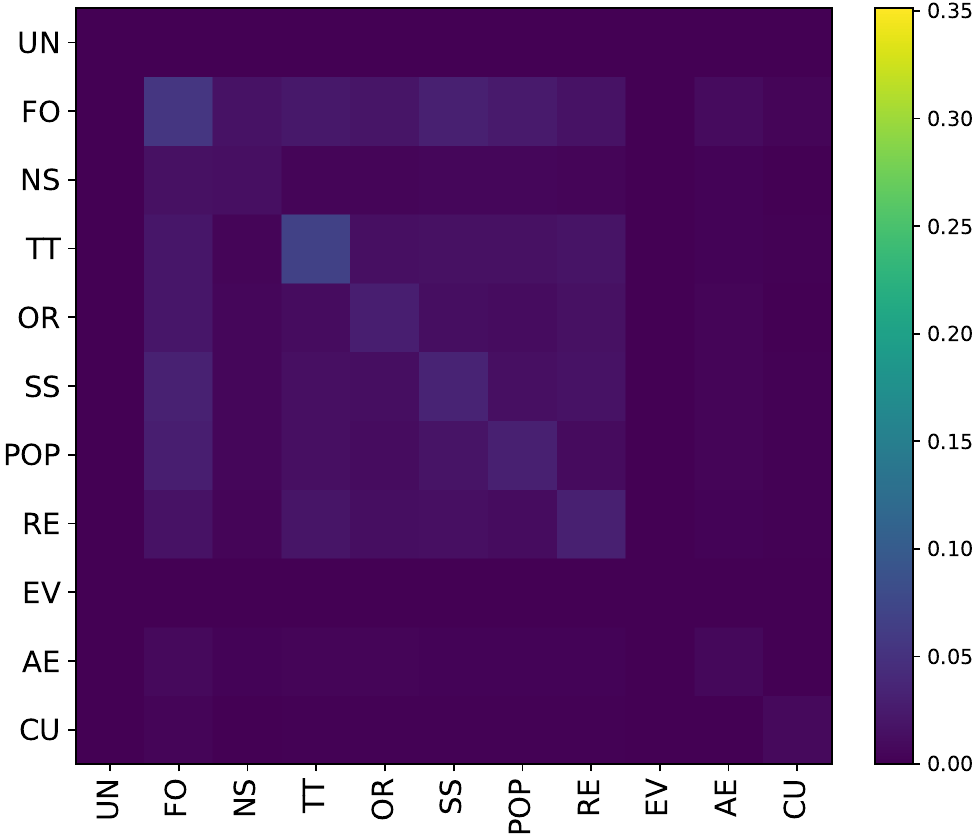}
    \caption{NYC: Real}
\end{subfigure}
\hfill
\begin{subfigure}[t]{0.30\textwidth}
    \centering
    \includegraphics[width=\linewidth]{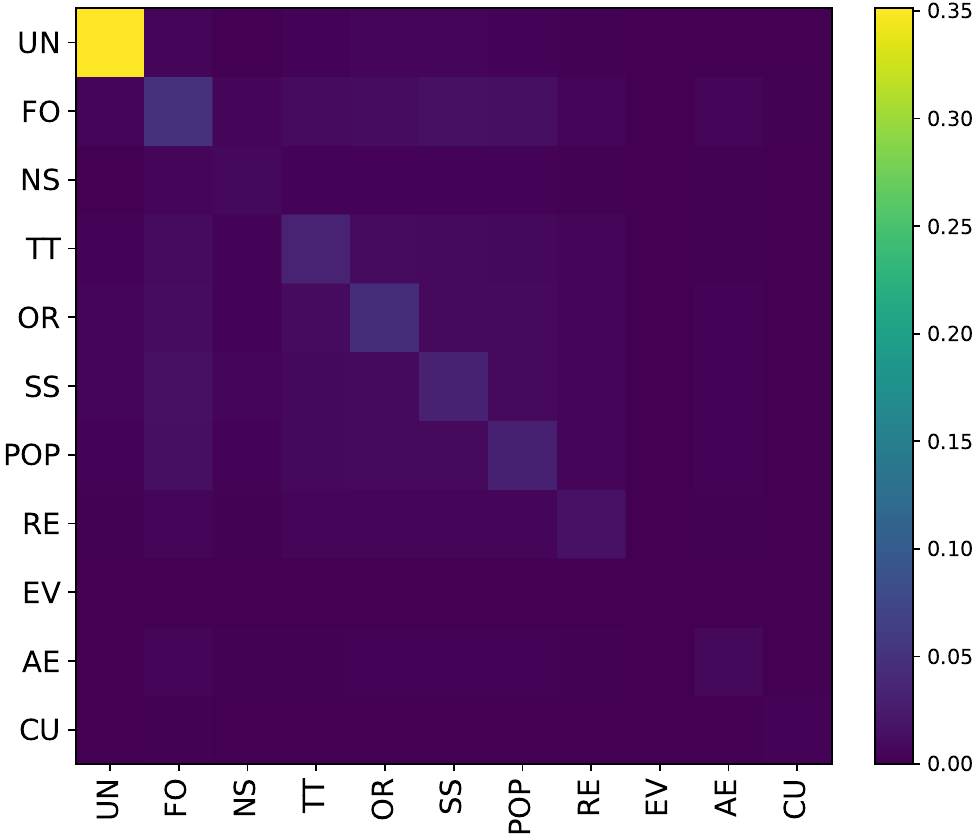}
    \caption{NYC: DiffTraj}
\end{subfigure}
\hfill
\begin{subfigure}[t]{0.30\textwidth}
    \centering
    \includegraphics[width=\linewidth]{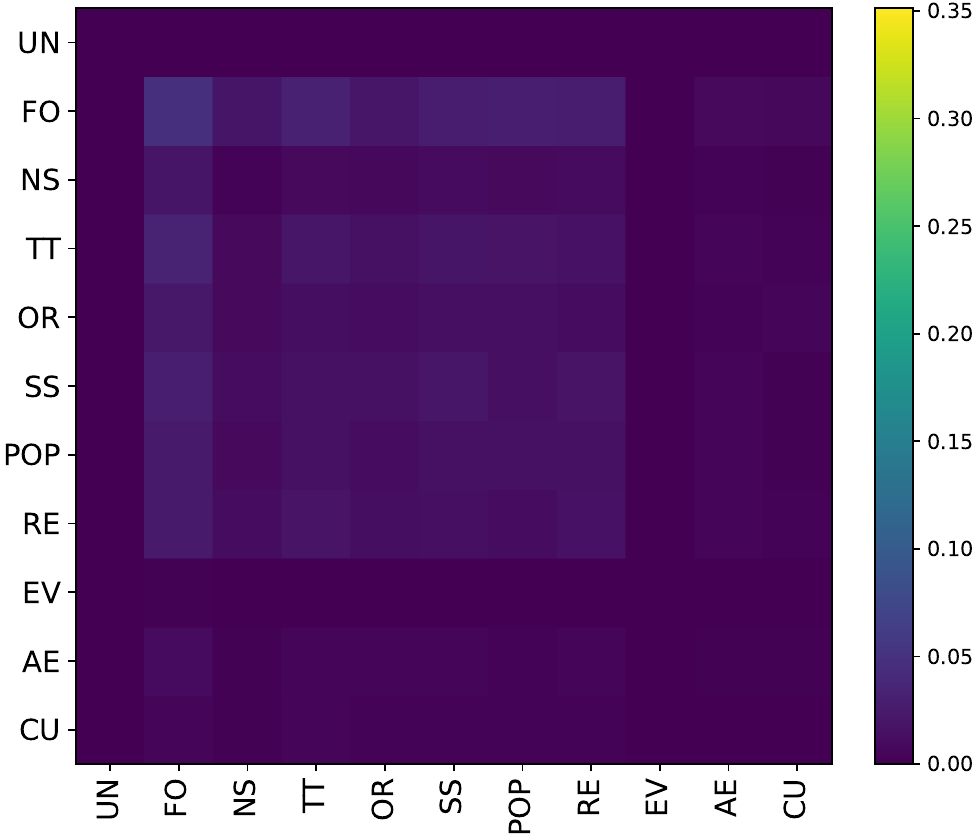}
    \caption{NYC: SeqGAN}
\end{subfigure}

\par\medskip

% Row 2: NYC (MoveSim, Geo-LLaMA, SynHAT)
\begin{subfigure}[t]{0.30\textwidth}
    \centering
    \includegraphics[width=\linewidth]{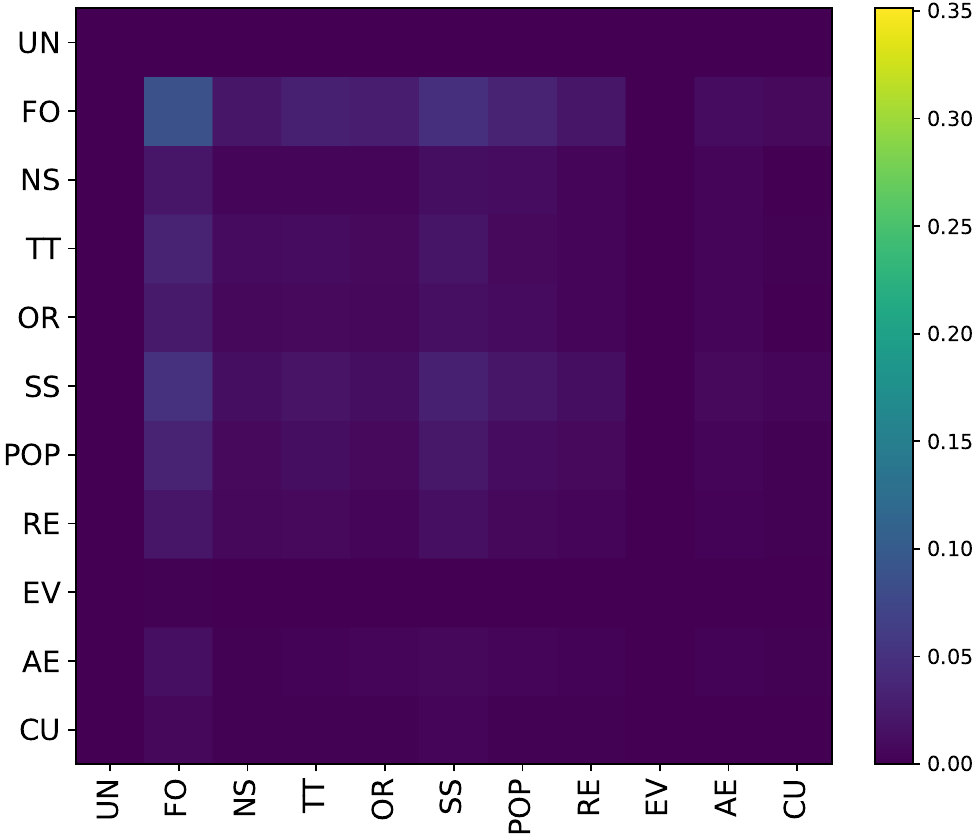}
    \caption{NYC: MoveSim}
\end{subfigure}
\hfill
\begin{subfigure}[t]{0.30\textwidth}
    \centering
    \includegraphics[width=\linewidth]{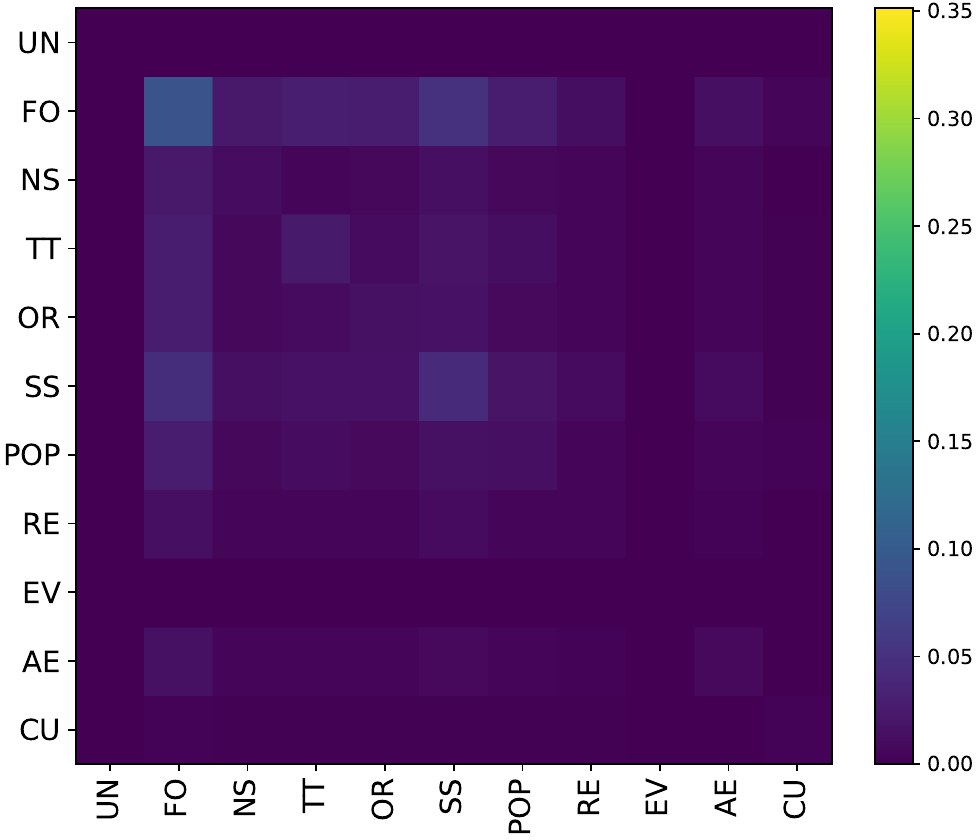}
    \caption{NYC: Geo-LLaMA}
\end{subfigure}
\hfill
\begin{subfigure}[t]{0.30\textwidth}
    \centering
    \includegraphics[width=\linewidth]{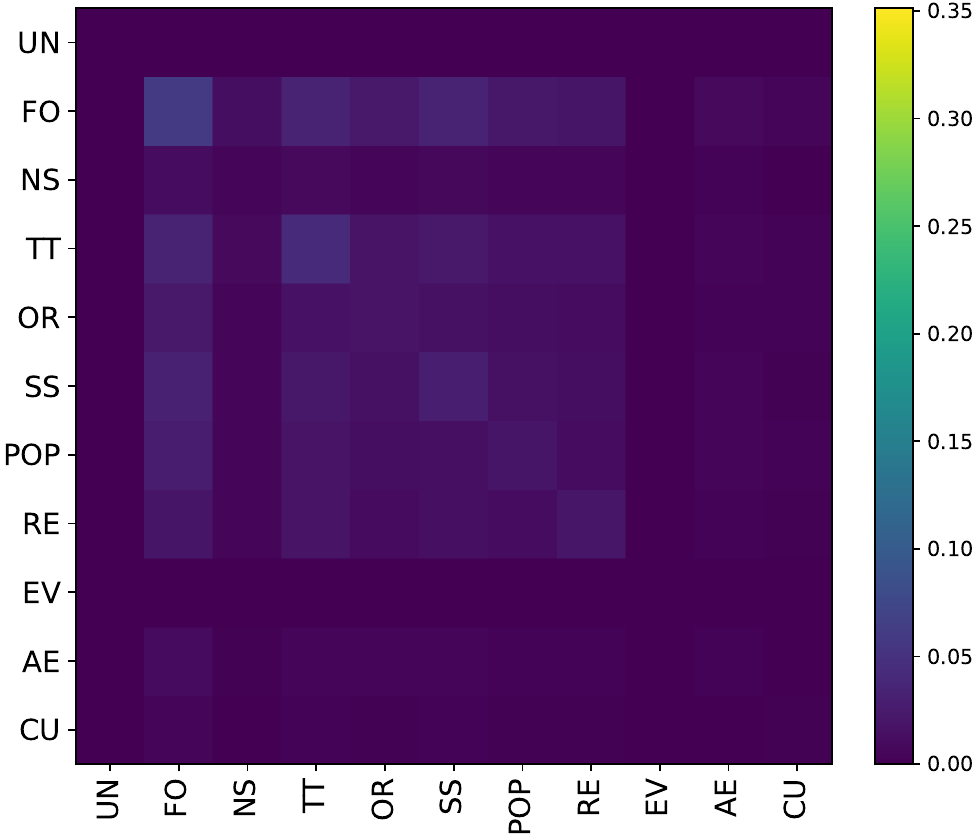}
    \caption{NYC: SynHAT (Ours)}
\end{subfigure}

\par\medskip

% Row 3: TKY (Real, DiffTraj, SeqGAN)
\begin{subfigure}[t]{0.30\textwidth}
    \centering
    \includegraphics[width=\linewidth]{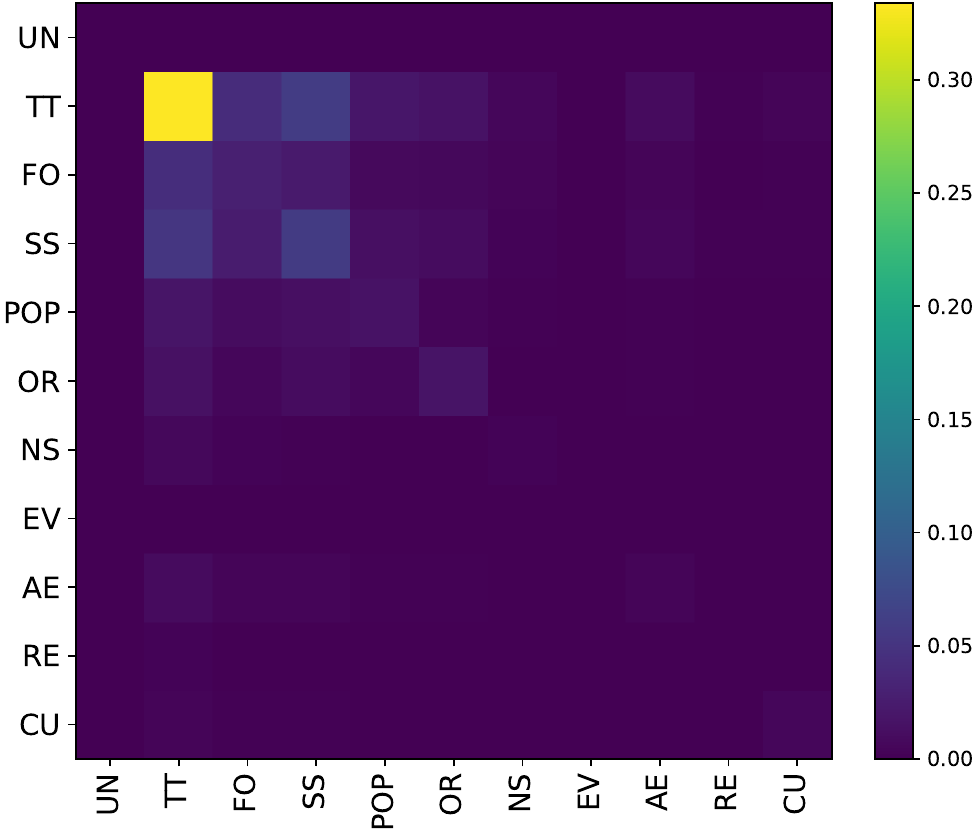}
    \caption{TKY: Real}
\end{subfigure}
\hfill
\begin{subfigure}[t]{0.30\textwidth}
    \centering
    \includegraphics[width=\linewidth]{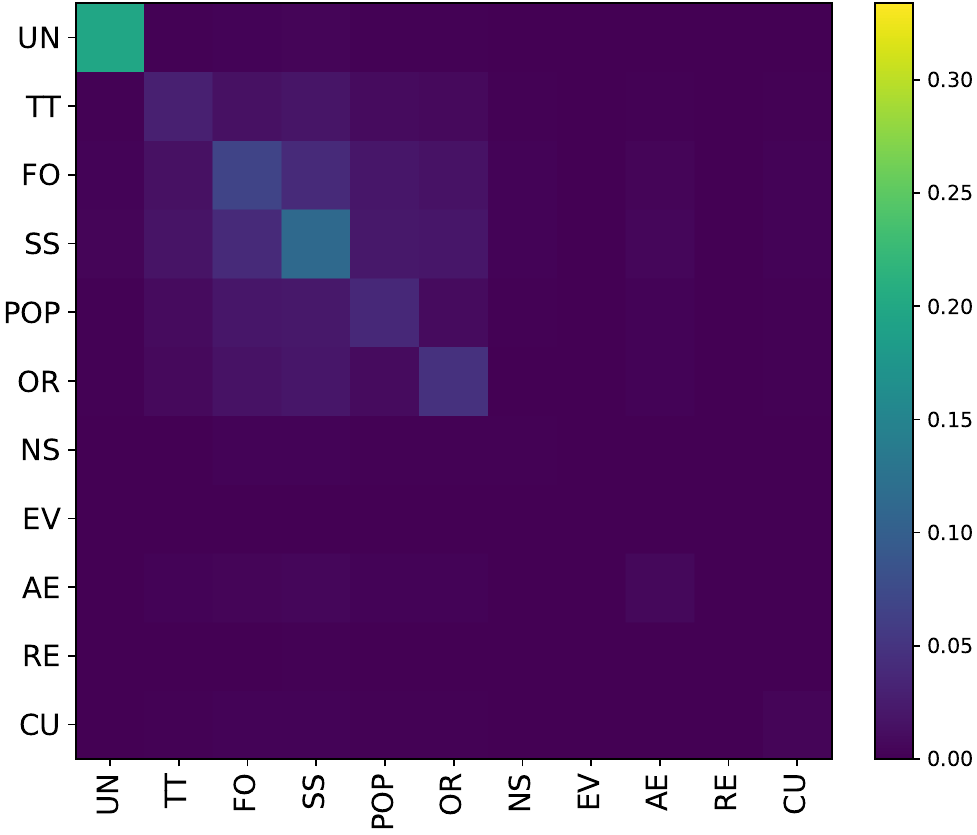}
    \caption{TKY: DiffTraj}
\end{subfigure}
\hfill
\begin{subfigure}[t]{0.30\textwidth}
    \centering
    \includegraphics[width=\linewidth]{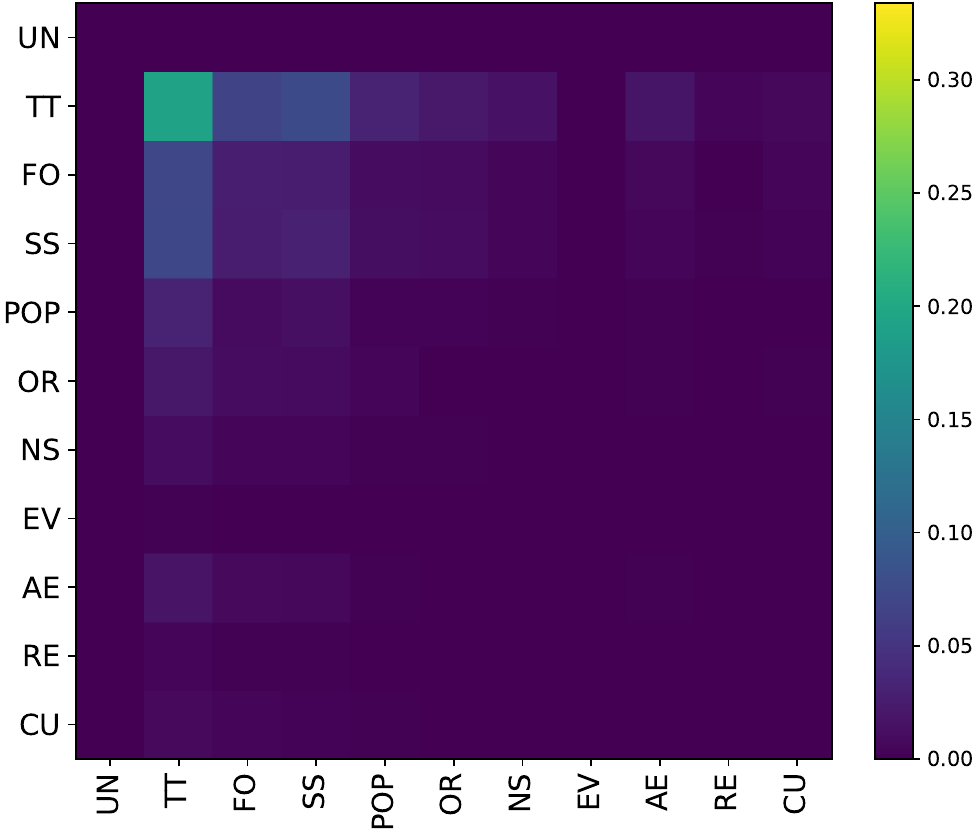}
    \caption{TKY: SeqGAN}
\end{subfigure}

\par\medskip

% Row 4: TKY (MoveSim, Geo-LLaMA, SynHAT)
\begin{subfigure}[t]{0.30\textwidth}
    \centering
    \includegraphics[width=\linewidth]{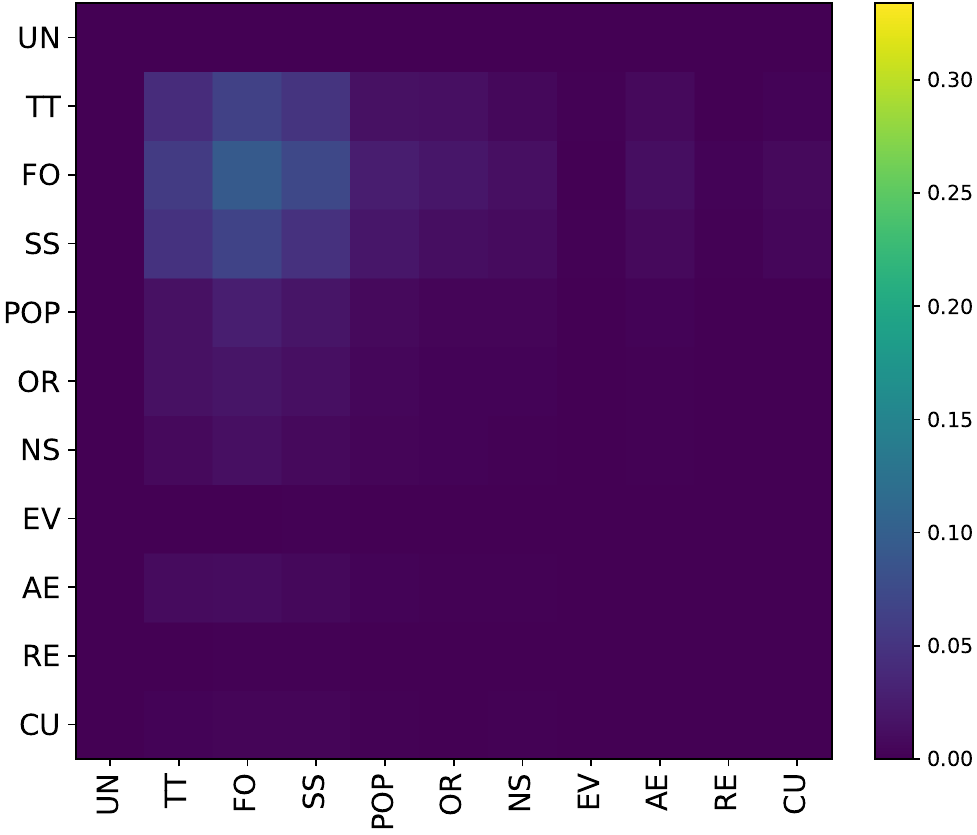}
    \caption{TKY: MoveSim}
\end{subfigure}
\hfill
\begin{subfigure}[t]{0.30\textwidth}
    \centering
    \includegraphics[width=\linewidth]{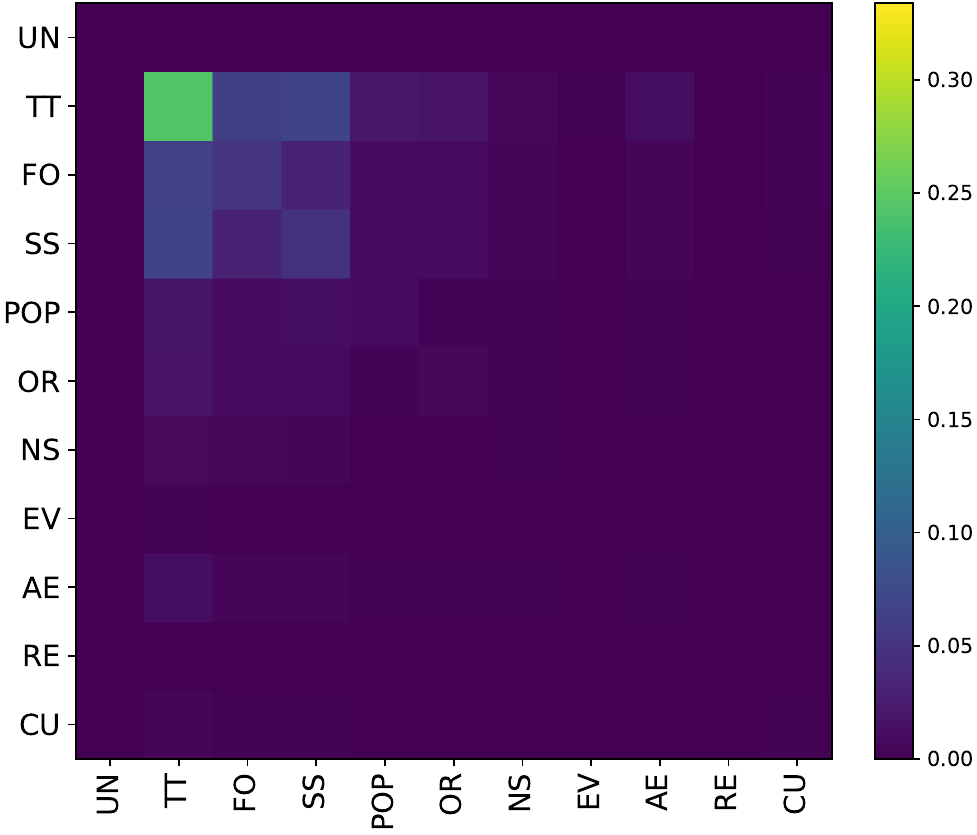}
    \caption{TKY: Geo-LLaMA}
\end{subfigure}
\hfill
\begin{subfigure}[t]{0.30\textwidth}
    \centering
    \includegraphics[width=\linewidth]{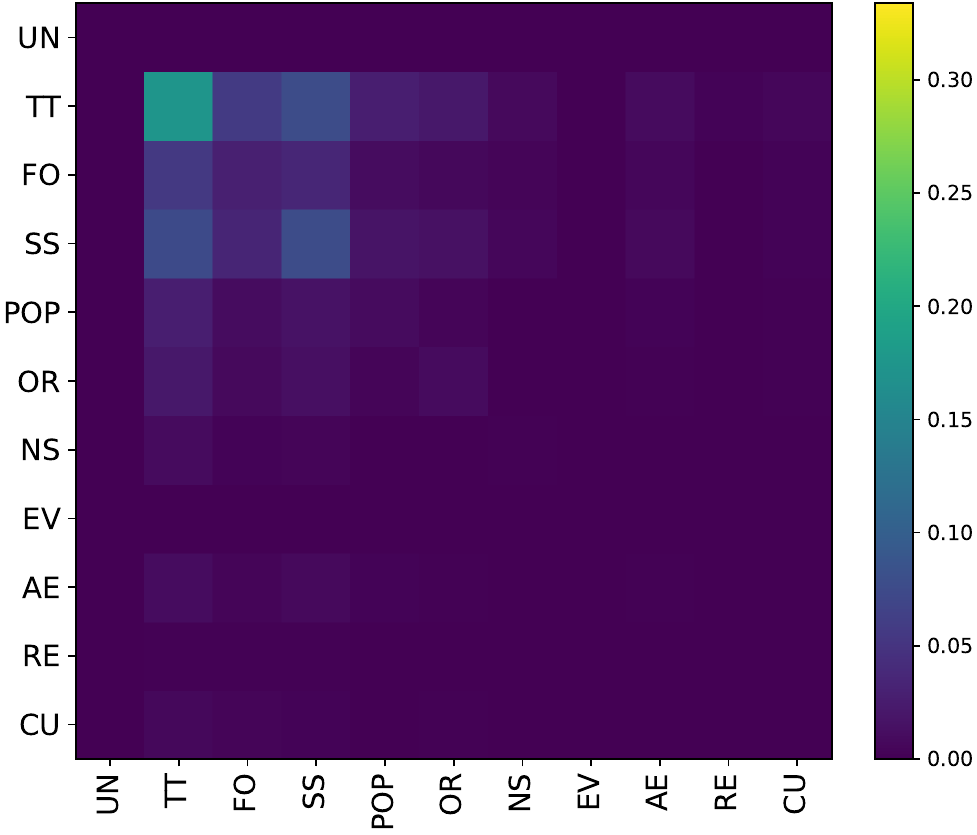}
    \caption{TKY: SynHAT (Ours)}
\end{subfigure}
\vspace{-10pt}
\caption{Activity transition density matrices of real and synthetic HAT datasets in NYC and TKY. 
}
\label{fig:tran_nyc_tky}
\end{figure*}

In the fidelity evaluation, activity transitions like “Work → Gym → Home” in synthetic HATs are also important.
We examined the high-level POI-anchored activity transitions in NYC, TKY, and ATX using Foursquare datasets.
Following the common practice~\cite{feng2020learning}, we define the high-level activities as follows: 
Food (FO),
Nightlife Spots (NS),
Travel \& Transport (TT),
Outdoors \& Recreation (OR),
Shops \& Services (SS),
Professional \& Other Places (POP),
Residences (RE),
Events (EV),
Arts \& Entertainment (AE),
College \& University (CU),
and Unknown (UN).
We compute the transition density matrix by aggregating consecutive activities in each HAT in the synthetic and real datasets.
We then compare the Jensen-Shannon Divergence (JSD) of each transition matrix of synthetic HATs with the real transition matrix, and the quantitative and qualitative results and visualizations are presented in Table~\ref{table:transition} and Figure~\ref{fig:tran_nyc_tky}, respectively.
The smaller the JSD is, the closer the transition matrix of the synthetic dataset is to the real transition matrix. 
It is evident that SynHAT provides strong activity-level transition modeling, ranking first on NYC and ATX and second on TKY. However,
continuous trajectory generation methods (e.g., DiffTraj and ControlTraj) struggle to capture activity transitions because they lack explicit activity-aware knowledge.

Although SynHAT does not explicitly incorporate prior knowledge about inter-event transitions, it still models activity transitions effectively. 
This is because the fine-grained Latent ST States themselves are trained to capture and reflect the underlying spatio-temporal dependencies of human movement.
Rather than relying on a hard-coded transition matrix, SynHAT models fine-grained Latent ST States of behavioral flow where the latent transitions inherently encapsulate logical behavioral sequencing. 
The Semantic Alignment module then acts as a grounded interface: by applying spatial filtering and temporal sampling under the specific constraints of these latent states, it ensures that the sampled POI-anchored activities remain faithful to the original intent of the generated trajectory. 
In other words, the logical behavioral transitions are maintained by the synthetic fine-grained latent ST states, while the alignment module simply maps these logical transitions into the physical POI space.

\begin{figure*}[t]
    \centering

    \begin{subfigure}[b]{\linewidth}
        \includegraphics[width=\linewidth]{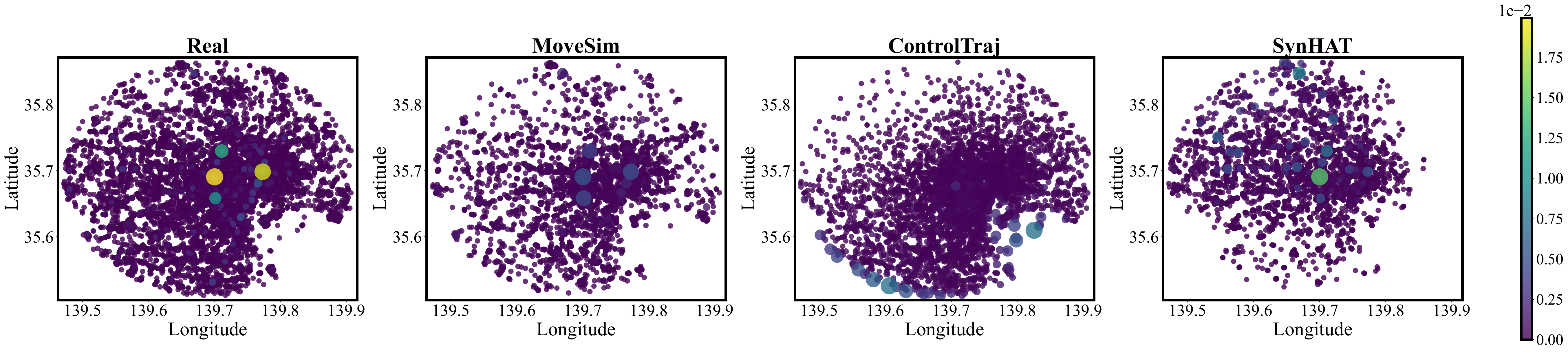}
        \caption{Density distribution in TKY.}
        \label{fig:density-tky}
    \end{subfigure}
    
    \begin{subfigure}[b]{\linewidth}
        \includegraphics[width=\linewidth]{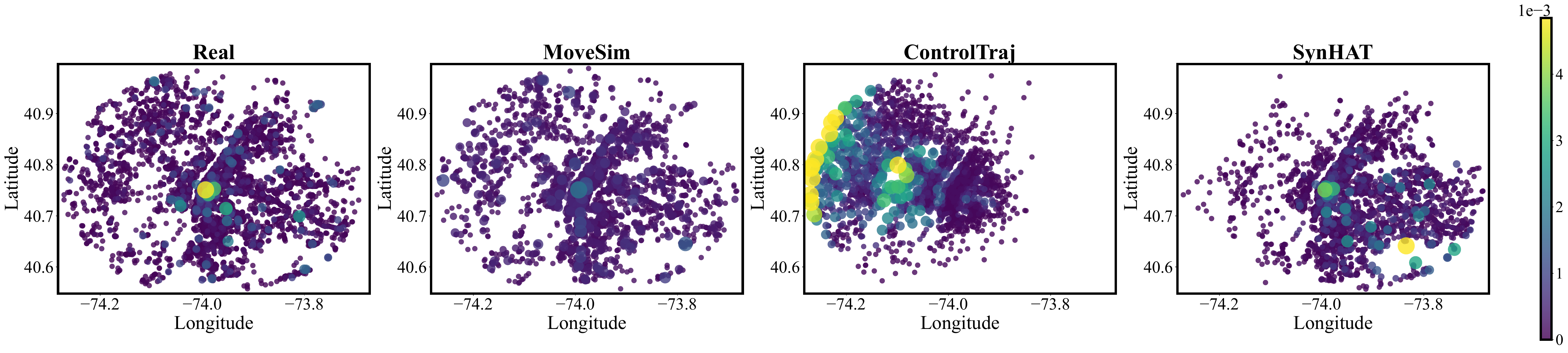}
        \caption{Density distribution in NYC}
        \label{fig:density-nyc}
    \end{subfigure}  
    
    \begin{subfigure}[b]{\linewidth}
        \includegraphics[width=\linewidth]{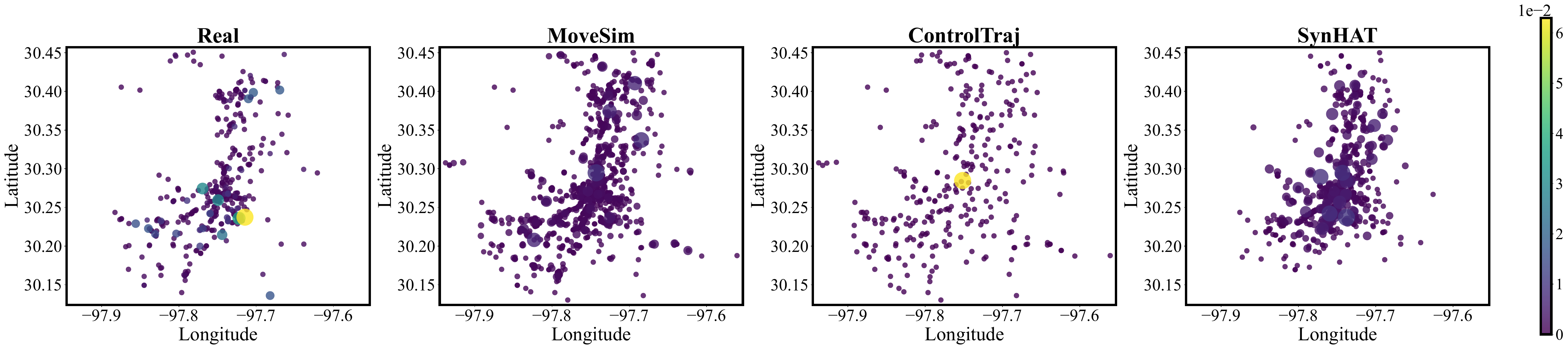}
        \caption{Density distribution in ATX.}
        \label{fig:density-atx}
    \end{subfigure}

    \begin{subfigure}[b]{\linewidth}
        \includegraphics[width=\linewidth]{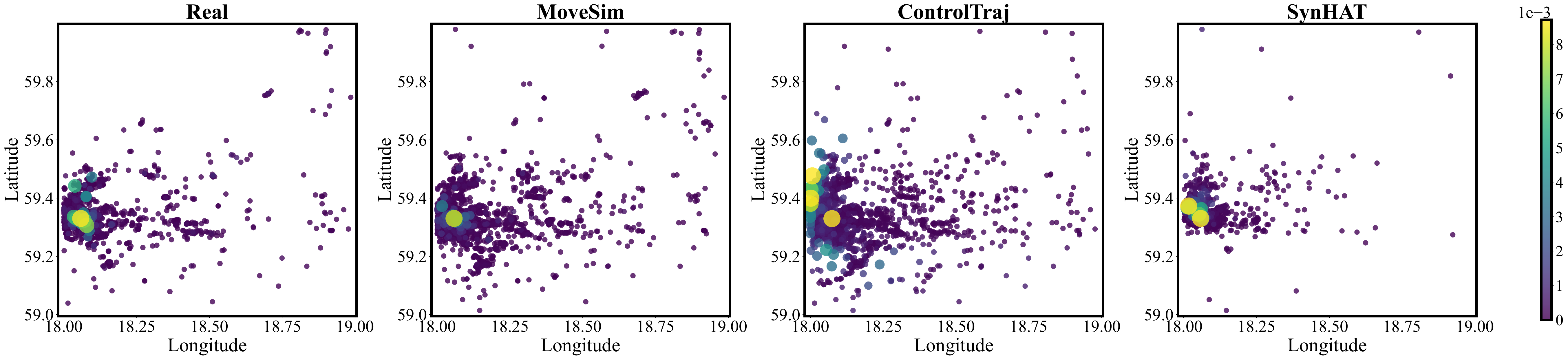}
        \caption{Density distribution in STO.}
        \label{fig:density-gw}
    \end{subfigure}
    
    \caption{POI-anchored activity distributions of different methods in four cities. The brighter the color is, the more frequently the POI is visited.}
    \label{fig:density}
\end{figure*}

We visualize the density of POI-anchored activity distributions in Fig.~\ref{fig:density}. Across all cities, \N\ best preserves both the spatial concentration and activity intensity, whereas other baselines either misplace major hubs or exhibit inconsistent density magnitudes. Specifically, MoveSim fails to highlight high-intensity regions effectively, resulting in less distinct urban centers and diluted hotspot structures. The results confirm the strong capability of \N\ to accurately replicate real-world spatial distributions across diverse urban environments.

\clearpage
\subsection{Extented Privacy-preserving Evaluations}

\begin{figure*}[!h]
\centering
\begin{subfigure}{0.49\textwidth}
    \includegraphics[width=\linewidth]{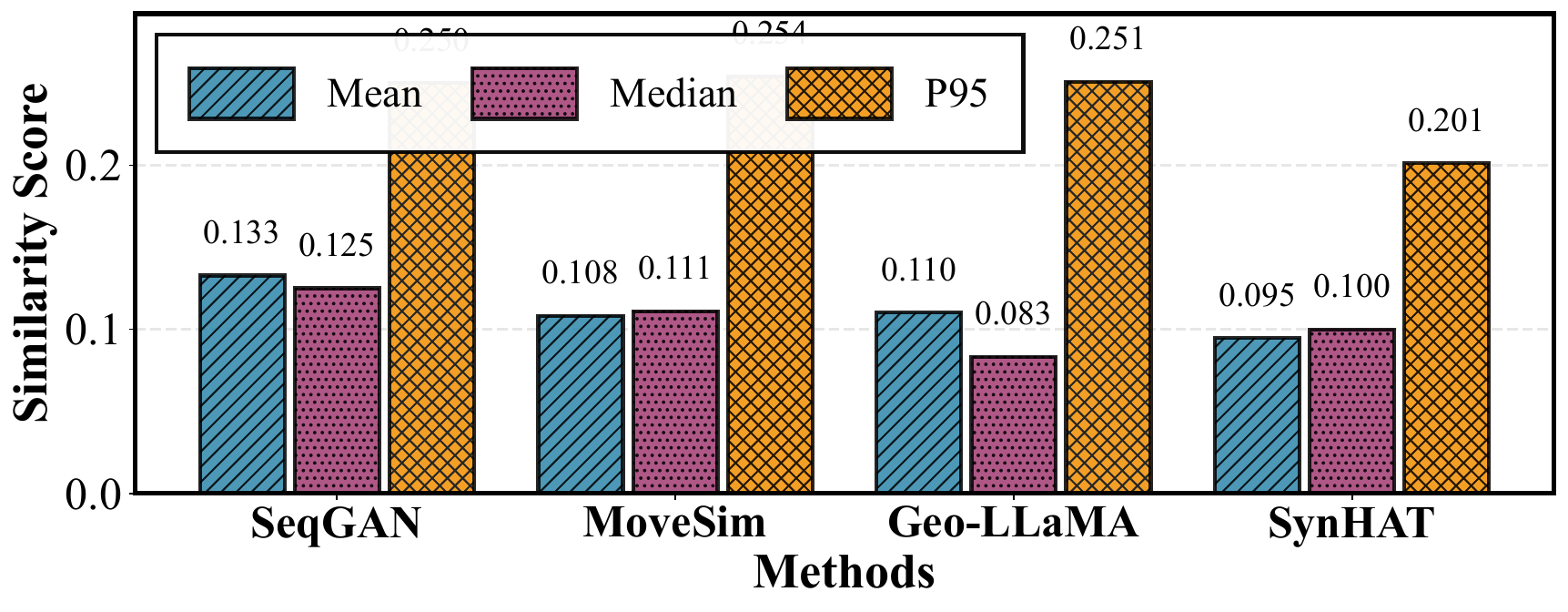}
    \caption{NYC ($tr_s$ = 0.2km, $tr_t$ = 0.5h)}
\end{subfigure}
\hfill
\begin{subfigure}{0.49\textwidth}
    \includegraphics[width=\linewidth]{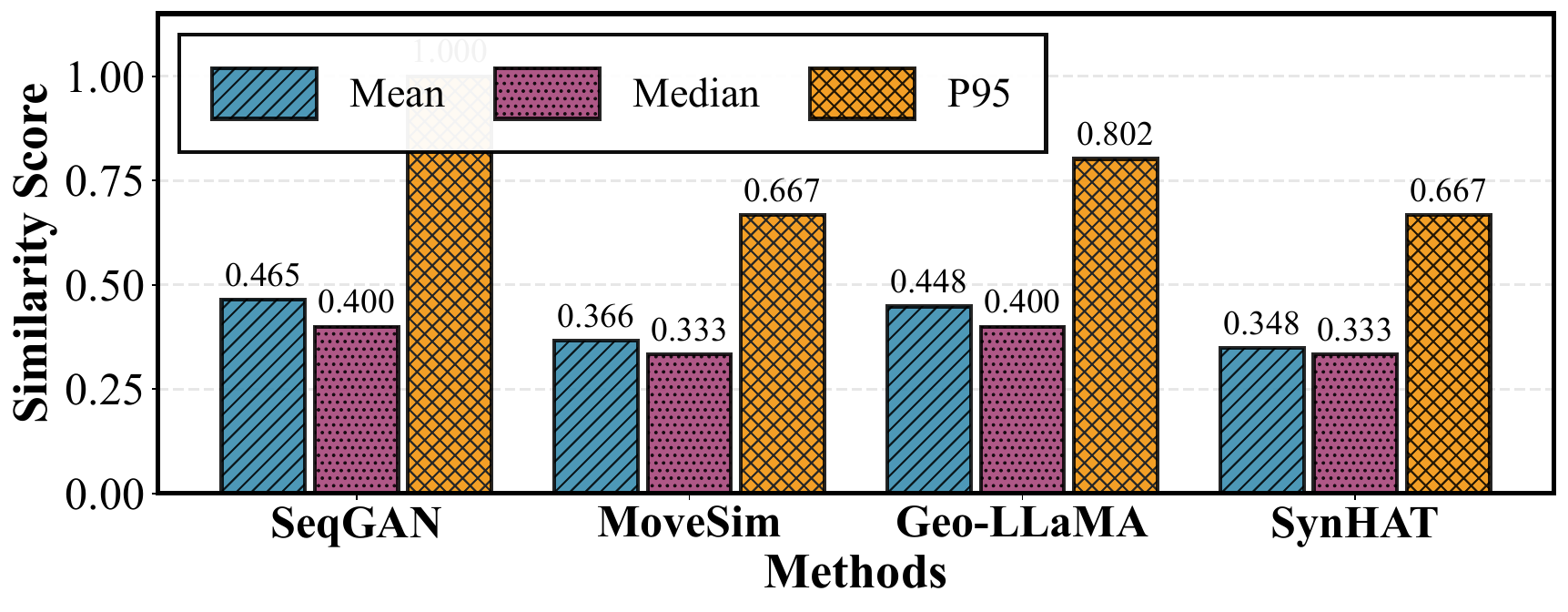}
    \caption{NYC ($tr_s$ = 2km, $tr_t$ = 2h)}
\end{subfigure}
\vspace{1ex}

\begin{subfigure}{0.49\textwidth}
    \includegraphics[width=\linewidth]{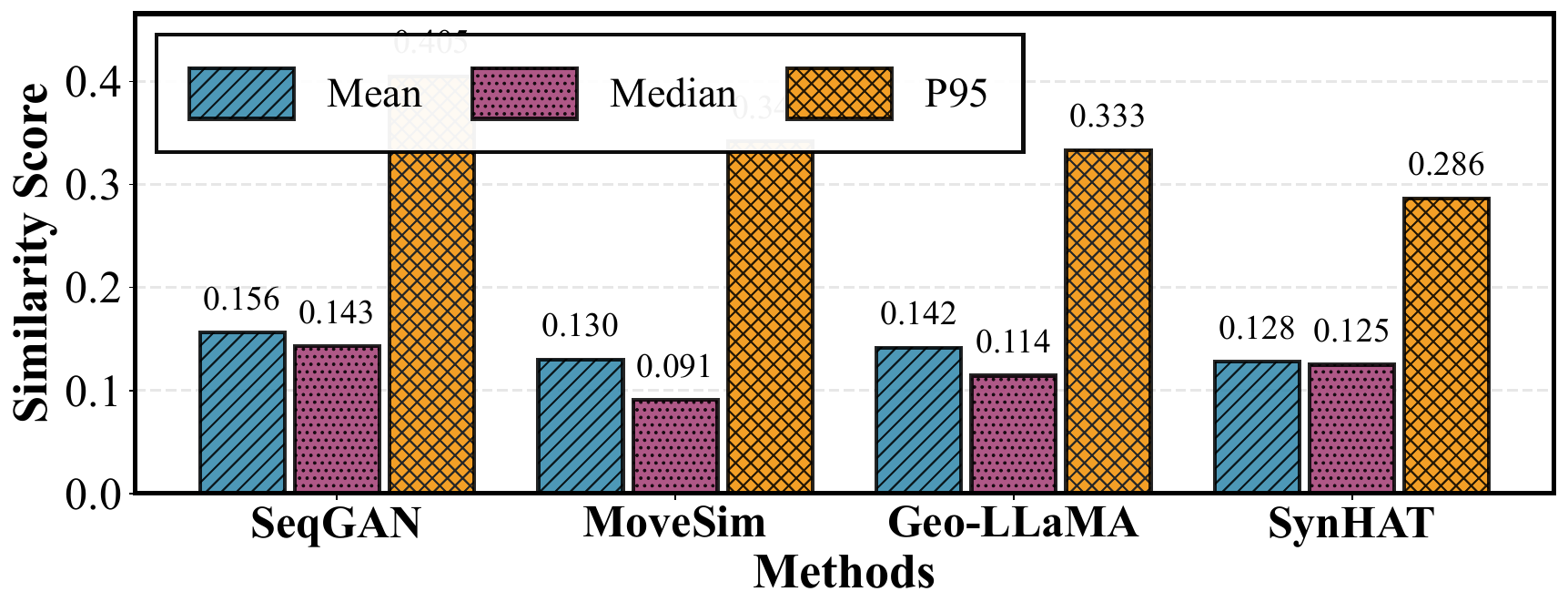}
    \caption{TKY ($tr_s$ = 0.2km, $tr_t$ = 0.5h)}
\end{subfigure}
\hfill
\begin{subfigure}{0.49\textwidth}
    \includegraphics[width=\linewidth]{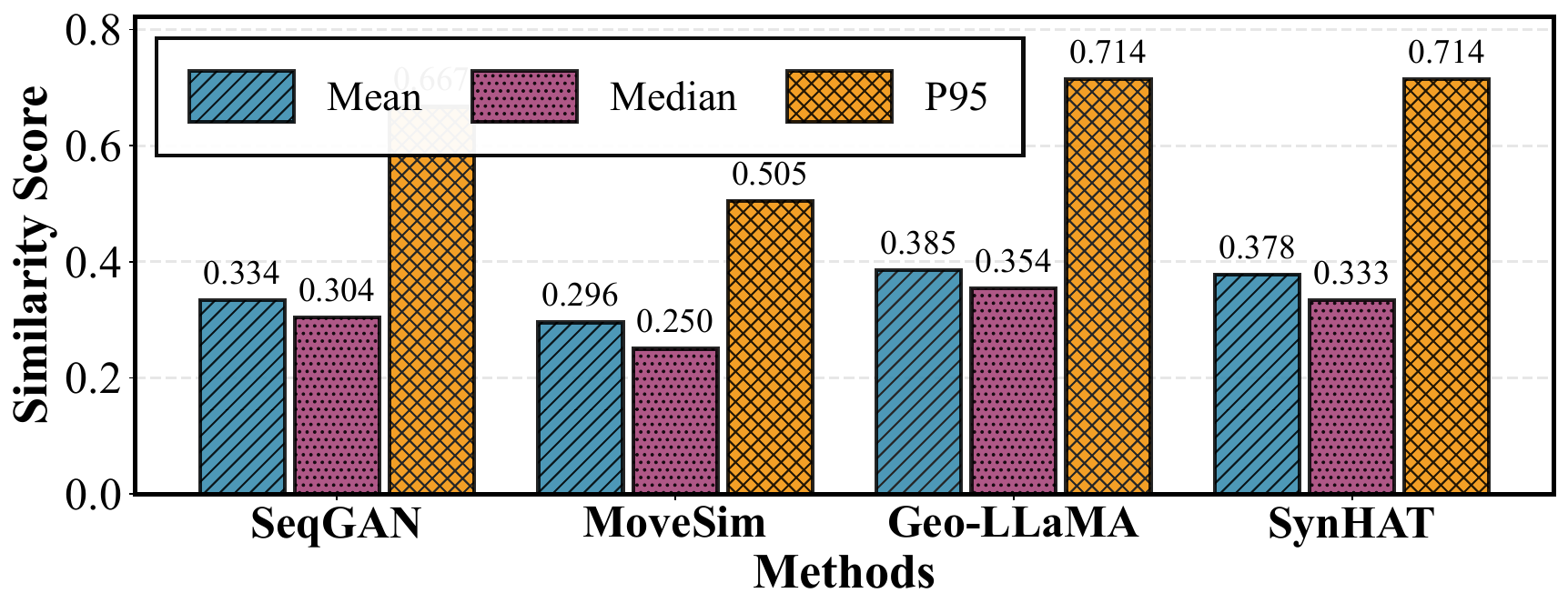}
    \caption{TKY ($tr_s$ = 2km, $tr_t$ = 2h)}
\end{subfigure}

\caption{Privacy-preserving statistics for HAT generation in NYC and TKY. P95 denotes the 95th percentile of similarity values in the synthetic datasets.}
\label{fig:privacy_bar}
\end{figure*}

Aside the Figure~\ref{fig:privacy_cdf} that qualitatively shows the privacy-preserving effectiveness, the statistics of the similarity distribution of each method in NYC and TKY are shown in Figure~\ref{fig:privacy_bar}.
Interestingly, across most settings, \N~achieves the lowest 95th-percentile (P95) similarity, indicating the smallest upper-tail memorization and thus the tightest upper bound on privacy leakage among all methods.
Although \N~does not rank first under the looser thresholds (e.g., TKY with \(tr_s=2\)\,km, \(tr_t=2\)\,h; ATX with \(tr_s=2\)\,km, \(tr_t=2\)\,h), this reflects the inherent fidelity–privacy trade-off; relative to other baselines, \N\ attains the strongest overall fidelity and downstream utility.

% \begin{figure*}
% \centering
% \begin{subfigure}{0.49\textwidth}
%     \includegraphics[width=\linewidth]{Figures/bars/privacy_bars_nyc_tolm200_tolt0.pdf}
%     \caption{TKY ($tr_s$ = 0.2km, $tr_t$ = 0.5h)}
% \end{subfigure}
% \hfill
% \begin{subfigure}{0.49\textwidth}
%     \includegraphics[width=\linewidth]{Figures/bars/privacy_bars_nyc_tolm2000_tolt2.pdf}
%     \caption{NYC ($tr_s$ = 0.2km, $tr_t$ = 0.5h)}
% \end{subfigure}
% \vspace{1ex}

% \begin{subfigure}{0.49\textwidth}
%     \includegraphics[width=\linewidth]{Figures/bars/privacy_bars_tky_tolm200_tolt0.pdf}
%     \caption{TKY ($tr_s$ = 0.2km, $tr_t$ = 0.5h)}
% \end{subfigure}
% \hfill
% \begin{subfigure}{0.49\textwidth}
%     \includegraphics[width=\linewidth]{Figures/bars/privacy_bars_tky_tolm2000_tolt2.pdf}
%     \caption{NYC ($tr_s$ = 0.2km, $tr_t$ = 0.5h)}
% \end{subfigure}

% \caption{Privacy-preserving statistics for HAT generation in NYC and TKY. P95 denotes the 95th percentile of similarity values in the synthetic datasets. }
% \label{fig:privacy_bar}
% \end{figure*}   

% preamble:
% \usepackage{stfloats}

\end{document}